\documentclass{article}

\usepackage[preprint]{neurips_2021}

\usepackage{xcolor}
\definecolor{darkblue}{rgb}{0.0, 0.0, 0.55}
\definecolor{anton}{rgb}{0, .67, .54}

\PassOptionsToPackage{square,numbers,sort&compress}{natbib}
\usepackage[colorlinks=true, bookmarksopen,
			linkcolor={darkblue},
			anchorcolor={black},
			citecolor={darkblue},
			filecolor={magenta},
			menucolor={black},
			plainpages=false,pdfpagelabels,
			urlcolor={darkblue}]{hyperref}

\usepackage[utf8]{inputenc} 
\usepackage[T1]{fontenc}	
\usepackage{url}			
\usepackage{booktabs}	   
\usepackage{amsfonts}	   
\usepackage{nicefrac}	   
\usepackage{microtype}	  
\usepackage{amsmath, amsthm}
\usepackage{bm}
\usepackage{graphicx}
\usepackage{chngcntr}
\usepackage[font=small, belowskip=-1em]{caption}
\usepackage{wrapfig}
\usepackage{bbm}

\newtheorem{theorem}{Theorem}

\newtheorem*{remark}{Remark}

\title{Offline Policy Comparison under Limited Historical Agent-Environment Interactions}

\author{%
	Anton Dereventsov \\
	Lirio AI Research\\
	Lirio, LLC\\
	Knoxville, TN 37923\\
	\texttt{adereventsov@lirio.com}
	\And
	Joseph Daws,~Jr.\\
	Lirio AI Research\\
	Lirio, LLC\\
	Knoxville, TN 37923\\
	\texttt{jdaws@lirio.com}
	\And
	Clayton G.~Webster\thanks{Behavioral Reinforcement and Learning Lab (BReLL), Lirio, LLC.}\\
	Lirio AI Research\\
	Lirio, LLC\\
	Knoxville, TN 37923\\
	\texttt{cwebster@lirio.com}
}

\begin{document}

\maketitle

\begin{abstract}
We address the challenge of policy evaluation in real-world applications of reinforcement learning systems where the available historical data is limited due to ethical, practical, or security considerations.
This constrained distribution of data samples often leads to biased policy evaluation estimates.
To remedy this, we propose that instead of policy evaluation, one should perform policy comparison, i.e. to rank the policies of interest in terms of their value based on available historical data.
In addition we present the Limited Data Estimator (LDE) as a simple method for evaluating and comparing policies from a small number of interactions with the environment.
According to our theoretical analysis, the LDE is shown to be statistically reliable on policy comparison tasks under mild assumptions on the distribution of the historical data.
Additionally, our numerical experiments compare the LDE to other policy evaluation methods on the task of policy ranking and demonstrate its advantage in various settings.
\end{abstract}

\section{Introduction}\label{sec:intro}
In recent years, the field of Reinforcement Learning (RL) has seen tremendous scientific progress impacting a wide range of applications: from classic tabletop games~\citep{silver2016mastering, silver2018general} to modern video games~\citep{mnih2013playing, berner2019dota, jaderberg2019human}, and even problems such as protein folding~\citep{alquraishi2019alphafold, AlphaFold2020}, optimizing chemical reactions~\citep{chou2017optimize}, resource management in computer clusters~\citep{mao2016resource}, robotics applications~\citep{kober2013reinforcement, levine2016end}, and many more.

However, while the advances of RL algorithms when applied to simulated environments are impressive and often surpass human-expert-level performance, see e.g.~\citep{silver2017mastering, vinyals2019grandmaster}, real-world applications typically impose a set of grand challenges, see e.g.~\citep{dulac2019challenges, dulac2020empirical, dulac2021challenges}, that often render conventional RL approaches infeasible, see e.g.~\citep{rlblogpost} and the references therein.

A particular difficulty of real-world RL is the restriction of the available logged data, specifically in applications involving agent-human communications. In this setting one generally cannot deploy an unverified agent with the intent of either policy learning or evaluation from the received feedback (i.e. online RL) and instead has to rely on algorithms that exclusively utilize previously collected {\it historical data} (i.e. offline RL, see e.g.~\citep{agarwal2020optimistic, levine2020offline, monier2020offline}.

While offline policy evaluation is an actively developing field, see e.g.~\citep{thomas2015confidence, Tennenholtz2020off, fu2021benchmarks} and the references therein, most conventional approaches require large amounts of historical data and rely on the assumption of data ergodicity which, in practice, can not be guaranteed~\citep{aslanides2017universal, bojun2020steady}.
Violating these assumptions results in a biased prediction of the policy towards the distribution of the historical data.
To combat this difficulty, similarly to~\citep{fu2021benchmarks} we argue that a more achievable and realistic alternative is to employ policy evaluation methods in order to \textit{compare} the target policies instead of evaluating them directly.
In other words, given several policies, arrange them from worst to best in terms of their value but not necessarily provide a precise estimate of the value itself.

For the purpose of this effort we consider a general contextual bandit setting with particular focus on the challenge precipitated by {\it limited historical data}, which encompasses the infeasibility of obtaining sufficiently large amounts of historical interactions, either by ethical, practical, or security considerations~\citep{Tennenholtz2020off}.
Under this assumption, our primary objective is to compare and evaluate (whenever possible) target policies using different policy evaluation methods.

Comparing the performance of policies is an important task for instances of RL (e.g., contextual bandit) problems where it is expensive, difficult, or unethical to deploy untested or unverified agents. Such comparisons may be achieved by directly estimating the value of the policies as well as policy correlations~\citep{voloshin2020empirical, fu2021benchmarks, irpan2019off, doroudi2018importance, paine2020hyperparameter}.
However, such classic and current popular methods rely on obtaining a sufficiently large number of agent-environment interactions in order to approximate probabilistic quantities of interest, e.g. the expected cumulative reward of a policy.
As such, we introduce the Limited Data Estimator (LDE): a quantifier suitable for comparing the performance of two or more undeployed policies using only a small number of interactions with the environment, which may be collected using a behavior policy whose parameterization is unknown.
Furthermore, we show that, in some cases, LDE is also a good approximation of the true policy value.

Finally, solving real-world problems often involves heuristic choices and assumptions about the underlying contextual bandit model, e.g. the choice neural network architecture used to parameterize a policy.
In such cases, it is difficult to verify the performance of a policy trained to solve a specific task~\citep{flaxman2005online}.
In addition, according to our theoretical analysis, the LDE is shown to be statistically reliable on policy comparison tasks under mild assumptions on the distribution of the historical data.

Our main contributions are the following:
\begin{itemize}
    \item We describe the \textit{limited data} setting which encompasses many critical real-world applications where the amount of available historical data is not sufficient for reliable policy evaluation.
    \item We formalize the task of \textit{policy comparison} as a generalization of policy evaluation and prove that this task is attainable even in cases of severely limited data.
    \item We introduce the \textit{Limited Data Estimator} and verify, both theoretically and numerically, the efficiency of this method on the policy comparison task.
\end{itemize}

\subsection{Related Work}\label{sec:related_work}
The area of policy evaluation in the contextual bandit setting has expanded in recent years, see e.g.~\citep{thomas2015high, agarwal2016making, wang2017optimal, vlassis2019design}.
In this paper we consider the following policy evaluation methods: the Direct Method, the Inverse Propensity Scoring~\citep{horvitz1952generalization, precup2000eligibility, agarwal2014taming}, and the Doubly Robust Estimator~\citep{robins1995semiparametric, bang2005doubly, dudik2011doubly, dudik2014doubly}.

An active area of research is to adjust the aforementioned methods to be applied to general reinforcement learning~\citep{jiang2016doubly}, to further reduce the variance~\citep{farajtabar2018more}, to incorporate a model estimate~\citep{thomas2016data}, or to account for long horizons~\citep{xie2019towards}.
Moreover, in applications where a sufficient amount of historical interactions are recorded, there is an interest in methods that can leverage the available data samples to learn from the logged data, see e.g.~\citep{strehl2010learning, swaminathan2015batch, swaminathan2015counterfactual, joachims2018deep}.
 
In this paper we consider the task of policy comparison in scenarios where a sufficient amount of logged interactions is not available.
Similar settings are considered in a concurrent work~\citep{fu2021benchmarks} where the authors provide related arguments and establish a set of benchmarks for the offline policy comparison, some of which are applicable in our case.
However, due to the recent emergence of this development, we do not utilize the proposed benchmarks in the current submission.

\section{Background}\label{sec:background}
In this paper we consider a general contextual bandit setting proposed in~\citep{langford2007epoch}.
Such a setting is not restrictive since we address the problem of offline policy evaluation, and thus, with appropriate modifications, any set of logged data from an RL environment can be viewed as a data from the contextual bandit, as we demonstrate in Section~\ref{sec:ex3}.

We use the notational standard MDPNv1~\citep{thomas2015notation}.
Namely, $\mathcal{S}$ denotes the state (context) space, $\mathcal{A}$ denotes the action space, $\mathcal{R} \subset \mathbb{R}$ denotes the reward space, and $r : \mathcal{S} \times \mathcal{A} \to \mathcal{R}$ denotes the reward function.
We assume that the reward space is bounded and that the reward function is deterministic.

While in the literature it is conventionally assumed that the logged data is collected under a dedicated exploratory policy, often in practice the history of interactions comes from different sources rather than a behavioral policy.
For the convenience of presentation, throughout the paper we define the {\it historical data} $\mathcal{D}$ as a set of $N > 0$ four-tuples consisting of an observed state $s \in \mathcal{S}$, a taken action $a \in \mathcal{A}$, a received reward $r(s,a)$, and a behavioral probability $p$:
\begin{equation}\label{eq:historical_data}
    \mathcal{D} = \big\{ (s_n, a_n, r_n, p_n) \big\}_{n=1}^N
\end{equation}
For simplicity, we assume that each tuple in $\mathcal{D}$ is unique, i.e. that each state-action pair was observed at most once or, equivalently, that a duplicate observation is overwritten by the most recent interaction.
We also note that in practice the exact behavioral probabilities $\{p_1, \ldots, p_N\}$ might be unknown.
However, for simplicity of implementation and comparison, we assume that the exact values of behavior probabilities are available.
In the case of our work this assumption can be relaxed as the Limited Data Estimator, that we introduce in Section~\ref{sec:lde}, does not depend on behavioral probabilities.

The value of a policy $\pi$ is the expected reward under this policy, i.e.
\begin{equation}\label{eq:policy_value}
    V(\pi) = \mathbb{E}_s \big[ r(s,a) \,|\, a \sim \pi(s) \big].
\end{equation}
A task of policy evaluation is the following: given a policy $\pi$ and a set of historical data $\mathcal{D}$, estimate the value $V(\pi)$.
In Section~\ref{sec:pem} we state some well-known methods for policy evaluation.

\subsection{Policy Evaluation Methods}\label{sec:pem}
Throughout this section let $\pi$ denote the target policy to be evaluated under the set of the available historical data $\mathcal{D}$, defined by~\eqref{eq:historical_data}.
The policy evaluation methods utilize the provided historical data to approximate the value $V(\pi)$ of the target policy, defined in~\eqref{eq:policy_value}.
For notational simplicity throughout the rest of the paper we do not specify the dependence on the historical dataset $\mathcal{D}$ and omit it from the value estimates.
References to the discussed methods can be found in Section~\ref{sec:related_work}.

\paragraph{Direct Method (DiM)}
The Direct Method (DiM) is a straightforward estimate of the policy value $V(\pi)$ that makes use of the recorded reward values but does not attempt to approximate the reward outside of the observed interactions:
\begin{equation}\label{eq:dim}
    V_{DiM}(\pi) = \frac{1}{N} \sum_{n=1}^N r_n \, \pi(a_n|s_n).
\end{equation}
Note that in some literature the Direct Method is referred to a different estimate, typically based on the approximation of the reward function.

\paragraph{Inverse Propensity Scoring (IPS)}
The Inverse Propensity Scoring (IPS) accounts for the potentially unequal data sampling density by making use of the recorded probabilities $p_n$ and, similarly to DiM, does not approximate the reward outside of the observed interactions:
\begin{equation}\label{eq:ips}
    V_{IPS}(\pi) = \frac{1}{N} \sum_{n=1}^N r_n \frac{\pi(a_n|s_n)}{p_n}.
\end{equation}
Despite the apparent simplicity, the IPS is \textit{unbiased}, i.e. the provided value estimate converges to the true value of the policy in expectation given enough data samples.

\paragraph{Doubly Robust Estimator (DRE)}
The Doubly Robust Estimator (DRE) not only accounts for the potentially unequal data sampling density but also utilizes an additional baseline function $\hat{r} : \mathcal{S} \times \mathcal{A} \to \mathbb{R}$ that approximates the reward on unobserved state-action pairs:
\begin{equation}\label{eq:dre}
    V_{DRE}(\pi) = \frac{1}{N} \sum_{n=1}^N \Bigg[ \big(r_n - \hat{r}(s_n,a_n)\big) \frac{\pi(a_n|s_n)}{p_n}
        + \sum_{a \in \mathcal{A}} \hat{r}(s_n,a) \, \pi(a|s_n) \Bigg].
\end{equation}
The DRE is typically the default option for policy evaluation since it's estimate is \textit{unbiased} as long as either the behavioral probability or the baseline is precise, and usually has a small variance.

\section{The Limited Data Setting}\label{sec:limited_data}
In related literature it is often conventional to imply an assumption of \textit{ergodicity} of the historical data, which guarantees the expected behavior of the policy evaluation methods, see e.g.~\citep{wang2020reliable, karampatziakis2021off}.
However such an assumption is unfeasible in many critical applications where one does not have the freedom of interacting with the environment arbitrarily frequently, e.g.~\citep{hauskrecht2000planning, shameer2018machine, esteva2019guide}.

As discussed in Section~\ref{sec:intro}, for many real-world scenarios it is not possible to collect large amounts of interaction data, see also~\citep{fu2021benchmarks}.
For example, for complex decision making that involves human feedback, e.g.~\citep{tomkins2020rapidly, hassouni2021ph}, it is too resource intensive and possibly unethical to collect a large interaction history using untested behavior policies.
In addition, since human behavior is difficult to model, see e.g.~\citep{enkhtaivan2020model}, creating meaningful simulations of environments involving human feedback is infeasible.

In this effort we focus on the \textit{limited data} setting where one cannot obtain sufficiently many recorded data points to reliably estimate the value of the target policy.
Essentially, in the limited data paradigm, estimating the expected reward for a target policy is equivalent to estimating the expected value of a random variable using just a few samples, which is generally unfeasible.

\subsection{Policy Comparison Task}\label{sec:policy_comparison}
In real-world applications practitioners are often faced with the problem of selecting the best policy out of (potentially many) available options, e.g. the policies that are either trained offline or on a simulator.
The most straightforward way to establish the superior policy from a collection is to estimate their values, either by deploying or by employing a policy evaluation method.
Unfortunately, as discussed in Section~\ref{sec:intro}, in many critical applications direct deployment is generally not feasible, and policy evaluation is typically not reliable under the limited data setting.
Hence it is desirable to have an alternative approach for establishing the superiority relationship between the policies of interest without explicitly predicting their values.
In such scenarios we propose to consider the task of \textit{policy comparison} which we formalize next.

As discussed above, in the limited data setting an estimate of the policy's value is likely to be biased due to the constrained distribution of the historical data, which renders the use of established policy evaluation methods as generally not reliable.
This challenge motivates our interest in a closely related problem~--- \textit{policy comparison}~--- in place of \textit{policy evaluation}.
Given a set of target policies $\{\pi_1, \ldots, \pi_K\}$ with $K \ge 2$ and the historical data $\mathcal{D}$, the \textit{policy comparison} problem is to sort this set with respect to the expected rewards, i.e. the values $\{V(\pi_1), \ldots, V(\pi_K)\}$, by utilizing the logged interactions $\mathcal{D}$.

Specifically, consider the target policies $\pi$ and $\mu$, and the historical data $\mathcal{D}$.
In essence, policy comparison is executed as follows: compute and compare the value surrogates $\widetilde{V}(\pi;\mathcal{D})$ and $\widetilde{V}(\mu;\mathcal{D})$.
The method used to compute a surrogate value is decided by an architect and should be chosen in a problem dependent way. 
Once the value surrogates are computed, the comparison is straightforward: if $\widetilde{V}(\pi;\mathcal{D}) > \widetilde{V}(\mu;\mathcal{D})$ then the policy $\pi$ is predicted to be better than $\mu$ and vice versa.
In the case of multiple target policies $\{\pi_1, \ldots, \pi_K\}$ the comparison is performed pair-wise and, due to transitivity, the target policies may be sorted according to their surrogate values.
Note also that while the number of pairs scales quadratically in $K$, the surrogate values $\widetilde{V}(\pi_k;\mathcal{D})$ only need to be computed once for each $1 \le k \le K$.

Notice that the policy comparison task is more general than the policy evaluation as it does not require one to estimate the expected reward.
Instead, it is sufficient to obtain a surrogate that is correlated with the underlying value of the policy.
However, the obvious way to compare policies is to employ a policy evaluation method to estimate the target policies' values and then to simply compare these values.
We note that in case of policy comparison the potential biasedness of the provided value estimates is not as critical since the goal is to compare the target policies and not necessarily to predict their expected return, and this goal will be achieved as long as the values' surrogates are biased ``in the same way''.

Thus even though both policy comparison and policy evaluation can be solved by the same methods, the outcome of policy comparison is typically appropriate even if the value surrogate is biased.
Our theoretical results in Section~\ref{sec:theory} and numerical experiments in Section~\ref{sec:numerics} show that even
when the policy evaluation task may not be solvable due to a paucity of data, it is still possible to solve the policy comparison task with high confidence.

\section{Limited Data Estimator}\label{sec:lde}
The Limited Data Estimator (LDE) is defined as
\begin{equation}\label{eq:lde}
V_{LDE}(\pi) = \frac{1}{N} \sum_{n=1}^N \Big[ \big(r_n - \hat{r}(s_n,a_n)\big) \, \pi(a_n|s_n)
        + \sum_{a \in \mathcal{A}} \hat{r}(s_n,a) \, \pi(a|s_n) \Big]
\end{equation}
where $\hat{r} : \mathcal{S} \times \mathcal{A} \to \mathbb{R}$ is a baseline function.
In essence, the LDE can be viewed as a generalization of the DiM that employs a baseline to approximate the reward on unobserved state-action pairs.

\begin{wrapfigure}{r}{.35\linewidth}
    \includegraphics[width=\linewidth]{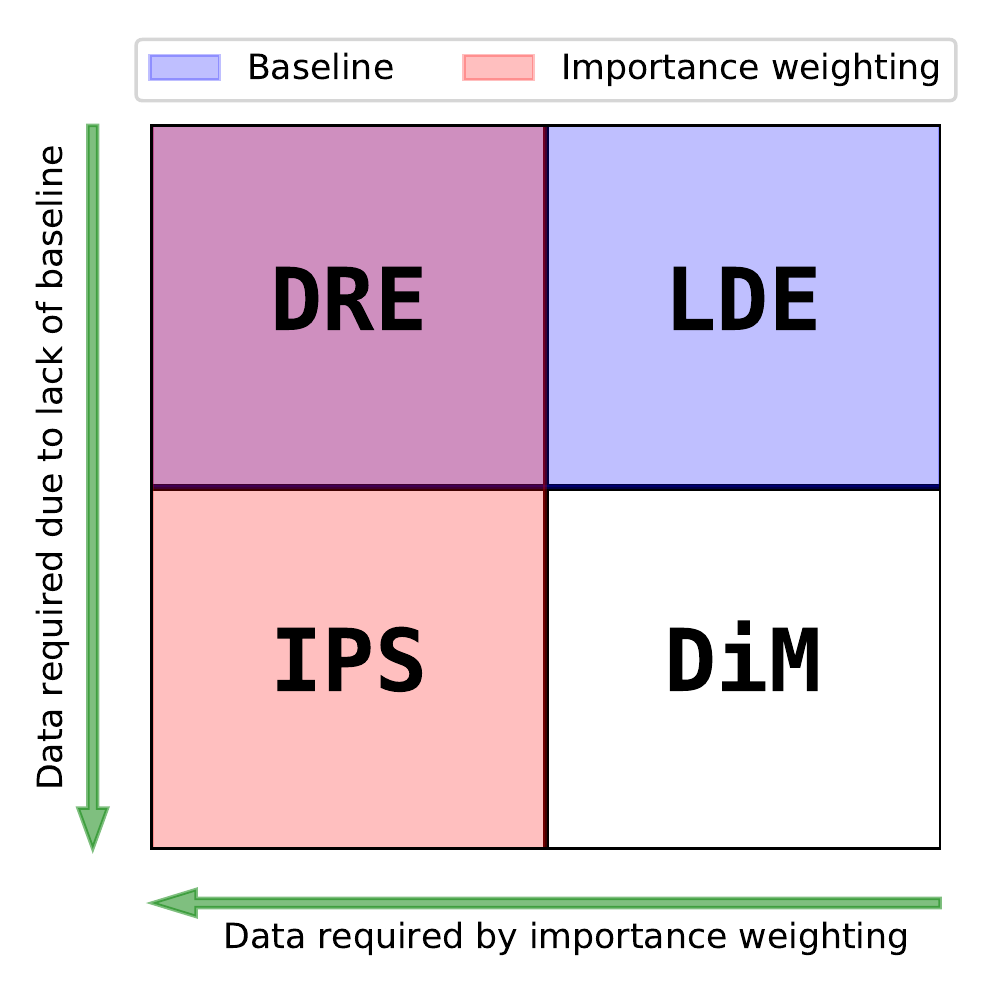}
    \caption{Spectrum of policy evaluation methods based on their use of a baseline and importance weighting.}
    \label{fig:pem_spectrum}
\end{wrapfigure}

Unlike the DRE and the IPS, the LDE does not utilize the importance weighting and thus may be affected by an irregular distribution of the historical data.
However, the LDE is intended to be used in a limited data setting where the amount of the available logged interactions is too small to discern the distribution density of the observed rewards over the state-action space. 
Given the abundance of the policy evaluation methods, it is natural to question which method is best in a given scenario.
From a design perspective, the policy evaluation methods discussed above can be categorized by their use of two mechanisms: baseline and importance weighting.
Based on the utilization of these mechanisms, one may speculate on the amount of historical data required for a policy evaluation method to be accurate.
Intuitively, the use of baseline allows for a smaller number of interactions since unobserved rewards will be approximated, while the use of importance weighting requires a sufficient amount of data so that the irregular distribution density of the behavioral policy can be leveraged, see e.g.~\citep{doroudi2018importance}.
A schematic representation of the policy evaluation methods based on these factors is presented in Figure~\ref{fig:pem_spectrum}.

\subsection{Theoretical Results}\label{sec:theory}
In this section we analyze the performance of the LDE on a general policy comparison task.
Since the comparison on a set of policies is performed pair-wise, see Section~\ref{sec:policy_comparison}, we consider the case of two target policies.
Specifically, let $\pi$ and $\mu$ be the target policies, and let $\mathcal{D}$ be the set of historical data.

Throughout this section we assume that the action space $\mathcal{A}$ is discrete and finite.
Although formally the LDE does not depend on a behavioral policy, in order to conduct a statistical analysis, we need to know the distribution of the historical actions.
To this end, we make an assumption of uniform distribution of the historical actions from the observed states:
\begin{equation}\label{eq:hist_assumption}
	\big\{ a \in \mathcal{A} \ |\ \big( s, a, r \big) \in \mathcal{D} \big\}
	\sim \mathcal{U} \big( 2^\mathcal{A} \big)
	\ \ \text{for any observed state}\ \ 
	s \in \mathcal{S}.
\end{equation}
Such an assumption is not restrictive in the case of limited historical data and is generally satisfied under a 
sufficiently exploratory behavioral policy.
We note that assumption~\eqref{eq:hist_assumption} does not imply that the behavioral policy is uniformly random, but rather suggests that the action space will be observed equally over the independent realizations of the historical dataset $\mathcal{D}$.

Even though LDE is intended for the policy comparison tasks, it is important to note that it is an unbiased estimator of the value of the policy as the following result states.
\begin{theorem}\label{thm:unbiasedness}
Let $\pi$ be the target policy and $\mathcal{D}$ be the historical data satisfying assumption~\eqref{eq:hist_assumption}.
Then the LDE is unbiased, i.e.
\[
    \mathbb{E}_\mathcal{D} \big[ V_{LDE}(\pi;\mathcal{D}) \big] = V(\pi).
\]
\end{theorem}

Our main result, stated below, shows that with high confidence the policy comparison performed by the LDE is correct.
\begin{theorem}\label{thm:policy_comp}
Let $\pi$ and $\mu$ be the target policies and $\mathcal{D}$ be the historical data satisfying assumption~\eqref{eq:hist_assumption}.
Then the probability $p_{comp}$ that the LDE with the baseline $\hat{r}$ provides the correct comparison between $\pi$ and $\mu$ admits the following estimate
\[
    p_{comp} = p_{comp}(\pi, \mu; \hat{r}, \mathcal{D})
    \ge 1 - \exp\Bigg( -\frac{|\mathcal{D}|^2 \big(V(\pi) - V(\mu)\big)^2}{8 |\mathcal{A}|^4 R^2} \Bigg)
\]
where $R := \max\{ \max\mathcal{R} - \mathbb{E}_s[\hat{r}(s)],\ \mathbb{E}_s[\hat{r}(s)] - \min\mathcal{R} \} < \infty$.
\end{theorem}

The above result highlights three main components to the success of the LDE for the policy comparison task: the size of the historical dataset $\mathcal{D}$, the difference of the (unknown) values of the target policies $\pi$ and $\mu$, and the baseline $\hat{r}$ selection, as it directly relates to the value of $R$.
Evidently, It is easier to compare two policies when there are more historical interactions available and when the policies have a larger difference in value.
Somewhat surprisingly, the choice of baseline does not seem to have much importance on the task of policy comparison, as it only changes the value of $R$, even though it is a crucial factor for policy evaluation.
In Section~\ref{sec:ex1} we numerically affirm the importance of these factors on synthetic data environments.

In special cases, our main result infers theoretical guarantees for the policy evaluation methods from Section~\ref{sec:pem} as stated below.
\begin{remark}
The statement of Theorem~\ref{thm:policy_comp} holds for
\begin{itemize}
    \item the DiM with $R = \max\big\{ \max\mathcal{R},\ -\min\mathcal{R} \big\} < \infty$;
    \item the IPS with $R = \max\big\{ \max\mathcal{R},\ -\min\mathcal{R} \big\} < \infty$ if the historical data $\mathcal{D}$ is collected under the uniformly random behavioral policy;
    \item the DRE if the historical data $\mathcal{D}$ is collected under the uniformly random behavioral policy.
\end{itemize}
\end{remark}
Additional theoretical results, proofs, and related discussion can be found in Section~\ref{sec:theory_appendix} in the Appendix.
Next, we confirm our theoretical findings in various practical settings.

\section{Numerical Experiments}\label{sec:numerics}
In this section we assess the performance of different policy evaluation methods on the task of policy comparison, as described in Section~\ref{sec:policy_comparison}.
The performance of a method on a particular policy comparison task is measured in the following way: given a set of target policies $\pi_1, \ldots, \pi_K$, the score of the method is the normalized number of correct pair-wise comparisons, i.e.
\begin{equation}\label{eq:comp_score}
    \text{score} = 1 - \frac{\text{number of incorrect pair-wise comparisons}}{\text{total number of pair-wise comparisons}}
    \in [0,1].
\end{equation}

In our numerical examples we consider the limited data setting, which does not allow for a reliable approximation of the reward on unobserved state-action pairs.
In order to avoid an unintentional bias in the baseline selection, we select baseline $\hat{r}$ to be the constant function derived as the average of the observed rewards from the historical data $\mathcal{D} = \{ (s_n, a_n, r_n, p_n) \}_{n=1}^N$, i.e.
\[
    \hat{r}(s,a) = \hat{r} = \frac{1}{N} \sum_{n=1}^N r_n.
\]
Generally, a more precise baseline provides better performance, and some of the common choices are discussed in Section~\ref{sec:baseline_appendix} in the Appendix.
However, in this paper we assume no additional knowledge of the environment other than the historical dataset $\mathcal{D}$, hence our modest choice of the baseline.

In this case the policy evaluation methods take the form
\begin{align*}
    V_{DiM}(\pi) &= \frac{1}{N} \sum_{n=1}^N r_n \, \pi(a_n|s_n)
    &
    V_{LDE}(\pi) &= \frac{1}{N} \sum_{n=1}^N (r_n - \hat{r}) \pi(a_n|s_n) + \hat{r}
    \\
    V_{IPS}(\pi) &= \frac{1}{N} \sum_{n=1}^N r_n \frac{\pi(a_n|s_n)}{p_n}
    &
    V_{DRE}(\pi) &= \frac{1}{N} \sum_{n=1}^N (r_n - \hat{r}) \frac{\pi(a_n|s_n)}{p_n} + \hat{r}
\end{align*}
with the available historical data $\mathcal{D} = \{ (s_n, a_n, r_n, p_n) \}_{n=1}^N$ and the corresponding baseline $\hat{r}$. 

The source code reproducing the presented experiments is available at~\url{https://github.com/joedaws/lde2021}.

\subsection{Example 1: Synthetic Data}\label{sec:ex1}
In this example we perform a policy comparison task on a synthetic contextual bandit.
According to our theoretical analysis in Section~\ref{sec:theory}, the complexity of the policy comparison task depends on the size of the historical dataset $\mathcal{D}$ and the difference of values of the target policies $\pi$ and $\mu$.
In order to highlight the effect of these parameters, we define by $m$ the number of sampled historical actions in $\mathcal{D}$ for each state and by $\alpha$ the normalized difference of values of $\pi$ and $\mu$, i.e.:
\begin{equation}\label{eq:value_diff}
    \alpha = \alpha(\pi, \mu) = \frac{V(\pi) - V(\mu)}{V_{max} - V_{min}} \in [0,1]
\end{equation}
where $V_{max}$ and $V_{min}$ are the values of the optimal and the pessimal policies respectively.

A single test consists of randomly generating target policies $\pi$ and $\mu$ (with a specified $\alpha$), sampling historical data $\mathcal{D}$ (with a specified $m$), and performing policy evaluation and comparison.
The reported results are obtained by repeating tests for 1,000 times for each pair of values of $\alpha$ and $m$.
Detailed information on the environment setup is provided in Section~\ref{sec:ex1_appendix} in the Appendix.

\begin{figure}[hbt!]
    \centering
    \includegraphics[width=.49\linewidth]{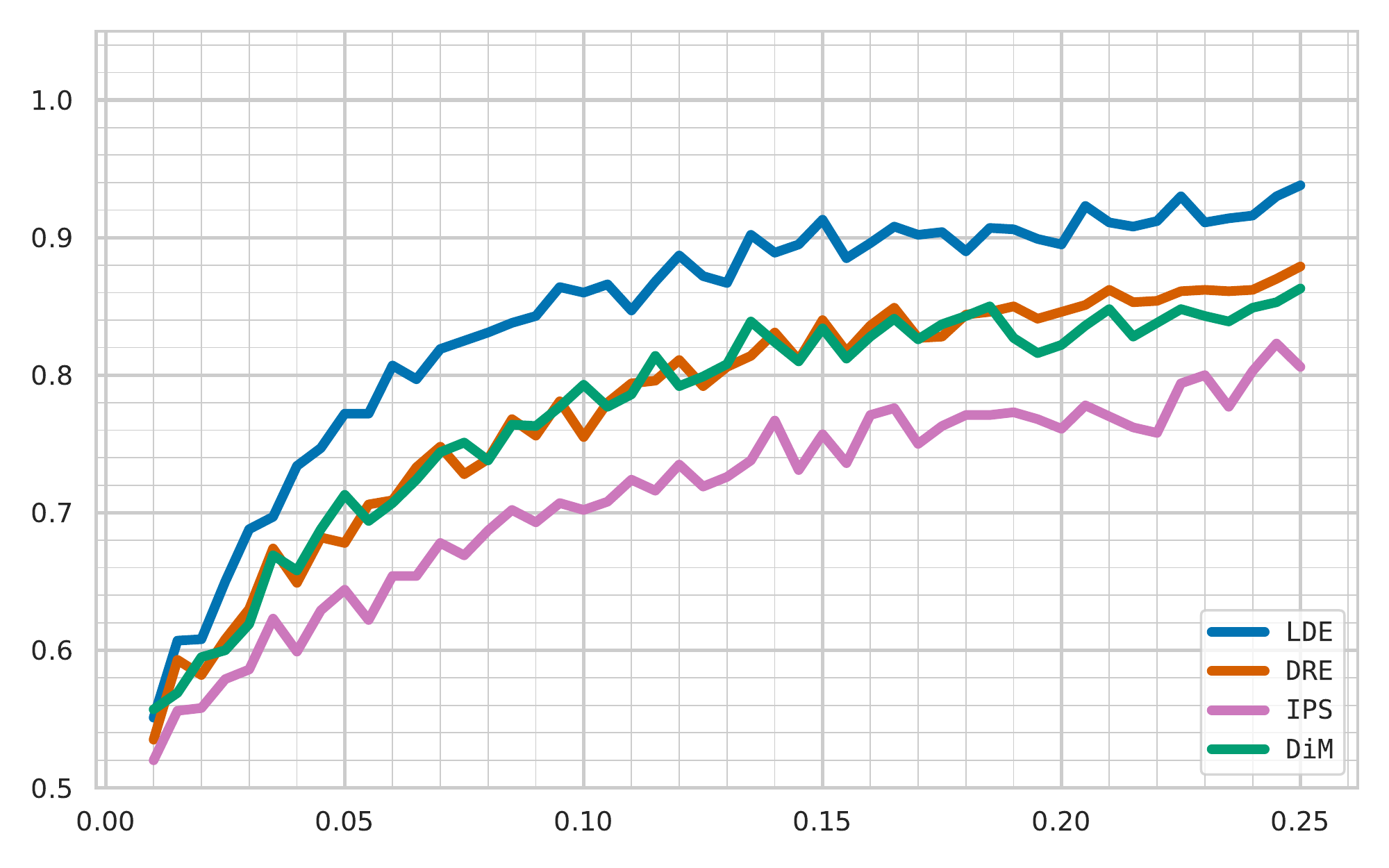}
    \includegraphics[width=.49\linewidth]{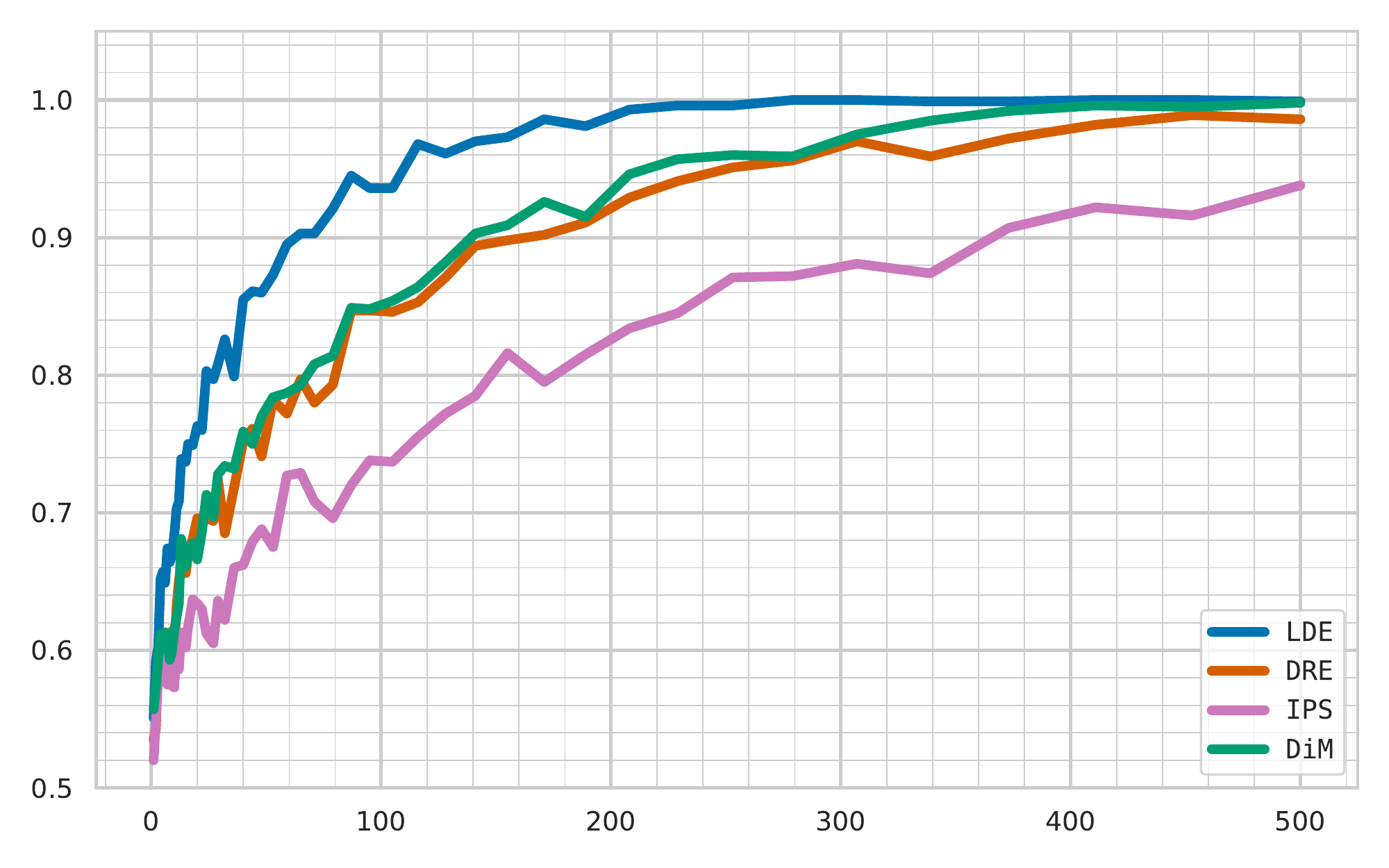}
    \caption{Policy comparison rate depending on the value difference $\alpha$ (left, $m = 1$) or the amount of historical data $m$ (right, $\alpha = .01$).}
    \label{fig:ex1.1_comp}
\end{figure}

\paragraph{Example 1.1}
Consider the reward function given by
\begin{equation}\label{eq:ex1.1_reward}
    r(s,a) = \Big( \exp(s + a) + \sin \big( 2\pi (s - a) \big) \Big)
        \times \Big( \cos \big( 2\pi (s + a) \big) + \exp(s - a) \Big).
\end{equation}
The policy comparison rates are presented in Figure~\ref{fig:ex1.1_comp} and the policy evaluation errors are given in Tables~\ref{tab:ex1.1_eval_a} and~\ref{tab:ex1.1_eval_m}, along with additional results, are provided in Section~\ref{sec:ex1.1_appendix} in the Appendix.

\begin{figure}[hbt!]
    \centering
    \includegraphics[width=.49\linewidth]{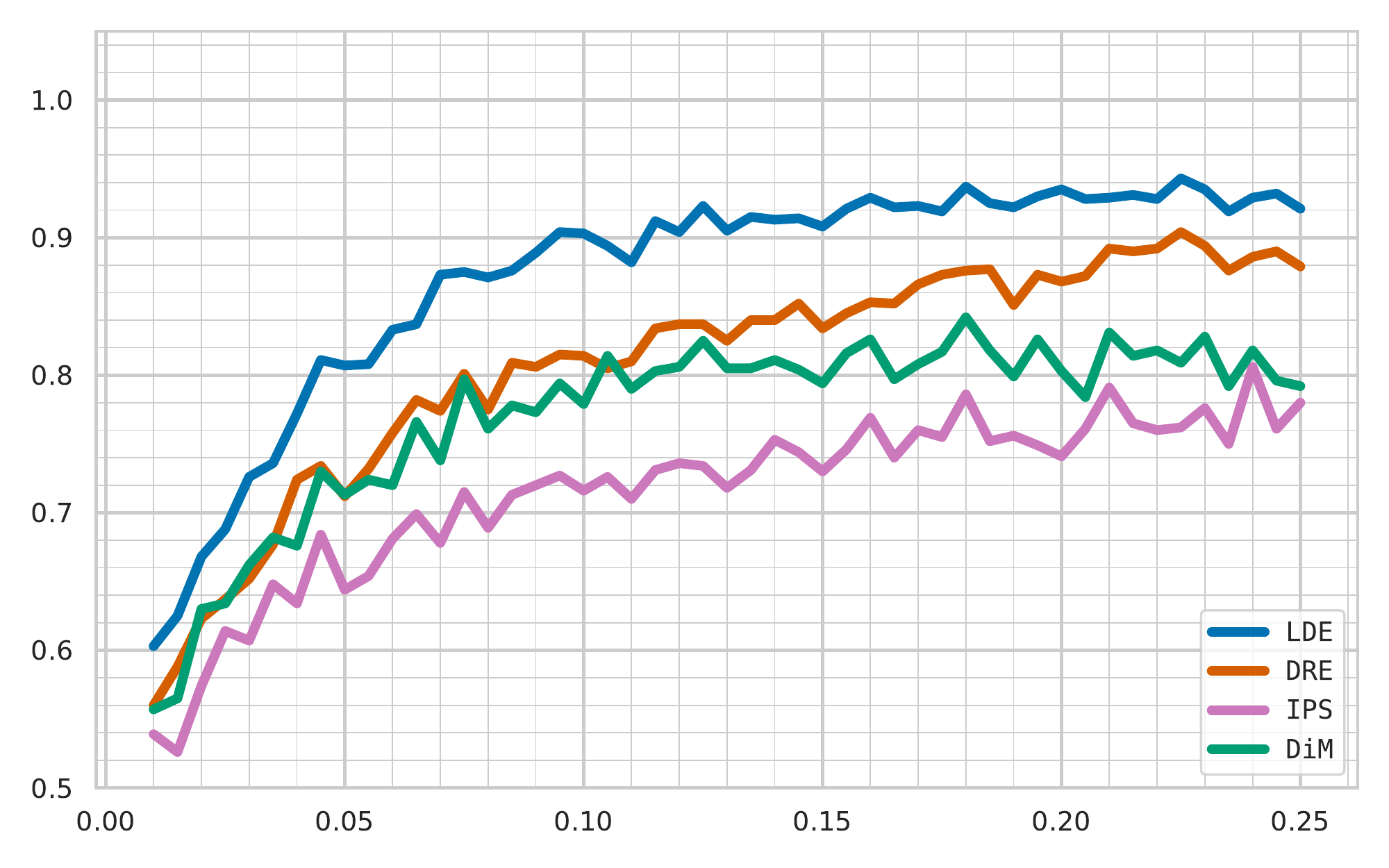}
    \includegraphics[width=.49\linewidth]{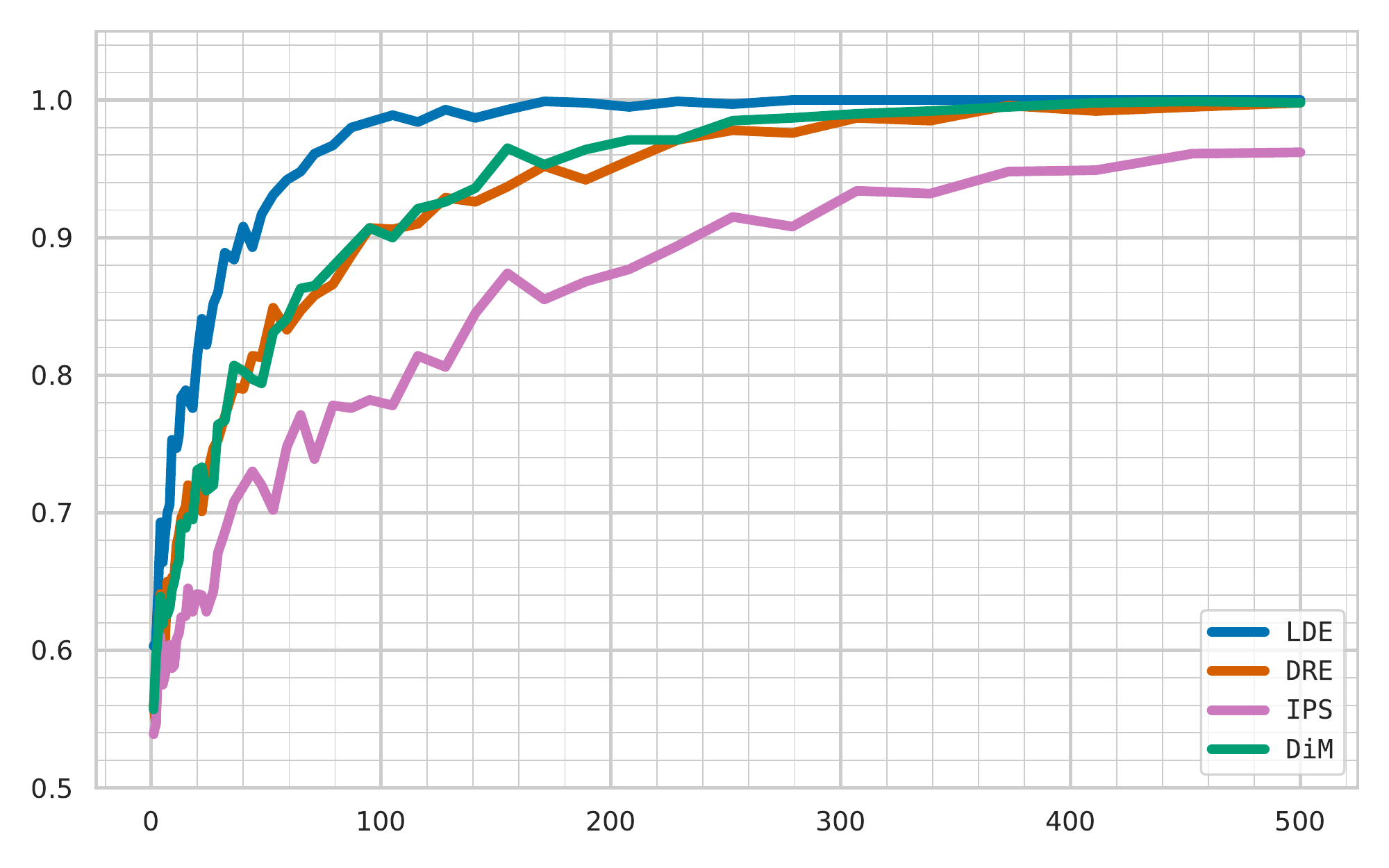}
    \caption{Policy comparison rate depending on the value difference $\alpha$ (left, $m = 1$) or the amount of historical data $m$ (right, $\alpha = .01$).}
    \label{fig:ex1.2_comp}
\end{figure}

\paragraph{Example 1.2}\label{sec:ex1.2}
Consider the reward function given by
\begin{equation}\label{eq:ex1.2_reward}
	r(s,a) = \frac{3 + \cos \big( 2\pi (s + a) \big)
		+ \cos \big( 10\pi (s - a) \big)}{2} \ \exp \Big( -\big( a + \cos(2\pi s) \big)^2 \Big).
\end{equation}
The policy comparison rates are presented in Figure~\ref{fig:ex1.2_comp} and the policy evaluation errors are given in Tables~\ref{tab:ex1.2_eval_a} and~\ref{tab:ex1.2_eval_m}.
Additional details and results are provided in Section~\ref{sec:ex1.2_appendix} in the Appendix.

\subsection{Example 2: Classification}\label{sec:ex2}
In~\citep{dudik2014doubly} the authors consider a policy evaluation task on a set of classifiers obtained via supervised learning.
Motivated by their example, we consider a closely related formulation of the classification problem and reframe it as a policy comparison task. 
Specifically, in this example we employ the policy evaluation methods to compare a set of $10$ target policies, trained via the Direct Loss Minimization~\citep{mcallester2010direct} on several datasets selected from~\citep{asuncion2007uci}.
The average of the test scores, computed by~\eqref{eq:comp_score}, with error bars over 1,000 tests 
is presented in Figure~\ref{fig:ex2_comp}.
Additional information on this example can be found in Section~\ref{sec:ex2_appendix} in the Appendix.
\begin{figure}[hbt!]
    \centering
    \includegraphics[width=\linewidth]{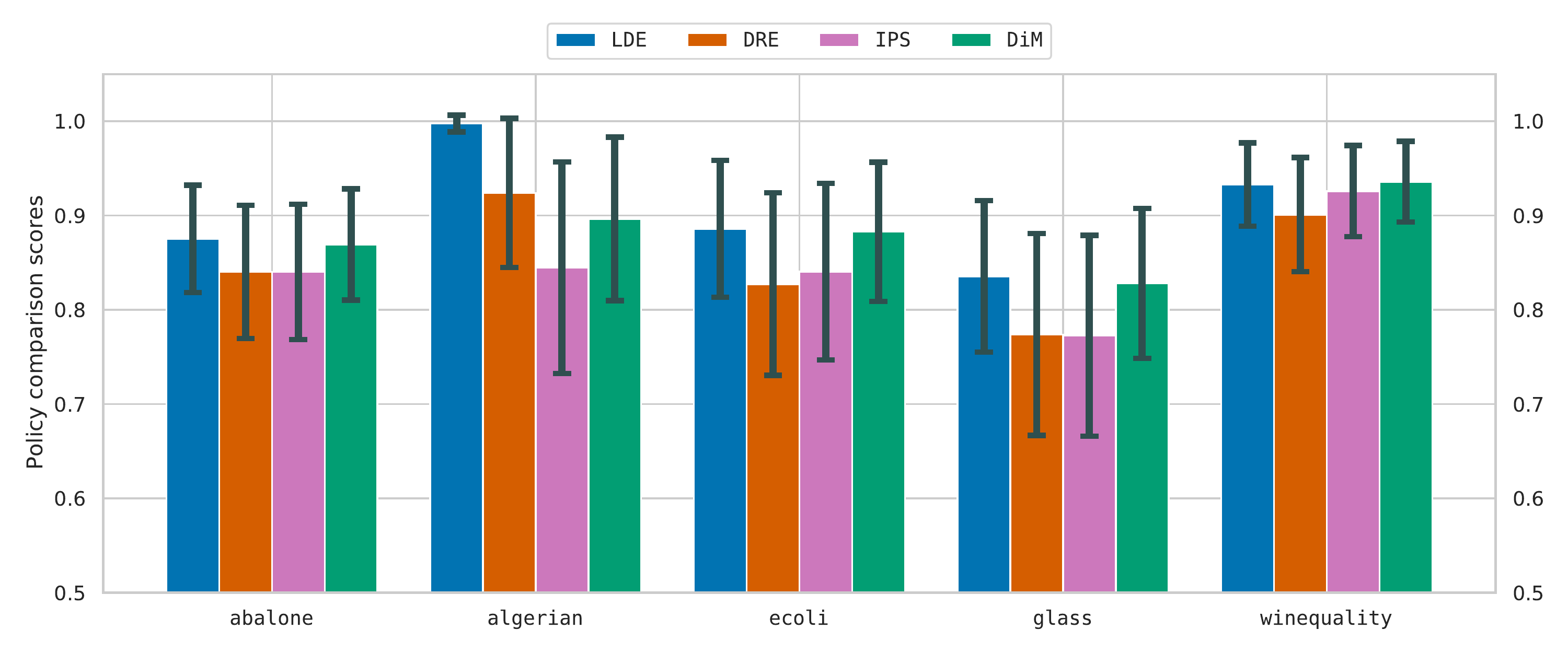}
    \caption{Policy comparison scores on various datasets.}
    \label{fig:ex2_comp}
\end{figure}

\subsection{Example 3: RL Environments}\label{sec:ex3}
\begin{figure}[hbt!]
    \centering
    \includegraphics[width=.49\linewidth]{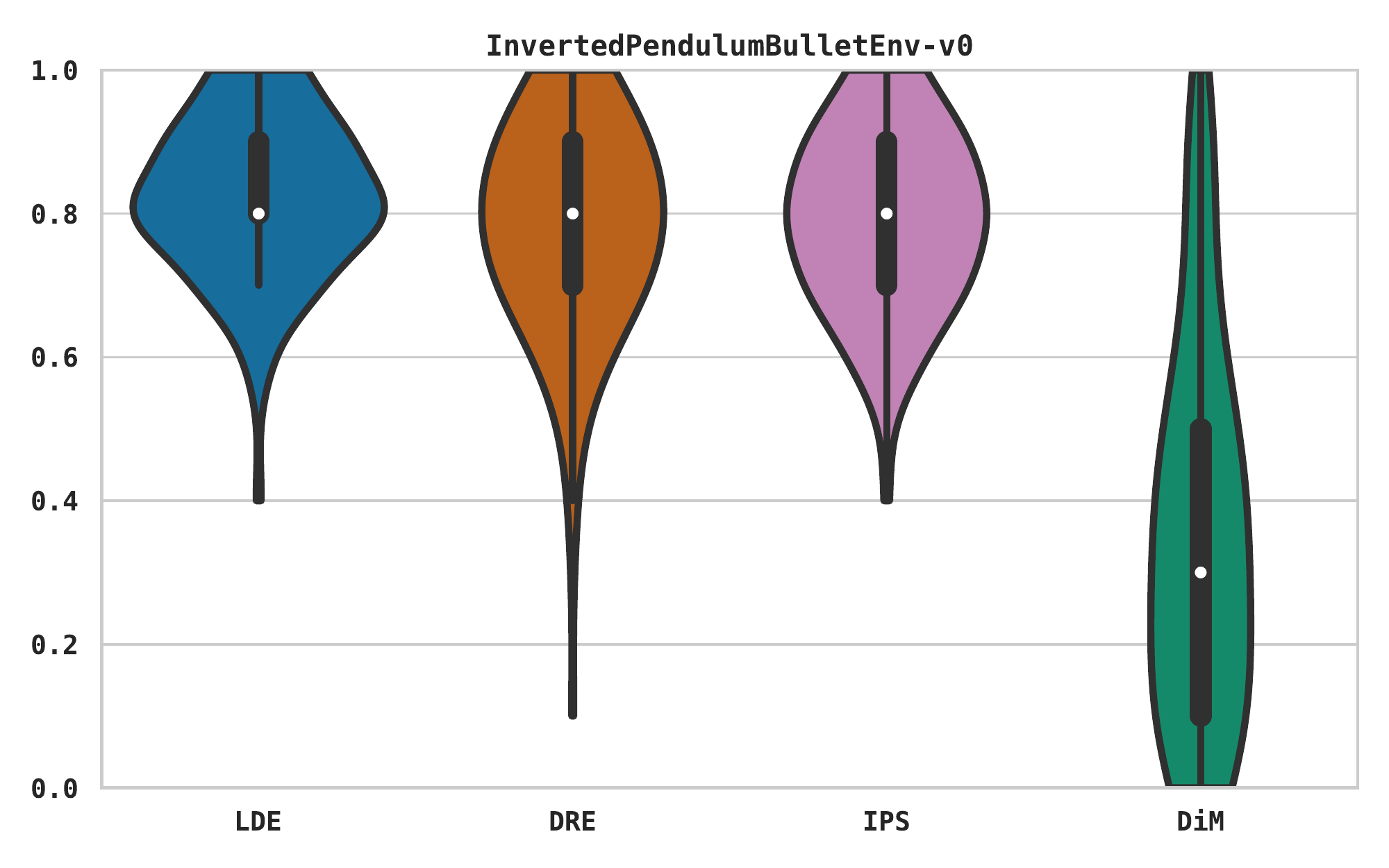}
    \includegraphics[width=.49\linewidth]{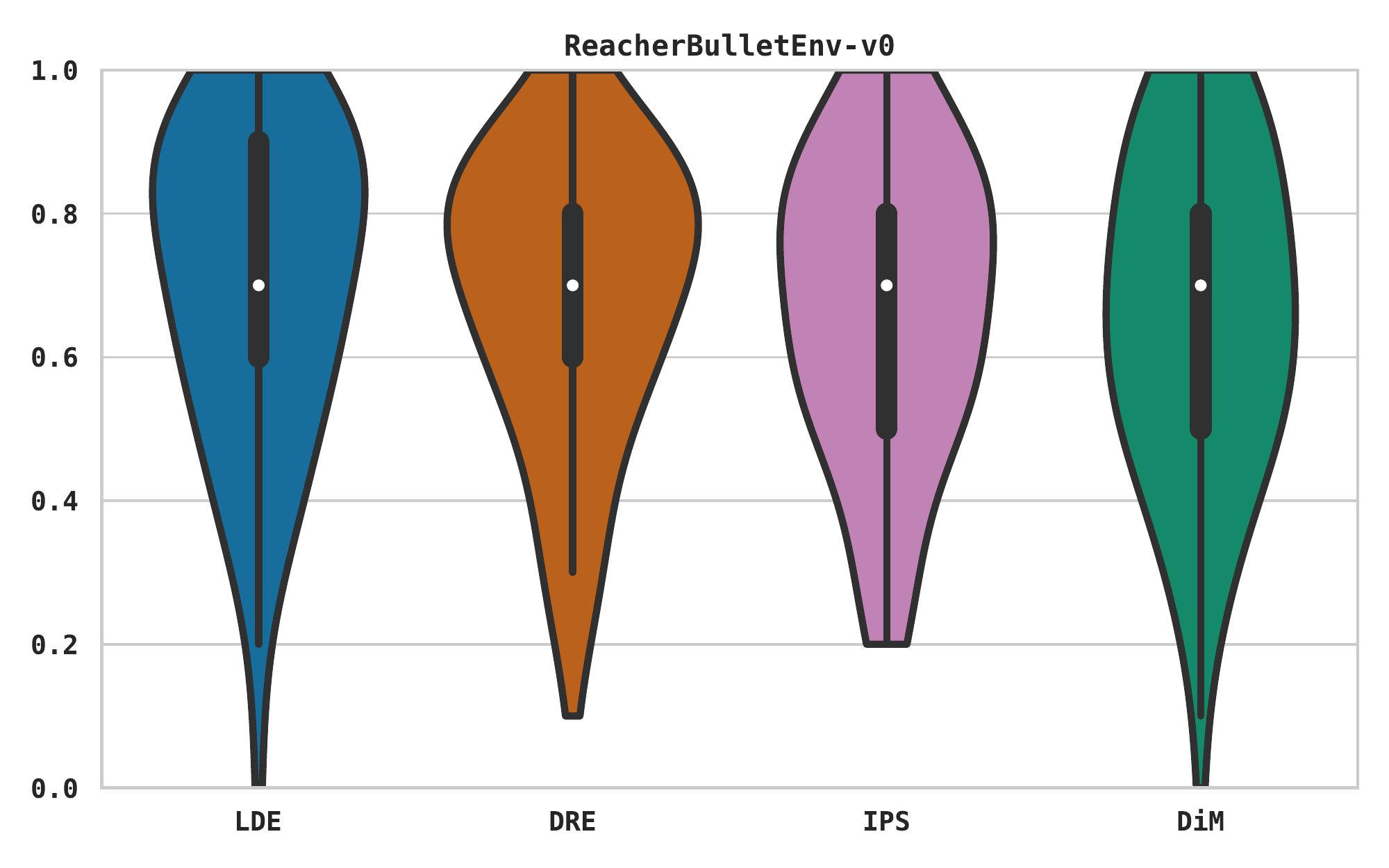}
    \\
    \includegraphics[width=.49\linewidth]{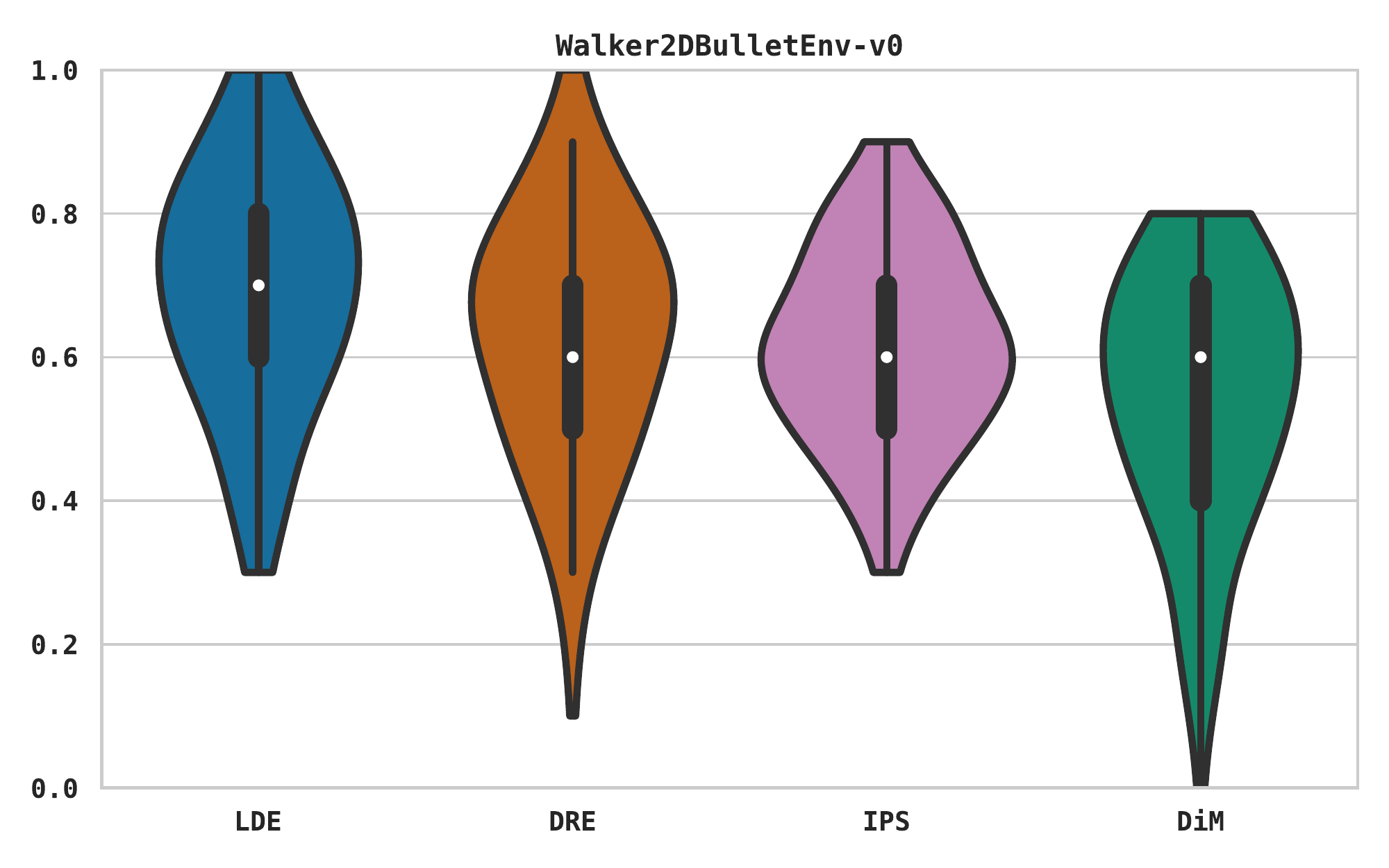}
    \includegraphics[width=.49\linewidth]{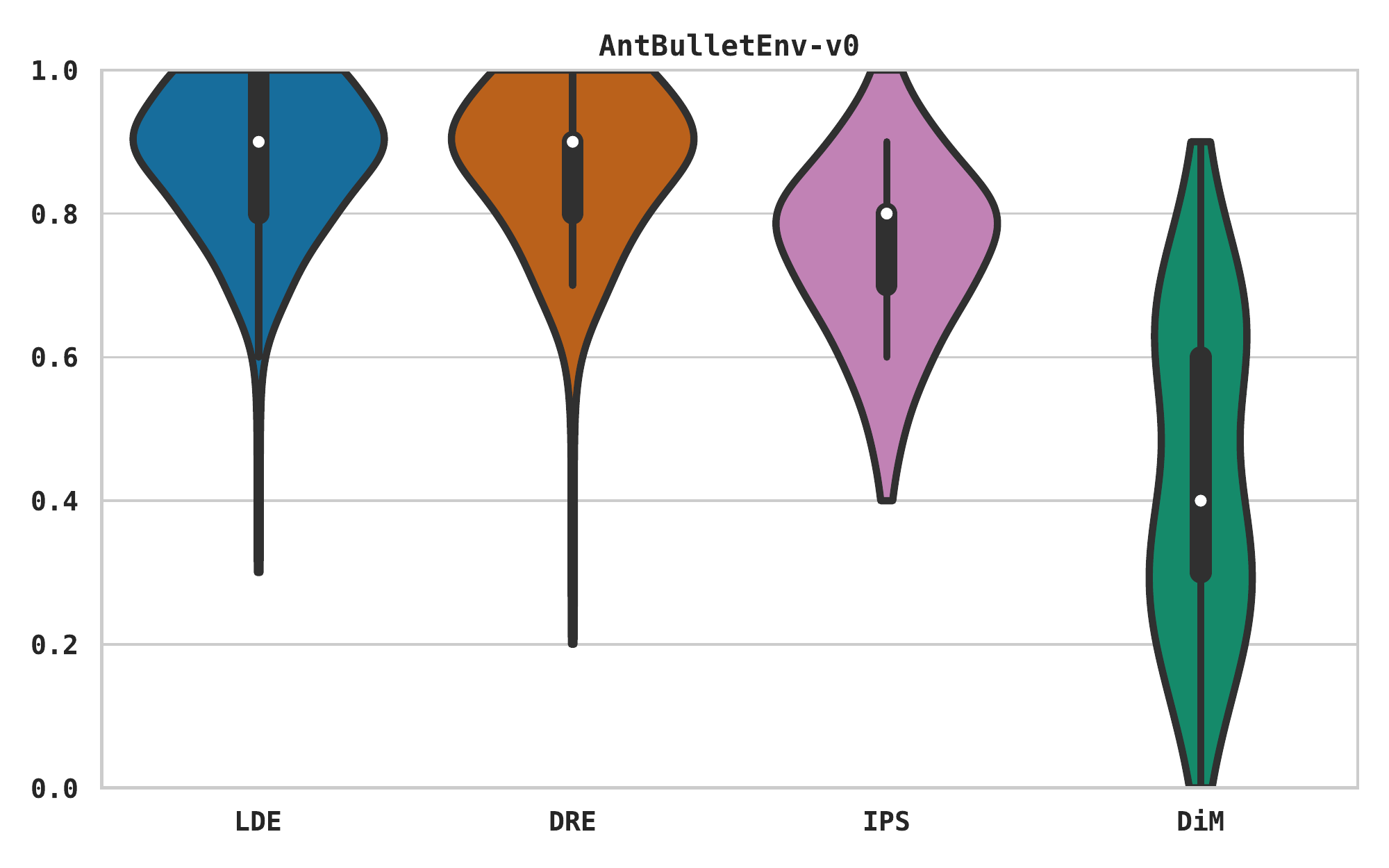}
    \caption{Policy comparison on PyBullet RL environments, see Table~\ref{tab:ex3}.}
    \label{fig:ex3_comp}
\end{figure}

In this example we perform a policy comparison task on conventional reinforcement learning environments from the \texttt{PyBullet}\footnote{\url{https://pybullet.org/}} engine.
Specifically, we use the \texttt{Stable-Baselines3}\footnote{\url{https://stable-baselines3.readthedocs.io/}}~\citep{stable-baselines3} library to train $5$ target policies using different optimization algorithms: \texttt{A2C}~\citep{mnih2016asynchronous}, \texttt{DDPG}~\citep{silver2014deterministic, lillicrap2015continuous}, \texttt{PPO}~\citep{schulman2017proximal}, \texttt{SAC}~\citep{haarnoja2018soft}, and \texttt{TD3}~\citep{fujimoto2018addressing}.
Then we deploy the policy evaluation methods to compare the target policies on the limited historical data $\mathcal{D}$.
The statistical distribution of the resulting scores, computed via~\eqref{eq:comp_score}, over 225 tests is presented in Table~\ref{tab:ex3} and in Figure~\ref{fig:ex3_comp}.
Performance on additional environments and an exact description of the experiment's setup can be found in Section~\ref{sec:ex3_appendix} in the Appendix.
\begin{table}[hbt!]
    \centering\small
    \caption{Policy comparison scores on PyBullet RL Environments: average (std).}
    \label{tab:ex3}
    \begin{tabular}{lcccc}
        \toprule
        Environment & LDE & DRE & IPS & DiM
        \\\midrule
        \texttt{InvertedPendulum-v0} & \textbf{0.8249} (0.109) & 0.7831 (0.146) & 0.7929 (0.124) & 0.3551 (0.258)
        \\
        \texttt{Reacher-v0} & \textbf{0.7076} (0.218) & 0.6818 (0.196) & 0.6693 (0.205) & 0.6582 (0.224)
        \\
        \texttt{Walker2D-v0} & \textbf{0.6969} (0.166) & 0.6289 (0.168) & 0.6160 (0.132) & 0.5458 (0.176)
        \\
        \texttt{Ant-v0} & \textbf{0.8760} (0.110) & 0.8680 (0.121) & 0.7498 (0.120) & 0.4400 (0.219)
        \\\bottomrule
    \end{tabular}
\end{table}

\section{Conclusion}
In this work we address the challenge of reinforcement learning systems being deployed in real-world scenarios where only a minute amount of logged data is available.
Such a setting is natural for many critical applications like healthcare, recommender systems, and hyper-personalization, where the recorded data is extremely sparse and often contains sensitive information.
We denote such scenarios as \textit{limited data} settings, where the conventional policy evaluation is not permissive due to the data constraints.
To treat such cases we propose performing the \textit{policy comparison} task, which we introduce and formalize.

We introduce the \textit{Limited Data Estimator} (LDE), which is suitable for policy comparison tasks in such challenging settings, and compare it to the known methods for policy evaluation.
Our theoretical analysis proves that with high confidence the LDE correctly compares target policies even when only a small number of agent-environment interactions are available.
Our numerical results demonstrate that the LDE typically surpasses other methods on the policy comparison task in various settings.
We intend the LDE to be useful in practical applications where conventional approaches might produce misleading results.

\subsection*{Ethical Considerations}\label{sec:ethics}
When considering which datasets and test environments to employ, we specifically sought out publicly available and widely-used options.
In our numerical examples in Section~\ref{sec:numerics}, the data is either synthetic or from commonly used open sources, namely the UCI dataset repository~\citep{asuncion2007uci} and the \texttt{PyBullet} (\url{https://pybullet.org/}) project.
Our synthetic data can be recreated by using the description of the experiment or by using the source code provided at~\url{https://github.com/joedaws/lde2021}.
While there is no inherent property of this work which increases risk of misuse, the results of this work may be used to find improved solutions to a wide range of contextual bandit and reinforcement learning problems.
As there are instances of machine learning systems having a negative impact on human life and on the environment, our method poses an implicit risk of assisting unintentionally unethical applications.
However, the risk of our methodology being used nefariously is no greater than that of any other machine learning algorithms employed in the real-world.

{\small\bibliography{references}}

\newpage
\appendix
\begin{center}
    \textbf{\Large Appendix}
    \bigskip
\end{center}
\section{Baseline Selection}\label{sec:baseline_appendix}
The value estimate provided by the Limited Data Estimator, defined in Section~\ref{sec:lde}, is affected by the selected baseline function $\hat{r} : \mathcal{S} \times \mathcal{A} \to \mathbb{R}$.
Generally the choice of baseline is determined by the domain knowledge and such problem-instance particulars as the availability of the environment model or the volume of the recorded interactions.
Depending on the available information, the following are some of the common use-cases that can be deployed in practice:
\begin{itemize}
    \item if a sufficiently reliable environment model is available, e.g.~\citep{chua2018deep}, then the modelled Q-function can be used as a baseline, i.e. $\hat{r}(s,a) = Q_{model}(s,a)$;
    \item if an actor-critic algorithm is deployed, e.g.~\citep{grondman2012survey}, then the value component of the critic can be used as a baseline, i.e. $\hat{r}(s,a) = V_{critic}(\nu|s)$, where $\nu$ is an actively deployed policy which interacts with the environment during the training;
    \item if sufficiently many historical interactions are recorded in $\mathcal{D}$, it might be possible to obtain an approximation $\tilde{r}(s,a;\mathcal{D})$ of the reward function $r(s,a)$, e.g. by a supervised learning algorithm~\citep{le2019batch}, and used as a baseline, i.e. $\hat{r}(s,a) = \tilde{r}(s,a;\mathcal{D})$;
    \item if a moderate amount of historical interactions in $\mathcal{D}$ are available but not sufficient for an accurate reward approximation, one may settle for a simple linear model $\bar{r}(s,a;\mathcal{D})$, e.g.~\citep{dudik2011doubly}, which, even though not being a precise estimate of the reward function, keeps the relation between the neighboring state-action pairs and does not overfit the historical data, i.e. $\hat{r}(s,a) = \bar{r}(s,a;\mathcal{D})$;
    \item if only a small number of historical data is observed in $\mathcal{D}$, one may use an empirical average as a crude estimate of the reward, i.e. $\hat{r}(s,a) = (r_1 + \ldots + r_N) / N$, which is what we use in our numerical experiments in Section~\ref{sec:numerics};
    \item in more complicated cases when the data is severely restricted or the distribution of the observed rewards seems unsuitable for extrapolation, one may heuristically select a fixed number from the reward space $\mathcal{R}$ to serve as a constant baseline, i.e. $\hat{r}(s,a) = \hat{r} \in \mathcal{R}$.
\end{itemize}
In this effort we address a general limited data setting where it is usually not possible to obtain a suitable approximation of the reward function; thus in our numerical results we consider the constant baseline given by the averaging of the observed rewards.
However, our theoretical analysis holds for a wider class of baseline functions with the same statements and minor adjustments to the proofs.

\newpage
\section{Theoretical Analysis}\label{sec:theory_appendix}
In this section we theoretically analyze the performance of the LDE on a general policy evaluation and comparison tasks.
We consider a contextual bandit environment $(\mathcal{S, A, R}, r)$ with a state pace $\mathcal{S}$, a discrete action space $\mathcal{A}$ of cardinality $n \in \mathbb{N}$, a bounded reward space $\mathcal{R} \subset \mathbb{R}$, and a deterministic reward function $r : \mathcal{S} \times \mathcal{A} \to \mathcal{R}$, i.e.
\[
	\mathcal{A} = \{a_1, a_2, \ldots, a_n\}
	\quad\text{and}\quad
	-\infty < \min\mathcal{R} < \max\mathcal{R} < \infty.
\]

We first address the problem of the policy evaluation: given a policy $\pi$ and a set of historical data $\mathcal{D}$, see~\eqref{eq:historical_data}, estimate the value $V(\pi)$ of the policy $\pi$, defined in~\eqref{eq:policy_value}.
Even though, due to its simplicity, the LDE is generally not well suited for the task of evaluation, we show that under a mild assumption on the distribution of the historical actions~\eqref{eq:hist_assumption} the LDE is unbiased, as Theorem~\ref{thm:unbiasedness} states.
The proof of this result is provided in Section~\ref{sec:unbiasedness_proof}.

Next, the policy comparison problem can be viewed as a generalization of the evaluation task and is stated as follows: given target policies $\pi$, $\mu$ and a set of historical data $\mathcal{D}$, predict which policy has the higher value, i.e. compare the values $V(\pi)$ and $V(\mu)$.
The rationality of such a task is justifiable in settings where the amount of available historical data $\mathcal{D}$ is not sufficient for conventional policy evaluation, as we discuss in Section~\ref{sec:limited_data}, since the policy comparison task is often feasible even in cases of severely restricted data.
The performance of the LDE on the policy comparison task is explained by Theorem~\ref{thm:policy_comp}, which states that under assumption~\eqref{eq:hist_assumption} the comparison of the target policies provided by the LDE is correct with high confidence.
The proof of this result is given in Section~\ref{sec:policy_comp_proof}.

In addition to the policy value comparison, we analyze the performance of the LDE on the task of comparing policies on individual states.
The rationale behind such an approach is the following: in order to make a policy comparison, one first has to compute a value surrogate, which includes a sum over the historical dataset $\mathcal{D}$.
Hence the whole comparison might be false if a particular term in this sum significantly diverges from its true value.

Specifically, given the target policies $\pi$ and $\mu$, the comparison on a state $s \in \mathcal{S}$ is determined by the sign of the difference of the LDE estimates $V_{LDE}(\pi|s) - V_{LDE}(\mu|s)$, as compared to the difference of the true estimates $V(\pi|s) - V(\mu|s)$.
The comparison on the state $s$ is correct if the signs of those differences are the same, and is incorrect otherwise.
Note that it is not realistic to expect correct predictions on every state $s \in \mathcal{S}$ and an incorrect comparison on a single state does not imply the wrong outcome on the policy comparison task.
In practice we prefer to avoid the per-state comparisons being ``very incorrect'', which we quantify by the magnitude of the product of the value differences $(V_{LDE}(\pi|s) - V_{LDE}(\mu|s)) \times (V(\pi|s) - V(\mu|s))$.
To formalize, for a \textit{misprediction threshold} $\delta > 0$ we define the probability $p(\delta|s)$, given as
\[
    p(\delta|s) = \mathbb{P}\Big[ \big( V_{LDE}(\pi|s) - V_{LDE}(\mu|s) \big)
		\times \big( V(\pi|s) - V(\mu|s) \big) \le -\delta \Big].
\]
From stability considerations, for large values of $\delta > 0$ we want $p(\delta|s)$ to be small for all states $s \in \mathcal{S}$.
Our next result affirms that with high probability the LDE comparisons on individual states are within the reasonable bounds.

\begin{theorem}\label{thm:state_comp}
Let $\pi$ and $\mu$ be the target policies and $\mathcal{D}$ be the historical date satisfying assumption~\eqref{eq:hist_assumption}.
Then for any state $s \in \mathcal{S}$ and for any misprediction threshold $\delta > 0$ we have
\[
	p(\delta|s) \le \Bigg( 1 + \frac{\delta + \frac{m}{n} \big(V(\pi|s) - V(\mu|s)\big)^2}{2R^2} \Bigg)^{-1}
\]
where
\[
    p(\delta|s) = p(\delta|s;\pi,\mu,\hat{r},\mathcal{D}) := \mathbb{P}\Big[ \big( V_{LDE}(\pi|s) - V_{LDE}(\mu|s) \big)
		\times \big( V(\pi|s) - V(\mu|s) \big) \le -\delta \Big]
\]
and $R := \max\{ \max\mathcal{R} - \hat{r}(s),\ \hat{r}(s) - \min\mathcal{R} \} < \infty$, and $m \le n$ is the number of historical actions taken from state $s$.
\end{theorem}

Theorem~\ref{thm:state_comp} shows that the chance of per-state misprediction of the LDE depends on the magnitude of the misprediction threshold $\delta$, the number $m$ of historical actions taken from the state $s$, and the difference of the values of $\pi$ and $\mu$ at the state $s$.
In Section~\ref{sec:state_comp_proof} we provide the proof of this result and in Section~\ref{sec:ex1_appendix} we numerically compare the performance of per-state predictions of the LDE by comparing it to other methods, see Example~1.1 (Figure~\ref{fig:ex1.1_grid}) and Example~1.2 (Figure~\ref{fig:ex1.2_grid}).

Note that for simplicity of presentation we carry out the analysis and prove our results for the class of baselines that are constant in the action space, i.e. $\hat{r}(s,a) = \hat{r}(s)$ for all $s \in \mathcal{S}$ and $a \in \mathcal{A}$.
However, our analysis works for a wider class of baselines with minor adjustment in the proofs.

\subsection{Proof of Theorem~\ref{thm:unbiasedness} (Unbiasedness)}\label{sec:unbiasedness_proof}
Let $\pi$ be the target policy and let $V(\pi|s)$ be the value of $\pi$ at the state $s \in \mathcal{S}$, given by
\[
    V(\pi|s)
    = \mathbb{E} \big[ r(s,a) \,|\, a \sim \pi(s) \big]
    = \sum_{i=1}^n r(s,a_i) \pi(a_i|s).
\]
Denote by $\mathcal{A}_s \subset \mathcal{A}$ the set of historical actions that were taken from $s$:
\[
	\mathcal{A}_s = \big\{ a \in \mathcal{A} \ |\ \big( s, a, r(s,a) \big) \in \mathcal{D} \big\}.
\]
Then we have
\[
    V_{LDE}(\pi|s) = \frac{1}{|\mathcal{A}_s|} \sum_{a \in \mathcal{A}_s} \big( r(s,a) - \hat{r}(s,a) \big) \pi(a|s) + \sum_{i=1}^n \hat{r}(s,a_i) \pi(a_i|s)
\]
and for any $a \in \mathcal{A}_s$ assumption~\eqref{eq:hist_assumption} provides
\begin{align*}
    \mathbb{E} \big[ \big( r(s,a) - \hat{r}(s,a) \big) \pi(a|s) \,\big|\, a \sim \mathcal{U(A)} \big]
    &= \frac{1}{n} \sum_{i=1}^n (r(s,a_i) - \hat{r}(s,a_i)) \pi(a_i|s)
    \\
    &= V(\pi|s) - \sum_{i=1}^n \hat{r}(s,a_i) \pi(a_i|s).
\end{align*}
Thus we deduce
\[
    \mathbb{E} \big[ V_{LDE}(\pi|s) \,\big|\, \mathcal{A}_s \sim \mathcal{U}(2^\mathcal{A}) \big]
    = V(\pi|s)
\]
and
\[
    \mathbb{E} \big[ V_{LDE}(\pi) \,\big|\, \mathcal{A}_s \sim \mathcal{U}(2^\mathcal{A}) \big]
    = \mathbb{E}_s \Big[ \mathbb{E} \big[ V_{LDE}(\pi|s) \,\big|\, \mathcal{A}_s \sim \mathcal{U}(2^\mathcal{A}) \big] \Big]
    = \mathbb{E}_s \big[ V(\pi|s) \big]
    = V(\pi),
\]
which concludes the proof of Theorem~\ref{thm:unbiasedness}.

\subsection{Proof of Theorem~\ref{thm:state_comp} (Per-state Comparison)}\label{sec:state_comp_proof}
In this section we analyze the comparison of policies $\pi$ and $\mu$ on a state $s \in \mathcal{S}$ by relating the difference of the LDE estimates with the difference of values at the state $s$:
\[
	V_{LDE}(\pi|s) - V_{LDE}(\mu|s) \ \sim\ V(\pi|s) - V(\mu|s).
\]
Specifically, for a misprediction threshold $\delta \ge 0$ we estimate the probability $p(\delta|s)$ given as
\begin{equation}\label{eq:p_state_delta}
	p(\delta|s) = \mathbb{P}\Big[ \big( V_{LDE}(\pi|s) - V_{LDE}(\mu|s) \big)
		\times \big( V(\pi|s) - V(\mu|s) \big) \le -\delta \Big].
\end{equation}
Fix a state $s \in \mathcal{S}$ and denote by $\mathcal{A}_s$ the set of historical actions that were taken from $s$:
\[
	\mathcal{A}_s = \big\{ a \in \mathcal{A} \ |\ \big( s, a, r(s,a) \big) \in \mathcal{D} \big\}.
\]
Let $\eta$ denote the difference of the policies $\pi$ and $\mu$ at the state $s$ and let $\varrho$ denote the difference between the reward $r$ and the baseline $\hat{r}$ at the state $s$:
\begin{align*}
	\eta(\,\cdot\,|s) &= \pi(\,\cdot\,|s) - \mu(\,\cdot\,|s) \ :\ \mathcal{A} \to [-1,1],
	\\
	\varrho(\,\cdot\,|s) &= r(s,\,\cdot\,) - \hat{r}(s) \ :\ \mathcal{A} \to \mathbb{R}.
\end{align*}

Note that $\eta(\,\cdot\,|s)$ is a signed measure on the action space $\mathcal{A}$ with
\begin{equation}\label{eq:eta_A}
	\eta(\mathcal{A}|s) = \sum\limits_{a \in \mathcal{A}} \eta(a|s) = 0
	\ \ \text{for any}\ \ 
	s \in \mathcal{S}.
\end{equation}
We express the difference of the LDE estimates in terms of $\varrho$ and $\eta$:
\[
	V_{LDE}(\pi|s) - V_{LDE}(\mu|s)
	= \sum_{a \in \mathcal{A}_s} \big(r(s,a) - \hat{r}(s)\big) \big(\pi(a|s) - \mu(a|s)\big)
	= \sum_{a \in \mathcal{A}_s} \varrho(a|s) \, \eta(a|s).
\]
Similarly, taking into account equality~\eqref{eq:eta_A}, we express the difference of values:
\[
	V(\pi|s) - V(\mu|s)
	= \sum_{a\in\mathcal{A}} r(s,a) \, \mu(a|s) - \sum_{a\in\mathcal{A}} r(s,a) \, \pi(a|s)
	= \sum_{a\in\mathcal{A}} \varrho(a|s) \, \eta(a|s).
\]
Next we rewrite probability~\eqref{eq:p_state_delta} as
\[
	p(\delta|s) = \mathbb{P}\Bigg[ \sum_{a\in\mathcal{A}_s} \varrho(a|s) \eta(a|s)
		\times \sum_{a\in\mathcal{A}} \varrho(a|s) \eta(a|s) \le -\delta \Bigg]
\]
and estimate it under the uniform distribution of the historical actions $\mathcal{A}_s \subset \mathcal{A}$ as per assumption~\eqref{eq:hist_assumption}.
Hence our problem becomes to estimate
\begin{equation}\label{eq:p_state_delta_U}
	p(\delta|s) = \mathbb{P}\Bigg[ \sum_{a\in\mathcal{A}_s} \varrho(a|s) \eta(a|s)
		\times \sum_{a\in\mathcal{A}} \varrho(a|s) \eta(a|s) \le -\delta
		\ \Big|\ \mathcal{A}_s \sim \mathcal{U}(2^\mathcal{A}) \Bigg].
\end{equation}

Let $\mathcal{A} = \{a_1, a_2, \ldots, a_n\}$ be a discrete action space and $\mathcal{A}_s = \{a'_1, a'_2, \ldots a'_m\} \sim \mathcal{U}(\mathcal{A}^m)$ be the set of historical actions taken from the state $s \in \mathcal{S}$ with some $m = m(s) \le n$.
Consider the random variable $\tau$ given by
\[
	\tau = \sum_{a\in\mathcal{A}_s} \varrho(a|s) \, \eta(a|s).
\]
Let $\{\Lambda_1, \Lambda_2, \ldots, \Lambda_K\}$ be the set of all subsets of $[n]$ of cardinality $m$ with $K = \binom{n}{m} = \frac{n!}{m! (n-m)!}$ and denote
\[
	\tau_k = \sum_{i\in\Lambda_k} \varrho(a_i|s) \, \eta(a_i|s) \in \mathbb{R}
	\ \ \text{for}\ \ 1 \le k \le K.
\]
Then $\tau \sim \mathcal{U}\{\tau_1, \tau_2, \ldots, \tau_K\}$ and the expectation of $\tau$, denoted by $\overline{\tau}$, is given as
\begin{align*}
	\overline{\tau}
	= \mathbb{E}[\tau]
	&= \frac{1}{K} \sum_{k=1}^K \tau_k
	= \frac{1}{K} \sum_{k=1}^K \sum_{i\in\Lambda_k} \varrho(a_i|s) \, \eta(a_i|s)
	\\
	&= \frac{\binom{n-1}{m-1}}{K} \sum_{i=1}^n \varrho(a_i|s) \, \eta(a_i|s)
	= \frac{m}{n} \sum_{a\in\mathcal{A}} \varrho(a|s) \, \eta(a|s).
\end{align*}
Hence we reframe problem~\eqref{eq:p_state_delta_U} in terms of $\tau$ and estimate via Cantelli's inequality:
\begin{align*}
	p(\delta|s) = \mathbb{P}\big[ \tau \overline{\tau}
	\le -\nicefrac{m \delta}{n} \big]
	&= \mathbb{P}\big[ -\tau \overline{\tau} + \overline{\tau}^2 \ge \nicefrac{m \delta}{n} + \overline{\tau}^2 \big]
	\\
	&\le \Bigg( 1 + \frac{\big(\nicefrac{m \delta}{n} + \overline{\tau}^2)^2}
		{\overline{\tau}^2 \mathrm{Var}[\tau]} \Bigg)^{-1}
	\le \Bigg( 1 + \frac{\nicefrac{m \delta}{n} + \overline{\tau}^2}{\mathrm{Var}[\tau]} \Bigg)^{-1}.
\end{align*}
From H{\"o}lder's inequality we obtain
\begin{align*}
	\mathrm{Var}[\tau]
	\le \mathbb{E}[\tau^2]
	&= \frac{1}{K} \sum_{k=1}^K \sum_{i\in\Lambda_k} \varrho(a_i|s)^2 \eta(a_i|s)^2
	\\
	&= \frac{m}{n} \sum_{a\in\mathcal{A}} \varrho(a|s)^2 \eta(a|s)^2
	\le \frac{m}{n} \|\varrho(\,\cdot\,|s)^2\|_\infty \|\eta(\,\cdot\,|s)^2\|_1,
\end{align*}
where
\[
	\|\varrho(\,\cdot\,|s)\|_\infty
	\le R
	:= \max\big\{ \max\mathcal{R} - \hat{r}(s),\ \hat{r}(s) - \min\mathcal{R} \big\}
	< \infty
\]
and
\[
	\|\eta(\,\cdot\,|s)^2\|_1
	\le \|\eta(\,\cdot\,|s)\|_1
	\le \|\pi(\,\cdot\,|s)\|_1 + \|\mu(\,\cdot\,|s)\|_1
	\le 2.
\]
Then taking into account that
\[
	\overline{\tau}
	= \frac{m}{n} \sum_{a\in\mathcal{A}} \varrho(a|s) \, \eta(a|s)
	= \frac{m}{n} \big( V(\pi|s) - V(\mu|s) \big).
\]
we obtain
\[
	\frac{\nicefrac{m \delta}{n} + \overline{\tau}^2}{\mathrm{Var}[\tau]}
	\ge \frac{n \delta + m \big( V(\pi|s) - V(\mu|s) \big)^2}{2nR^2}
\]
and arrive to the final estimate
\[
	p(\delta|s) \le \Bigg( 1 + \frac{n \delta + m \big( V(\pi|s) - V(\mu|s) \big)^2}{2nR^2} \Bigg)^{-1},
\]
which completes the proof of Theorem~\ref{thm:state_comp}.

\subsection{Proof of Theorem~\ref{thm:policy_comp} (Policy Comparison)}\label{sec:policy_comp_proof}
In this section we analyze the caliber comparison of policies $\pi$ and $\mu$ on the whole state space $\mathcal{S}$ by relating the difference of the LDE estimates with the difference of values:
\[
	V_{LDE}(\pi) - V_{LDE}(\mu) \ \sim\ V(\pi) - V(\mu).
\]
Specifically, we estimate the probability $p$ that the LDE comparison of policies $\pi$ and $\mu$ is incorrect, given as
\begin{equation}\label{eq:p_policy}
	p = \mathbb{P}\Big[ \big( V_{LDE}(\pi) - V_{LDE}(\mu) \big)
		\times \big( V(\pi) - V(\mu) \big) \le 0 \Big].
\end{equation}
Following the notations from Section~\ref{sec:state_comp_proof} we write
\begin{gather*}
	V_{LDE}(\pi) - V_{LDE}(\mu)
	= \mathbb{E}_s\Big[ V_{LDE}(\pi|s) - V_{LDE}(\mu|s) \Big]
	= \mathbb{E}_s\Big[ \sum_{a \in \mathcal{A}_s} \varrho(a|s) \, \eta(a|s) \Big],
	\\
	V(\pi) - V(\mu)
	= \mathbb{E}_s\Big[ V(\pi|s) - V(\mu|s) \Big]
	= \sum_{a\in\mathcal{A}} \mathbb{E}_s\big[ \varrho(a|s) \, \eta(a|s) \big].
\end{gather*}
Hence we rewrite probability~\eqref{eq:p_policy} as
\[
	p = \mathbb{P}\Bigg[ \mathbb{E}_s\Big[ \sum_{a \in \mathcal{A}_s} \varrho(a|s) \, \eta(a|s) \Big]
		\times \sum_{a \in \mathcal{A}} \mathbb{E}_s\big[ \varrho(a|s) \, \eta(a|s) \big] \le 0 \Bigg].
\]
Denote by $m$ the number of historical actions taken from each state $s \in \mathcal{S}$, i.e. $|\mathcal{A}_s| = m \le n$ for each $s \in \mathcal{S}$, then under assumption~\eqref{eq:hist_assumption} our problem becomes
\begin{equation}\label{eq:p_policy_U}
	p = \mathbb{P}\Bigg[ \sum_{a \in \mathcal{A}_m} \mathbb{E}_s\big[ \varrho(a|s) \, \eta(a|s) \big]
	\times \sum_{a \in \mathcal{A}} \mathbb{E}_s\big[ \varrho(a|s) \, \eta(a|s) \big] \le 0
		\ \Big|\ \mathcal{A}_m \sim \mathcal{U(A)} \Bigg].
\end{equation}

Consider the random variable $\tau$ given by
\[
	\tau = \sum_{a\in\mathcal{A}_m} \mathbb{E}_s\big[ \varrho(a|s) \, \eta(a|s) \big].
\]
Repeating the estimates from Section~\ref{sec:state_comp_proof} we derive
\[
	\overline{\tau}
	= \mathbb{E}[\tau]
	= \frac{m}{n} \sum_{a\in\mathcal{A}} \mathbb{E}_s\big[ \varrho(a|s) \, \eta(a|s) \big]
	= \frac{m}{n} \big( V(\pi) - V(\mu) \big),
\]
and $\tau \overline{\tau} \in [-2\overline{\tau} R, 2\overline{\tau} R]$ with $R = \max\big\{ \max\mathcal{R} - \mathbb{E}_s[\hat{r}(s)],\ \mathbb{E}_s[\hat{r}(s)] - \min\mathcal{R} \big\} < \infty$.
Then by using Hoeffding's inequality we estimate probability~\eqref{eq:p_policy_U} as
\begin{align*}
	p = \mathbb{P}\big[ \tau \overline{\tau} \le 0 \big]
	&= \mathbb{P}\big[ -\tau \overline{\tau} + \overline{\tau}^2 \ge \overline{\tau}^2 \big]
	\\
	&\le \exp\Bigg( -\frac{\overline{\tau}^2}{8R^2} \Bigg)
	= \exp\Bigg( -\frac{m^2 \big( V(\pi) - V(\mu) \big)^2}{8n^2R^2} \Bigg),
\end{align*}
which completes the proof of Theorem~\ref{thm:policy_comp}.

\newpage
\section{Additional Detail for Numerical Experiments}\label{sec:numerics_appendix}
In this section we provide a detailed explanation of the numerical examples presented in Section~\ref{sec:numerics}.
Our experiments are performed in \texttt{Python 3.6} and the source code reproducing the reported results is provided at~\url{https://github.com/joedaws/lde2021}.

\subsection{Example 1: Synthetic data}\label{sec:ex1_appendix}
In this example we construct a synthetic contextual bandit environment as follows:
\begin{itemize}
    \item The state space $\mathcal{S}$ consists of $100$ uniformly distributed points from the interval $[0,1]$;
    \item The action space $\mathcal{A}$ consists of $100$ uniformly distributed points from the interval $[-1,1]$;
    \item The reward space $\mathcal{R}$ and the reward function $r : \mathcal{S} \times \mathcal{A} \to \mathcal{R}$ are specified for each example.
\end{itemize}
In this setting we explore how the performance of the policy evaluation methods depends on the target policies' value difference $\alpha$, defined in~\eqref{eq:value_diff}, and the amount $m$ of historical actions per state.
To this end, the target policies $\pi$ and $\mu$ are sampled from the normal distribution, normalized via the softmax function, and then adjusted linearly so that $V(\pi) - V(\mu) = \alpha$.
The historical data $\mathcal{D}$ is collected under a randomly generated exploratory behavioral policy, which is sampled from the normal distribution and normalized via the softmax function.
Once the environment is set up, a single test consists of the following steps:
\begin{enumerate}
    \item The target policies $\pi$ and $\mu$ are generated randomly with the prescribed value of $\alpha$;
    \item The historical data $\mathcal{D}$ is collected under a randomly generated exploratory behavioral policy by sampling $m$ (potentially repeating) historical actions for each state $s \in \mathcal{S}$;
    \item The policy evaluation methods (specifically, the LDE, the DRE, the IPS, and the DiM) are employed to evaluate and compare the target policies $\pi$ and $\mu$ on the historical dataset $\mathcal{D}$.
\end{enumerate}
For each example we perform 1,000 tests and analyze the accumulated results.
Specifically, we report the performance of the considered methods on the task of policy evaluation and the task of policy comparison.

In the case of evaluation, the normalized approximation error is measured in each test, i.e.
\begin{equation}\label{eq:eval_score}
    \textrm{error}_{method}(\pi;\mathcal{D}) = \frac{V_{method}(\pi;\mathcal{D}) - V(\pi)}{V_{max} - V_{min}} \in [-1,1]
\end{equation}
where $V_{max}$ and $V_{min}$ are the values of the optimal and the pessimal policies (for the given environment) respectively.

In the case of comparison, the percentage of correct comparisons between the target policies is measured in each test via~\eqref{eq:comp_score} and reported in Section~\ref{sec:ex1}.
Additionally, our theoretical analysis, particularly Theorem~\ref{thm:state_comp} from Section~\ref{sec:theory_appendix}, suggests that the LDE is capable of performing comparison on the individual states $s \in \mathcal{S}$, in addition to the usual policy comparison over the whole state space $\mathcal{S}$.
We present the numerical verification of this claim by computing the normalized histogram of the products of differences of per-state comparisons, as explained in Section~\ref{sec:state_comp_proof}.

Additionally, for each method we compute the correlation statistics over all tests between the differences of the true values of $\pi$ and $\mu$, and their predicted estimates, i.e. between the sequences
\begin{equation}\label{eq:correlation}
    \Big\{ V(\pi_t) - V(\mu_t) \Big\}_{t=1}^{1,000}
    \quad\text{and}\quad
    \Big\{ V_{method}(\pi_t;\mathcal{D}_t) - V_{method}(\mu_t;\mathcal{D}_t) \Big\}_{t=1}^{1,000}
\end{equation}
where $\pi_t, \mu_t$ are the target policies generated during test number $1 \le t \le 1,000$, and $\mathcal{D}_t$ is the corresponding historical dataset.
In our experiments we observe that for the LDE this correlation is stronger than for the other methods, which supports our claim of the advantage of the LDE on the policy comparison tasks.

\clearpage
\subsubsection{Example 1.1}\label{sec:ex1.1_appendix}
\begin{figure}[hbt!]
    \centering
    \includegraphics[width=.49\linewidth, trim={5em 5em 0em 0em}, clip]{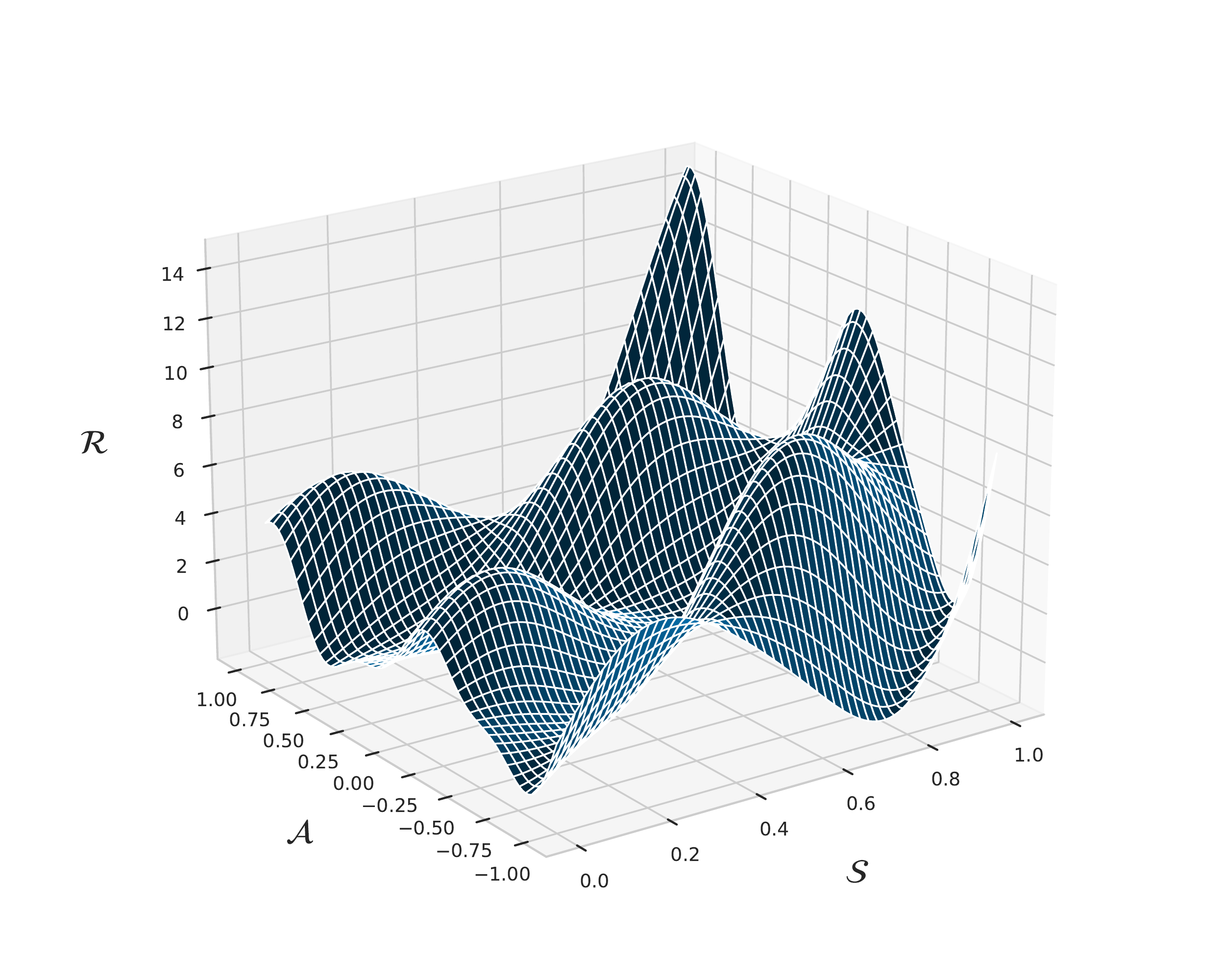}
    \includegraphics[width=.49\linewidth]{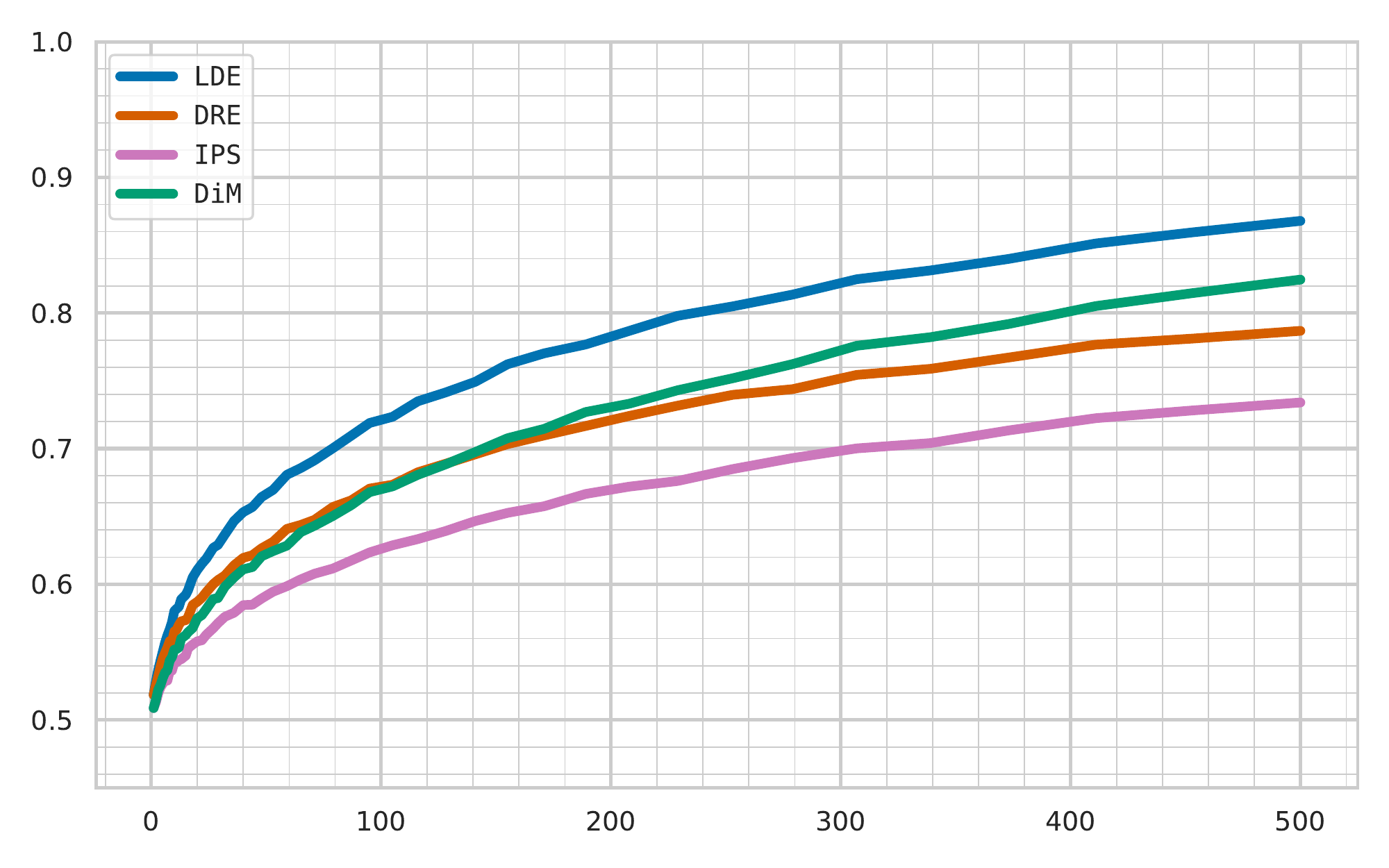}
    \caption{Reward function (left) and the percentage of correct per-state comparisons depending on the amount of historical data $m$ (right, $\alpha = .01$).}
    \label{fig:ex1.1_state}
\end{figure}

\begin{figure}[hbt!]
    \centering
    \includegraphics[width=.49\linewidth]{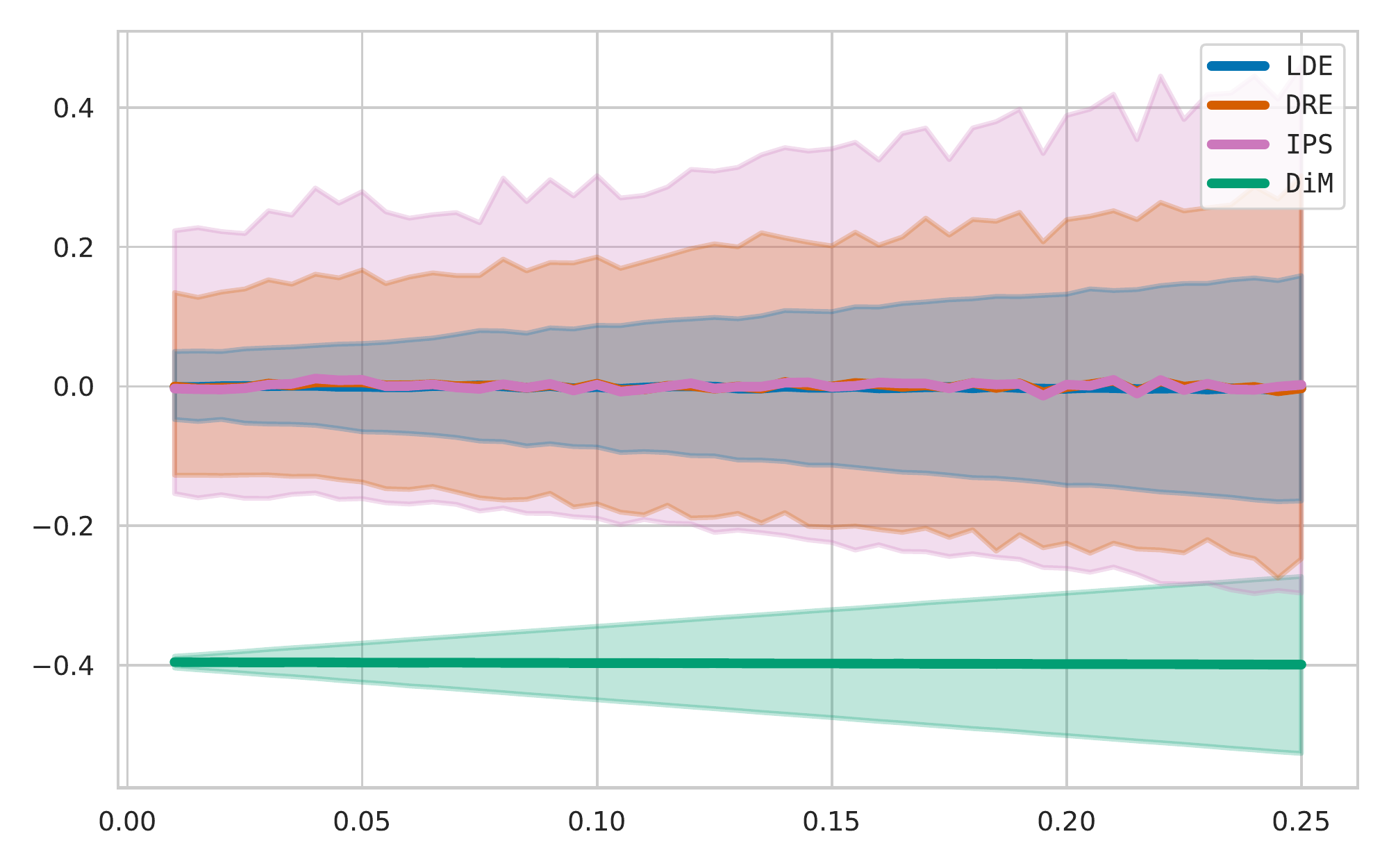}
    \includegraphics[width=.49\linewidth]{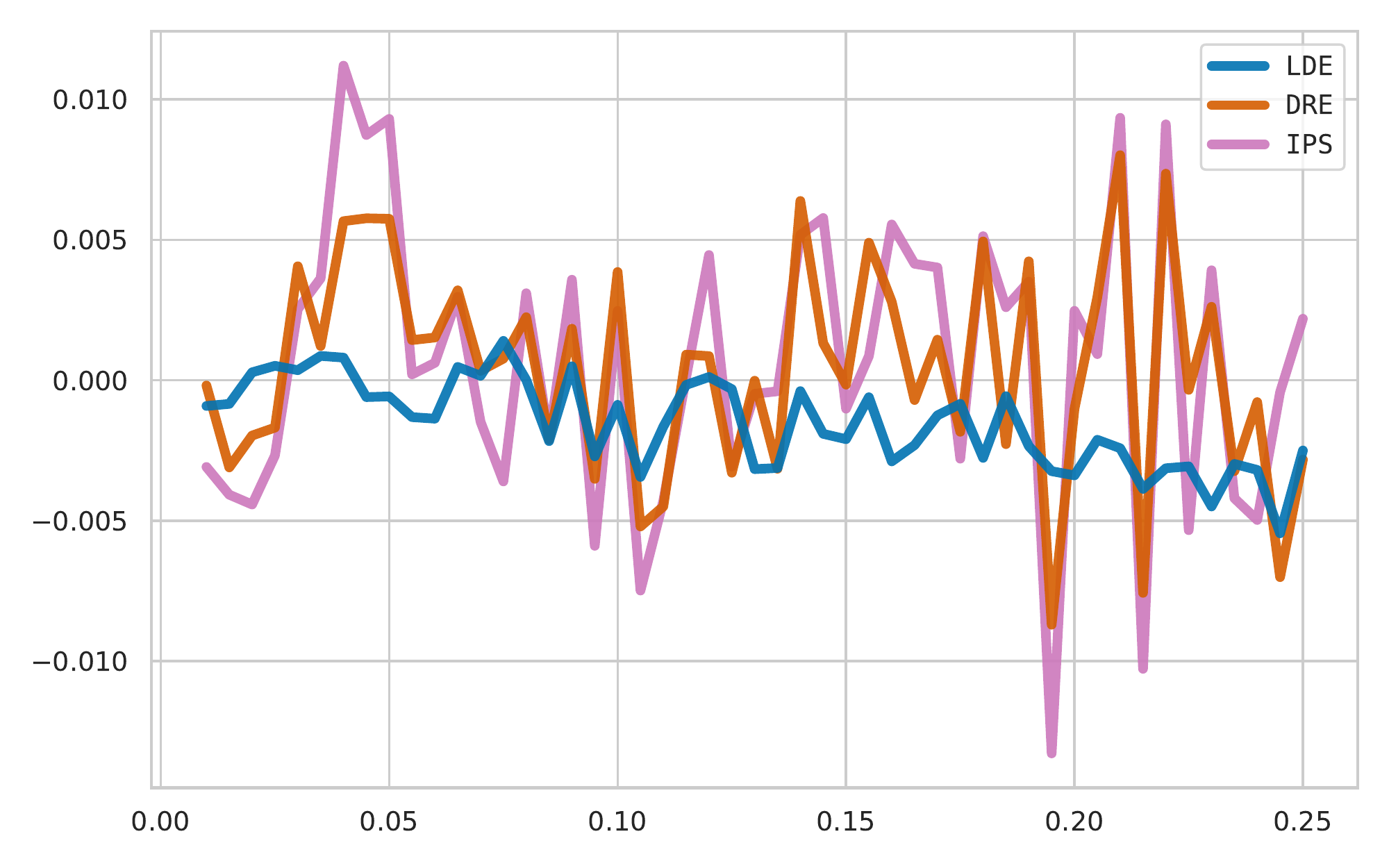}
    \\
    \includegraphics[width=.49\linewidth]{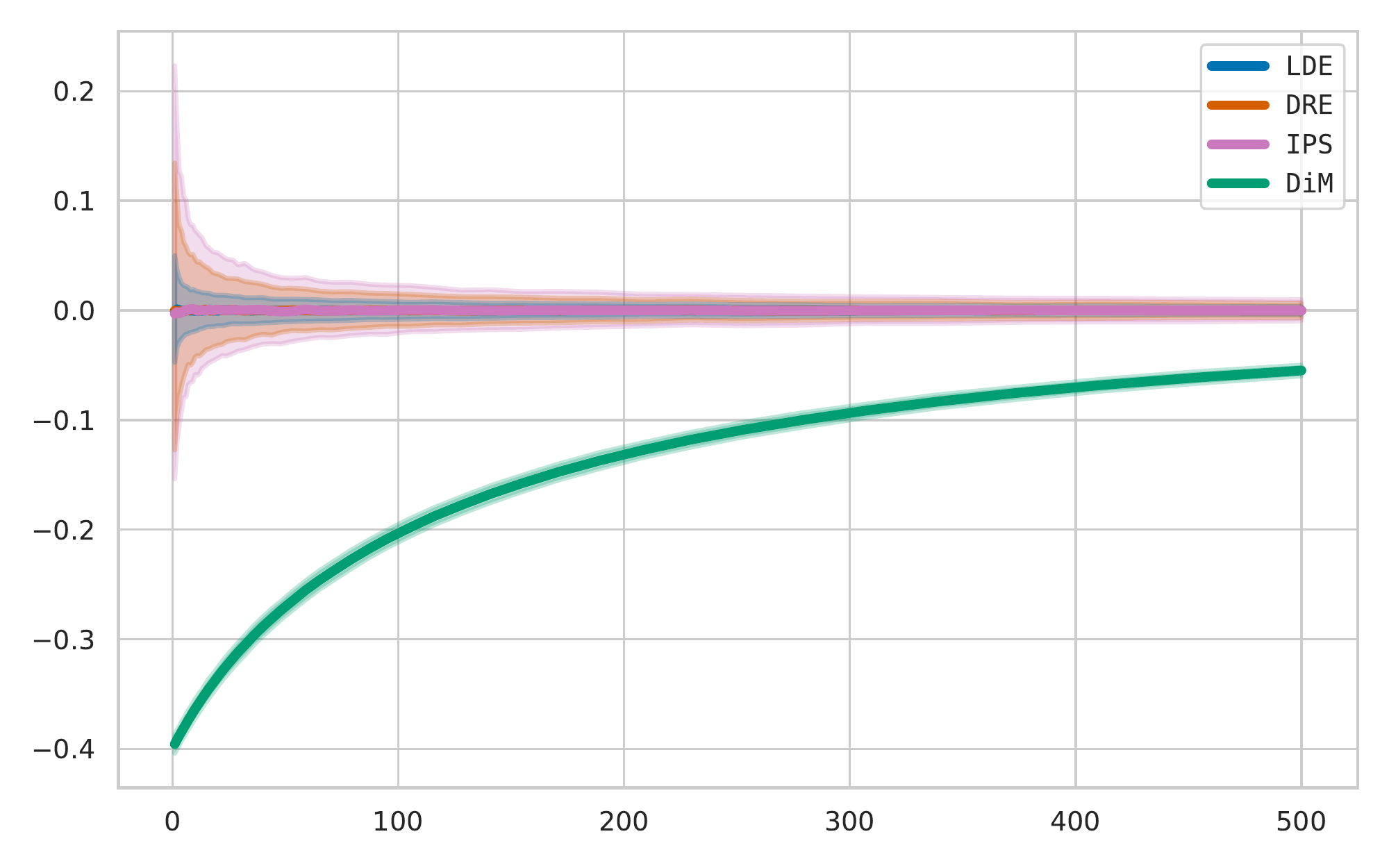}
    \includegraphics[width=.49\linewidth]{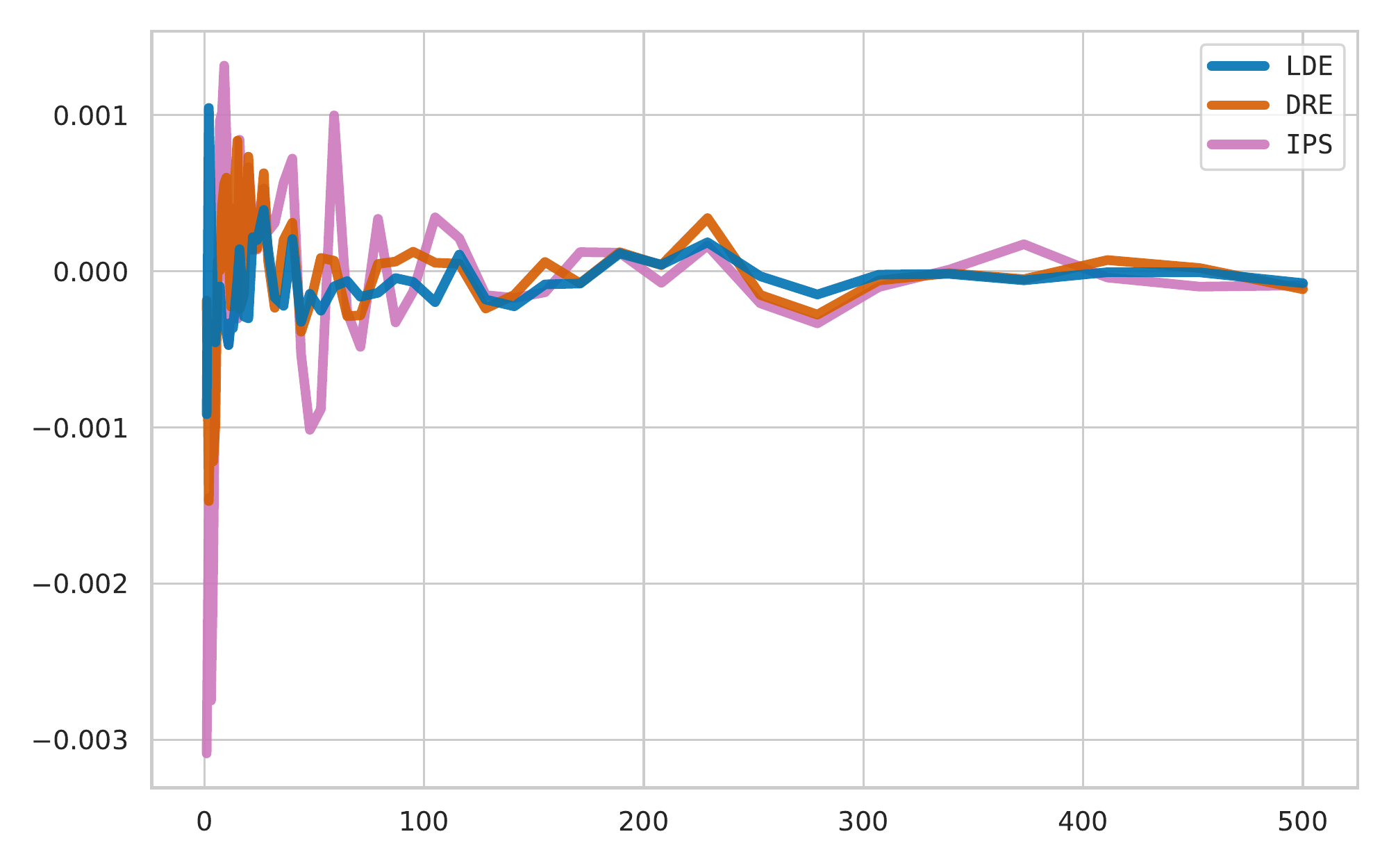}
    \caption{Policy evaluation error (normalized) depending on the value difference $\alpha$ (top, $m = 1$, see Table~\ref{tab:ex1.1_eval_a}) or the amount of historical data $m$ (bottom, $\alpha = .01$, see Table~\ref{tab:ex1.1_eval_m}).}
    \label{fig:ex1.1_eval}
\end{figure}

\begin{figure}[hbt!]
    \centering
    \includegraphics[width=.24\linewidth, trim={3em 3em 6em 3em}, clip]{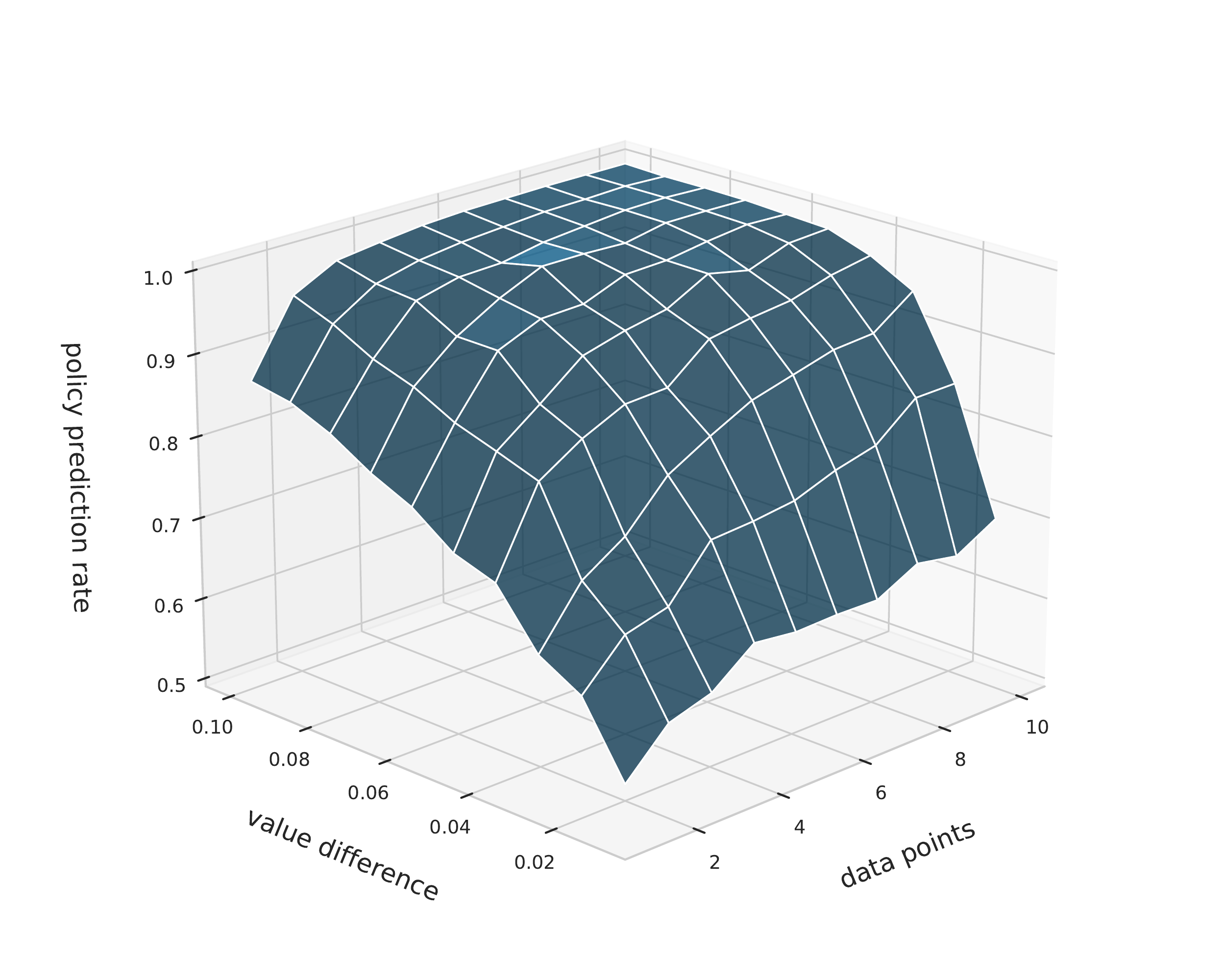}
    \includegraphics[width=.24\linewidth, trim={3em 3em 6em 3em}, clip]{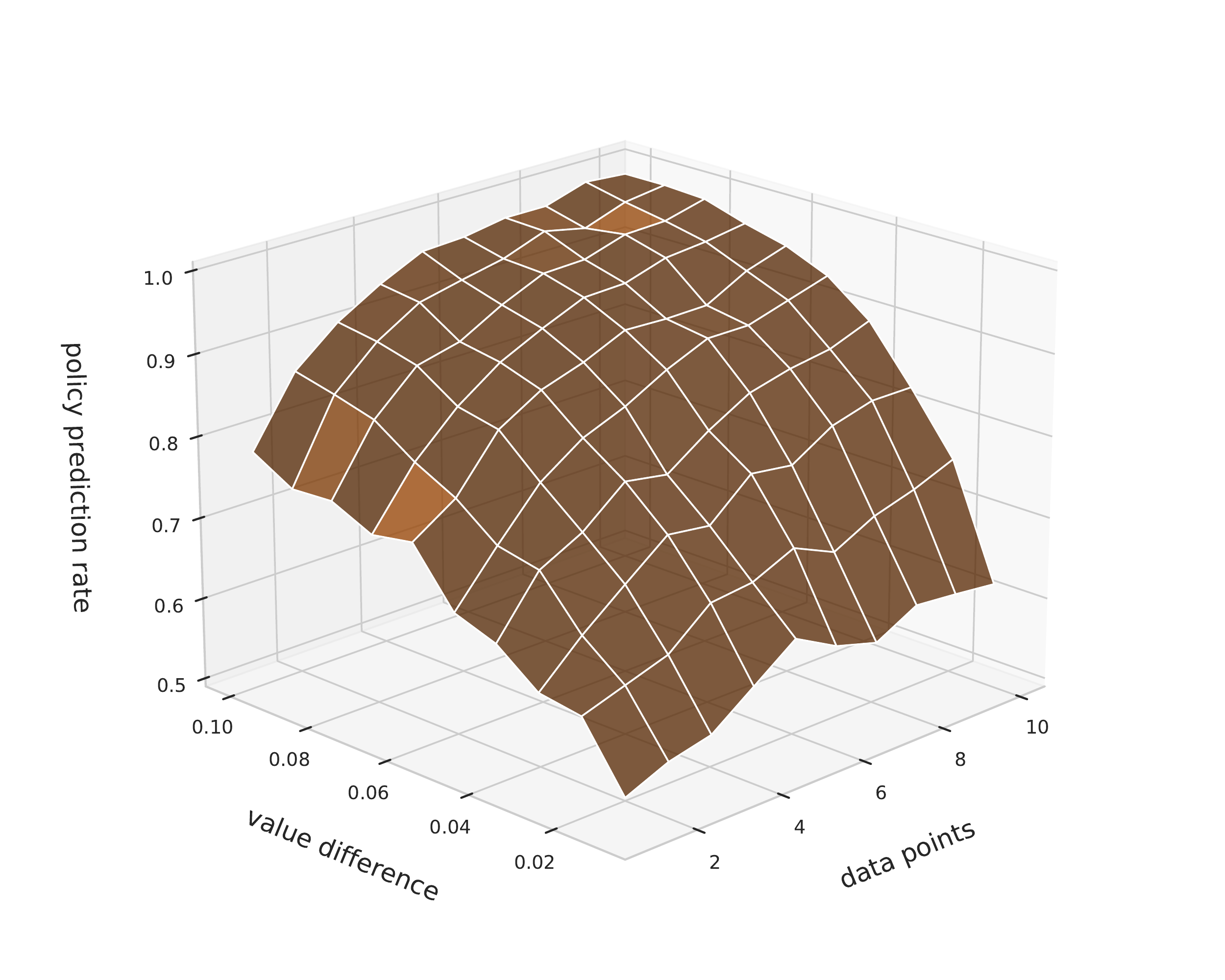}
    \includegraphics[width=.24\linewidth, trim={3em 3em 6em 3em}, clip]{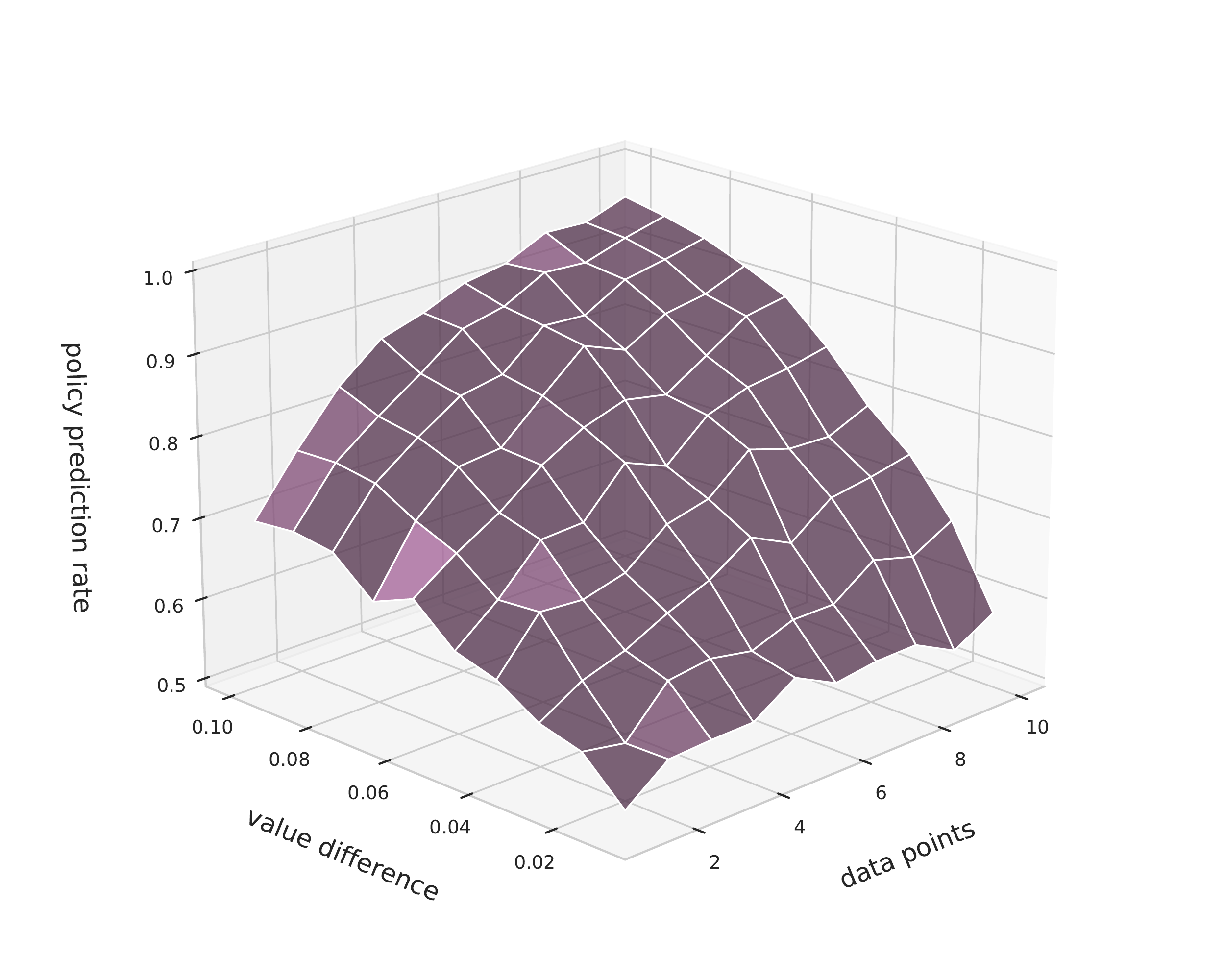}
    \includegraphics[width=.24\linewidth, trim={3em 3em 6em 3em}, clip]{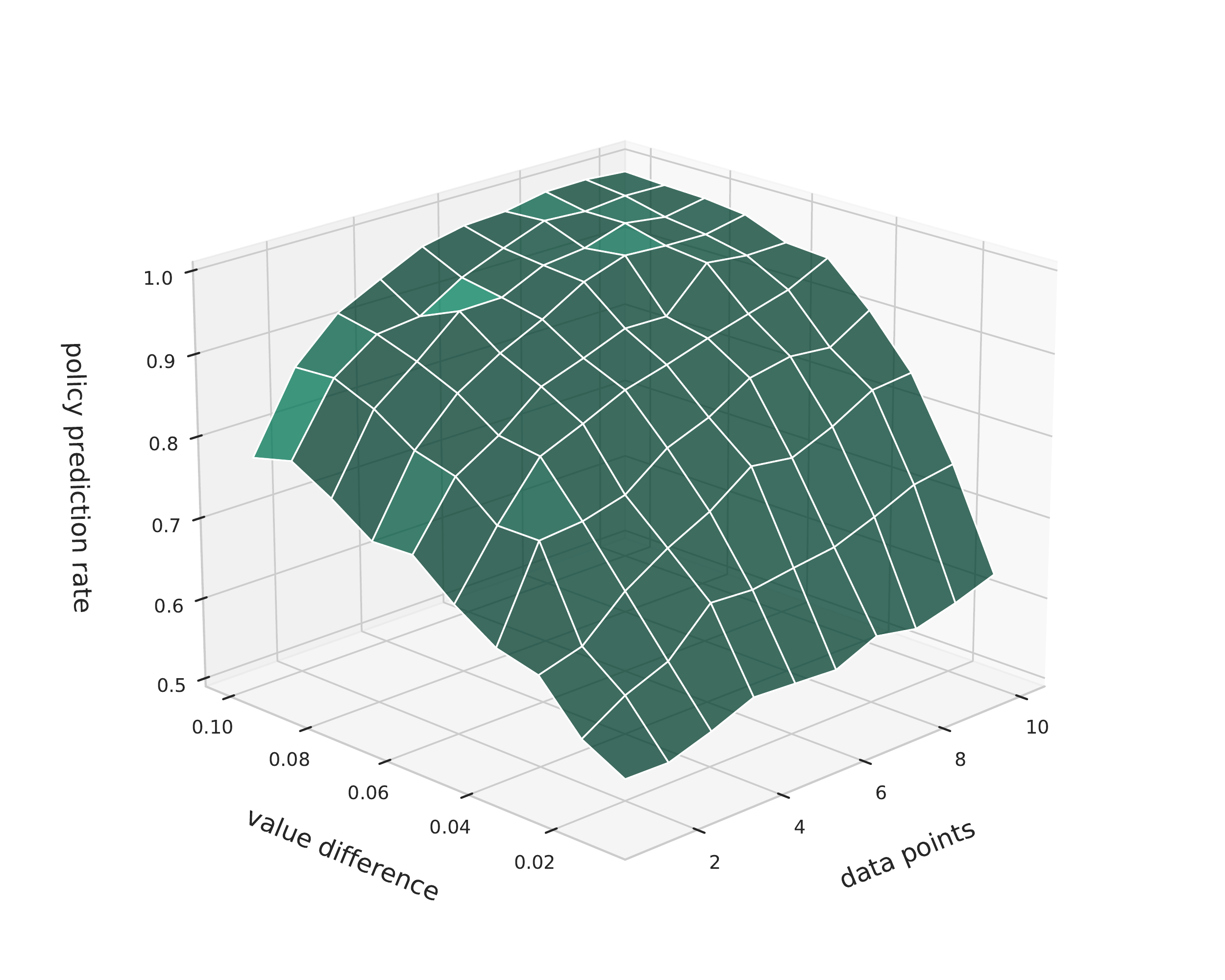}
    \\
    \includegraphics[width=.24\linewidth, trim={6em 3em 3em 3em}, clip]{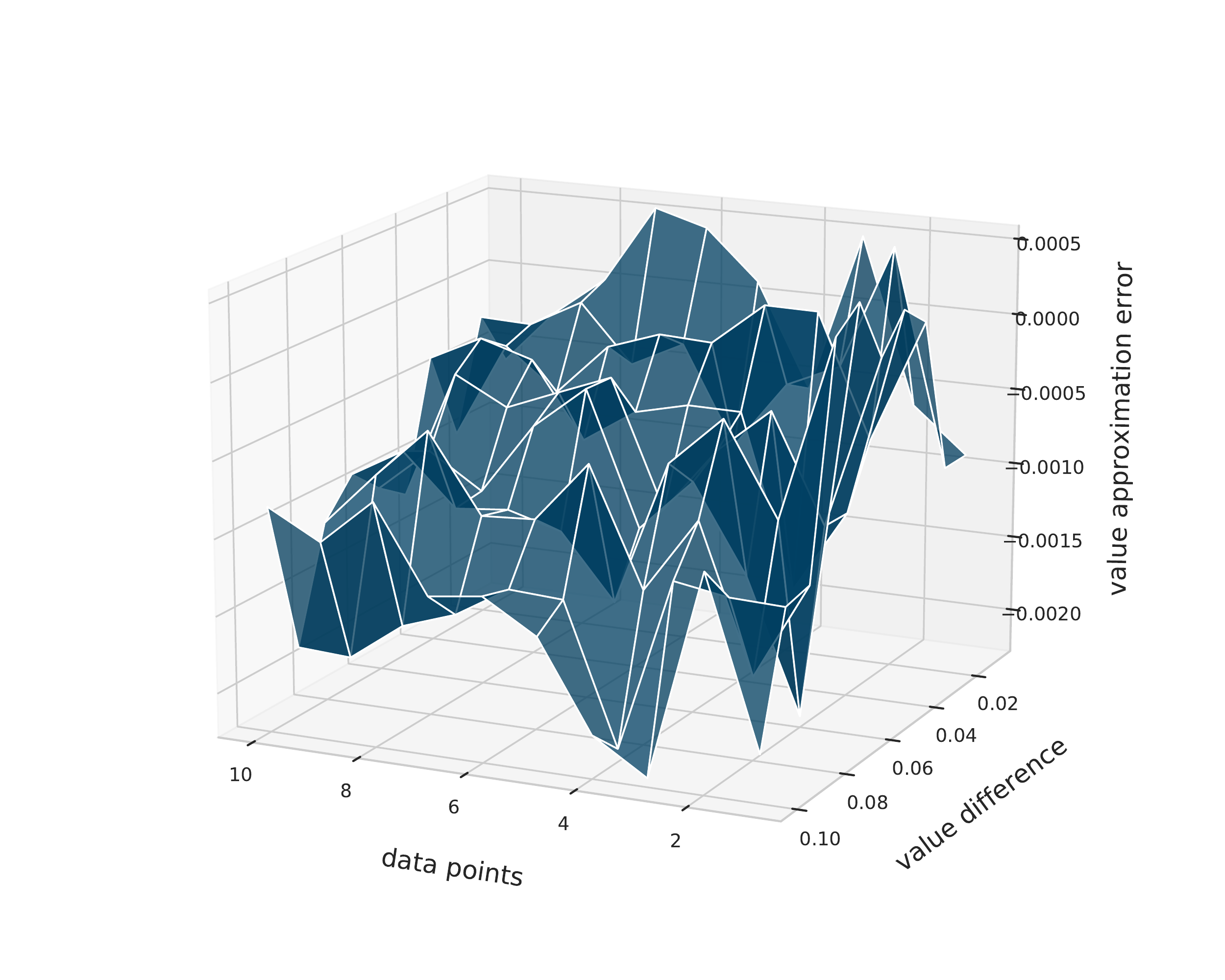}
    \includegraphics[width=.24\linewidth, trim={6em 3em 3em 3em}, clip]{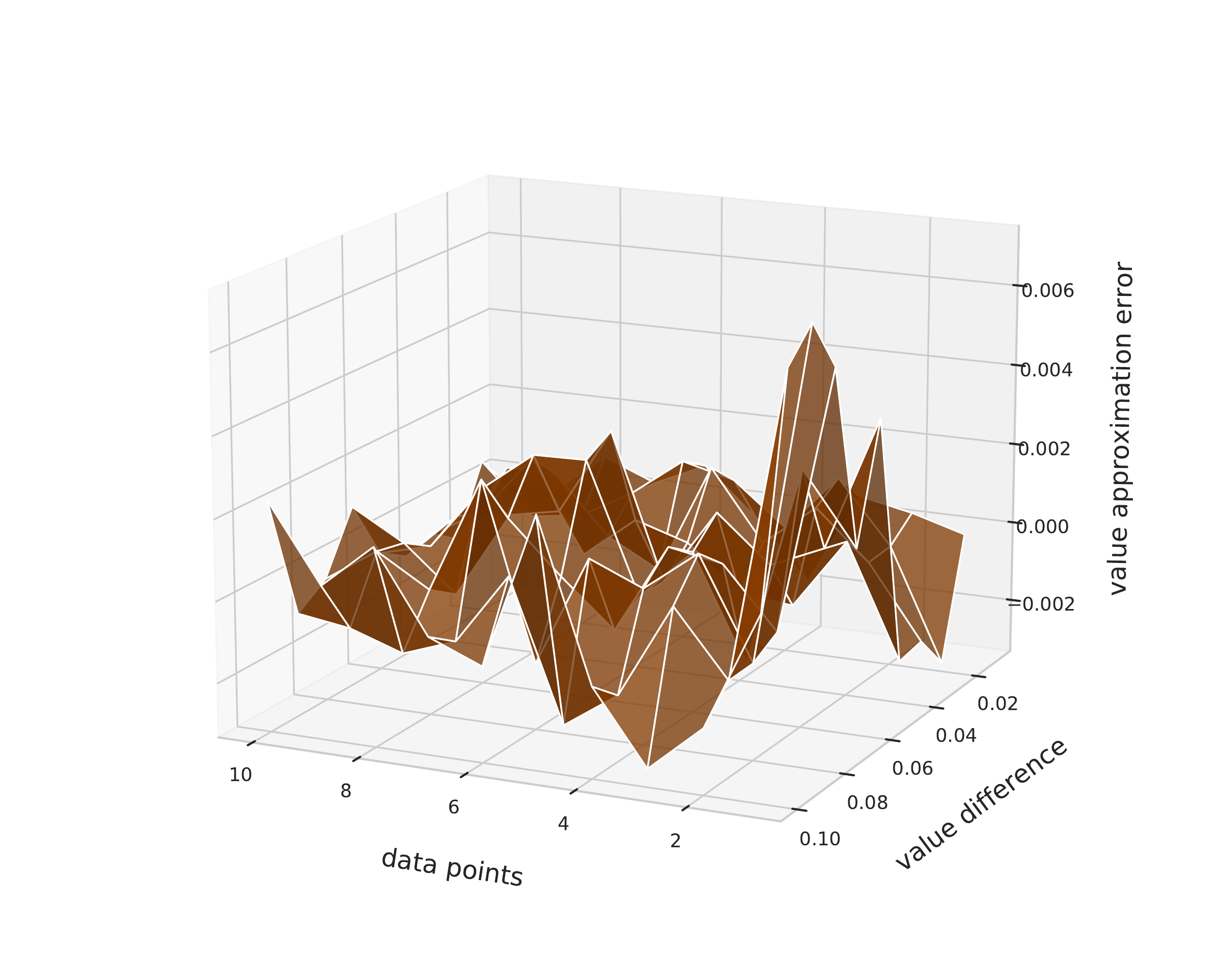}
    \includegraphics[width=.24\linewidth, trim={6em 3em 3em 3em}, clip]{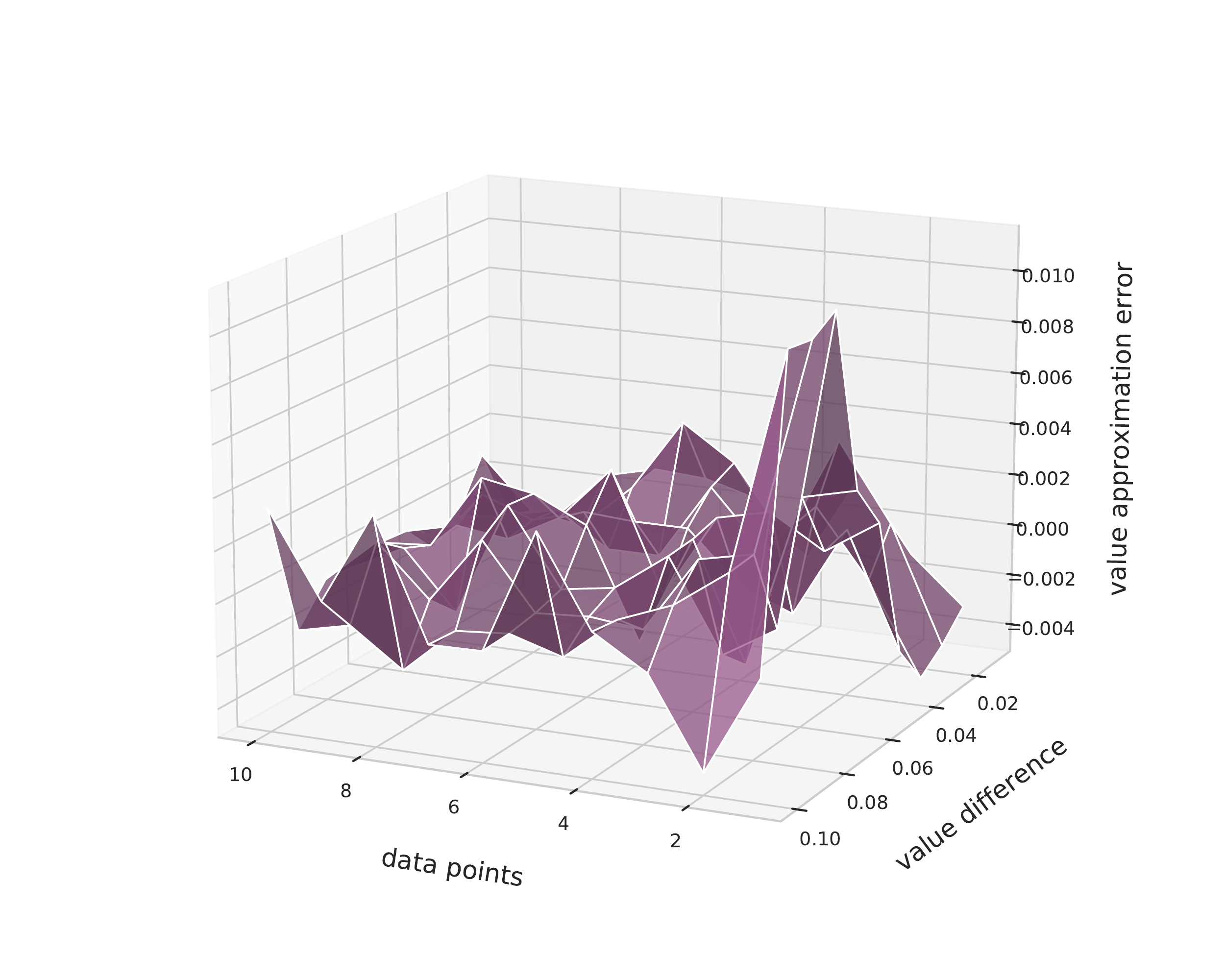}
    \includegraphics[width=.24\linewidth, trim={6em 3em 3em 3em}, clip]{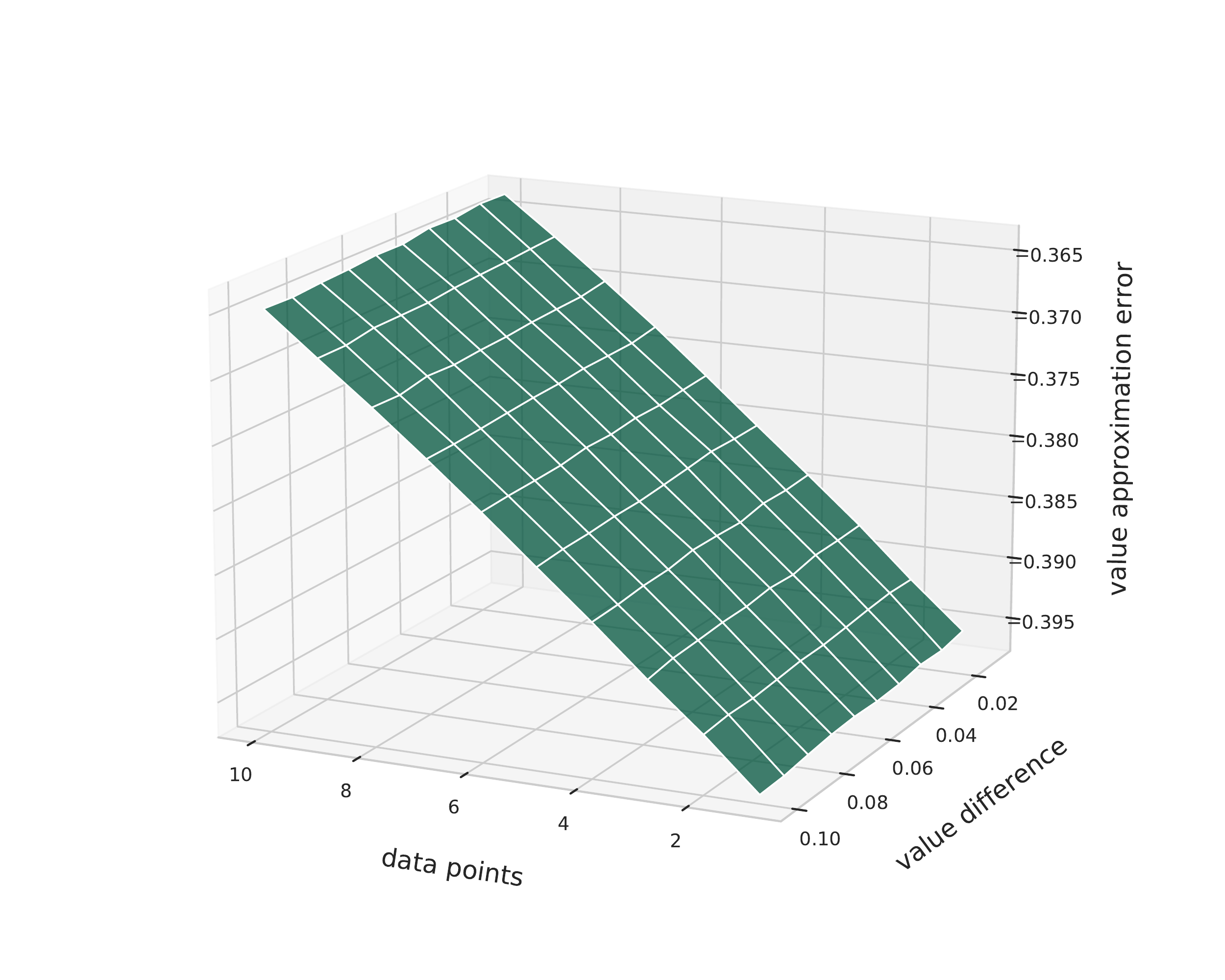}
    \caption{Policy comparison rate (top) and policy evaluation error (bottom) of the LDE, the DRE, the IPS, and the DiM respectively as a function of the value difference $\alpha$ and the number of data points $m$.}
    \label{fig:ex1.1_3d}
\end{figure}

\begin{figure}[hbt!]
    \centering
    \includegraphics[width=\linewidth, trim={4em 6em 8em 9em}, clip]{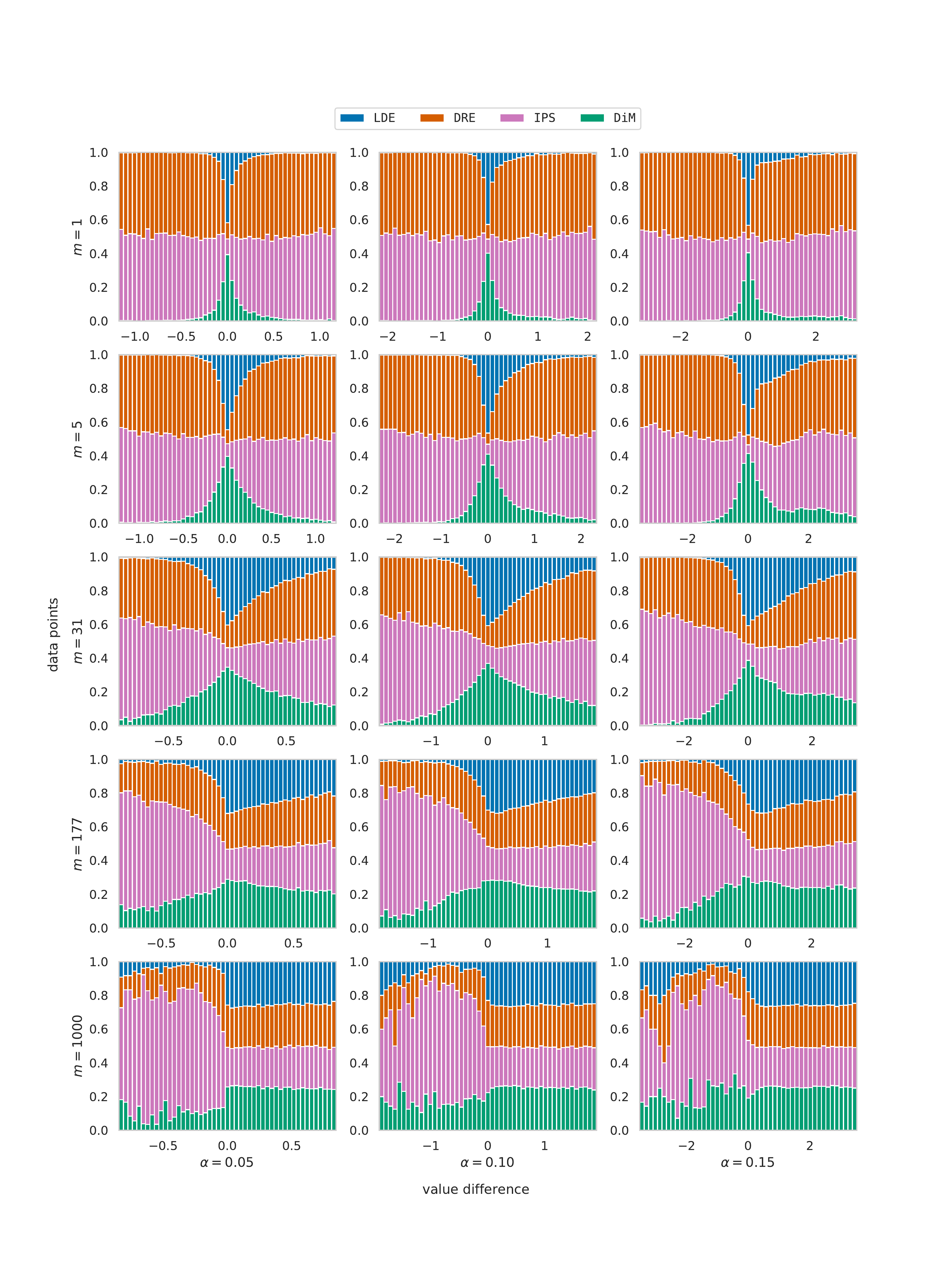}
    \caption{Normalized histograms of products of per-state differences with various values of $\alpha$ and $m$.}
    \label{fig:ex1.1_grid}
\end{figure}

This section contains supplementary figures for Example~1.1 from Section~\ref{sec:ex1}.
The reward function $r(s,a)$, given by~\eqref{eq:ex1.1_reward}, is shown in Figure~\ref{fig:ex1.1_state} (left).
The policy evaluation errors~\eqref{eq:eval_score}, depending on either the value difference $\alpha$, defined in~\eqref{eq:value_diff}, or the number of historical actions $m$, are shown in Figure~\ref{fig:ex1.1_eval}.
In Figure~\ref{fig:ex1.1_3d} we demonstrate the performance of the policy evaluation methods on the tasks of policy comparison (top) and evaluation (bottom) depending on the values of $\alpha$ and $m$.
The correlation between the differences of true values and the predicted estimates are presented in Table~\ref{tab:ex1.1_correlation}.
We verify the capacity of the LDE to perform per-state policy comparisons and show the per-state comparison rate depending on the number of data points $m$ in Figure~\ref{fig:ex1.1_state} (right) and the normalized histogram of the products of per-state differences in Figure~\ref{fig:ex1.1_grid}, as discussed in Section~\ref{sec:theory_appendix}.

We note that in this example, due to the regular reward distribution between the state-action pairs, the LDE performs well on both the comparison and evaluation tasks (see Figure~\ref{fig:ex1.1_eval} and Tables~\ref{tab:ex1.1_eval_a} and~\ref{tab:ex1.1_eval_m}), demonstrating the comparable average approximation error while enjoying a smaller variance.

\begin{table}[hbt!]
    \centering\small
    \caption{Policy evaluation errors depending on the value difference $\alpha$, see Figure~\ref{fig:ex1.1_eval} (top).}
    \label{tab:ex1.1_eval_a}
    \begin{tabular}{ccccccccc}
        \toprule
        & \multicolumn{4}{c}{Approximation errors (average)} & \multicolumn{4}{c}{Approximation errors (std)}
        \\\cmidrule(lr){2-5}\cmidrule(lr){6-9}
        $\alpha$ & LDE & DRE & IPS & DiM & LDE & DRE & IPS & DiM
        \\\midrule
        0.05 & \textbf{-5.77e-04} & \ 5.75e-03 & \ 9.31e-03 & -3.96e-01 & 3.86e-02 & 9.81e-02 & 1.43e-01 & 2.49e-02
        \\
        0.10 & \textbf{-8.75e-04} & \ 3.86e-03 & \ 2.47e-03 & -3.97e-01 & 5.80e-02 & 1.28e-01 & 1.77e-01 & 4.96e-02
        \\
        0.15 & -2.10e-03 & \textbf{-1.57e-04} & -1.01e-03 & -3.98e-01 & 7.98e-02 & 1.68e-01 & 2.25e-01 & 7.43e-02
        \\
        0.20 & -3.39e-03 & \textbf{-1.02e-03} & \ 2.47e-03 & -3.98e-01 & 1.03e-01 & 2.13e-01 & 2.58e-01 & 9.91e-02
        \\
        0.25 & -2.50e-03 & -2.82e-03 & \ \textbf{2.19e-03} & -3.99e-01 & 1.27e-01 & 2.33e-01 & 2.91e-01 & 1.24e-01
        \\\bottomrule
    \end{tabular}
\end{table}

\begin{table}[hbt!]
    \centering\small
    \caption{Policy evaluation errors depending on the number of data points $m$, see Figure~\ref{fig:ex1.1_eval} (bottom).}
    \label{tab:ex1.1_eval_m}
    \begin{tabular}{ccccccccc}
        \toprule
        & \multicolumn{4}{c}{Approximation errors (average)} & \multicolumn{4}{c}{Approximation errors (std)}
        \\\cmidrule(lr){2-5}\cmidrule(lr){6-9}
        $m$ & LDE & DRE & IPS & DiM & LDE & DRE & IPS & DiM
        \\\midrule
        10 & \textbf{-3.98e-04} & \ 5.99e-04 & \ 8.54e-04 & -3.64e-01 & 1.12e-02 & 2.83e-02 & 4.29e-02 & 5.68e-03
        \\
        27 & \ \textbf{3.93e-04} & \ 6.27e-04 & \ 5.30e-04 & -3.17e-01 & 7.58e-03 & 1.71e-02 & 2.60e-02 & 5.44e-03
        \\
        71 & \textbf{-1.63e-04} & -2.82e-04 & -4.84e-04 & -2.38e-01 & 5.64e-03 & 1.05e-02 & 1.55e-02 & 5.19e-03
        \\
        189 & \ \textbf{1.14e-04} & \ 1.22e-04 & \ 1.19e-04 & -1.37e-01 & 3.75e-03 & 6.53e-03 & 9.59e-03 & 4.49e-03
        \\
        500 & -7.60e-05 & \textbf{-1.15e-04} & -9.10e-05 & -5.48e-02 & 2.22e-03 & 4.01e-03 & 5.91e-03 & 3.04e-03
        \\\bottomrule
    \end{tabular}
\end{table}

\begin{table}[hbt!]
    \centering\small
    \caption{Correlation statistics computed on sequences~\eqref{eq:correlation} from Figure~\ref{fig:ex1.1_3d}.}
    \label{tab:ex1.1_correlation}
    \begin{tabular}{cccccccccc}
        \toprule
        & \multicolumn{4}{c}{Pearson's coefficient} & \multicolumn{4}{c}{Spearman's coefficient}
        \\\cmidrule(lr){2-5}\cmidrule(lr){6-9}
        $m$ & LDE & DRE & IPS & DiM & LDE & DRE & IPS & DiM
        \\\midrule
        1 & \textbf{0.3594} & 0.2323 & 0.1701 & 0.2678 & \textbf{0.3470} & 0.2321 & 0.1649 & 0.2569
        \\
        2 & \textbf{0.4775} & 0.3079 & 0.2292 & 0.3627 & \textbf{0.4730} & 0.3168 & 0.2307 & 0.3490
        \\
        3 & \textbf{0.5495} & 0.3773 & 0.2653 & 0.4132 & \textbf{0.5496} & 0.3870 & 0.2707 & 0.4039
        \\
        4 & \textbf{0.5961} & 0.4052 & 0.3058 & 0.4671 & \textbf{0.5985} & 0.4227 & 0.3099 & 0.4607
        \\
        5 & \textbf{0.6404} & 0.4542 & 0.3402 & 0.5072 & \textbf{0.6484} & 0.4691 & 0.3412 & 0.5037
        \\
        6 & \textbf{0.6783} & 0.4935 & 0.3769 & 0.5488 & \textbf{0.6848} & 0.5069 & 0.3824 & 0.5475
        \\
        7 & \textbf{0.7113} & 0.5121 & 0.3845 & 0.5763 & \textbf{0.7211} & 0.5306 & 0.3949 & 0.5761
        \\
        8 & \textbf{0.7333} & 0.5442 & 0.4214 & 0.6172 & \textbf{0.7446} & 0.5630 & 0.4274 & 0.6193
        \\
        9 & \textbf{0.7497} & 0.5743 & 0.4498 & 0.6265 & \textbf{0.7614} & 0.5908 & 0.4590 & 0.6311
        \\
        10 & \textbf{0.7629} & 0.5886 & 0.4619 & 0.6425 & \textbf{0.7730} & 0.6088 & 0.4685 & 0.6468
        \\\bottomrule
    \end{tabular}
\end{table}

\clearpage
\subsubsection{Example 1.2}\label{sec:ex1.2_appendix}
\begin{figure}[hbt!]
    \centering
    \includegraphics[width=.49\linewidth, trim={5em 5em 0em 0em}, clip]{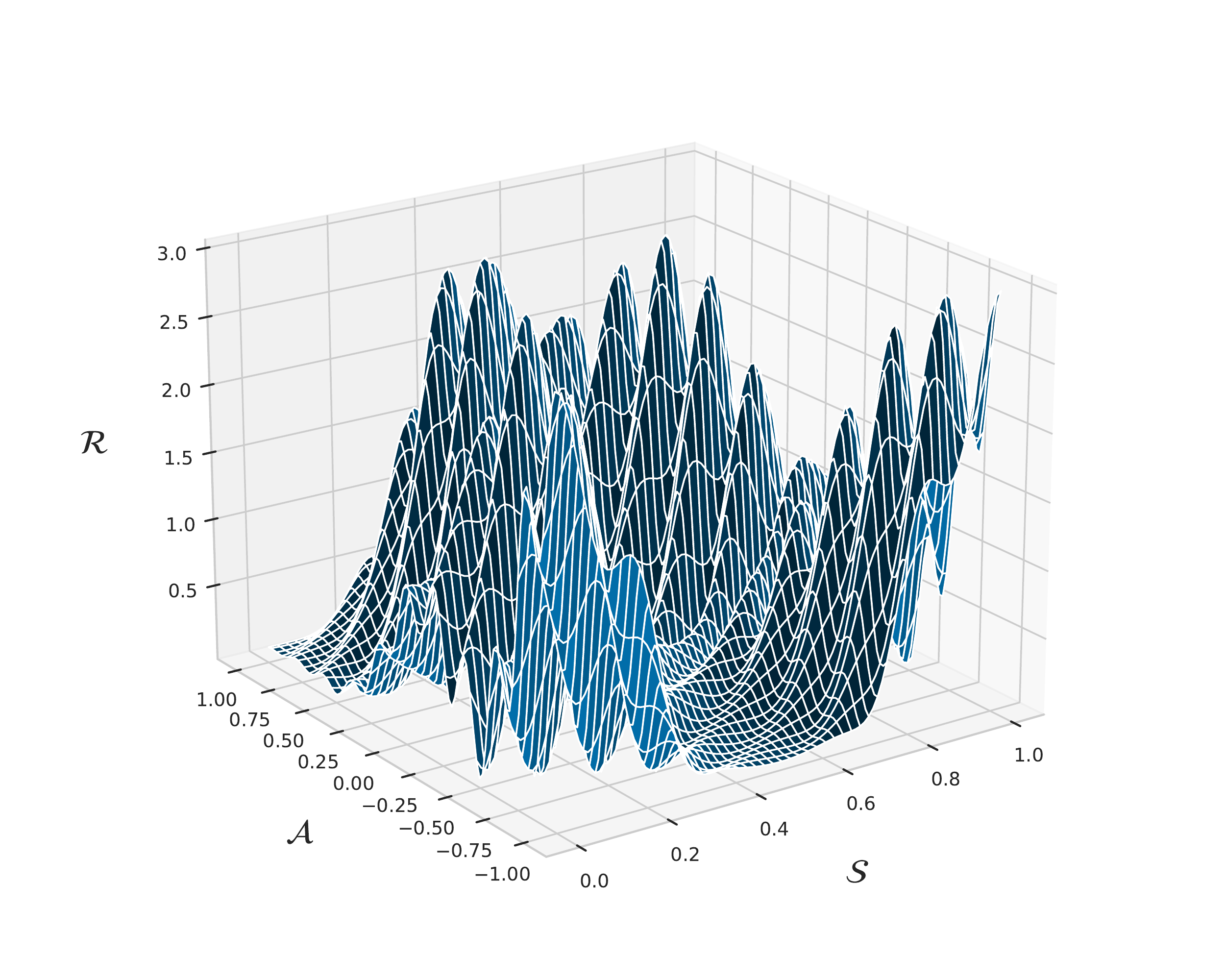}
    \includegraphics[width=.49\linewidth]{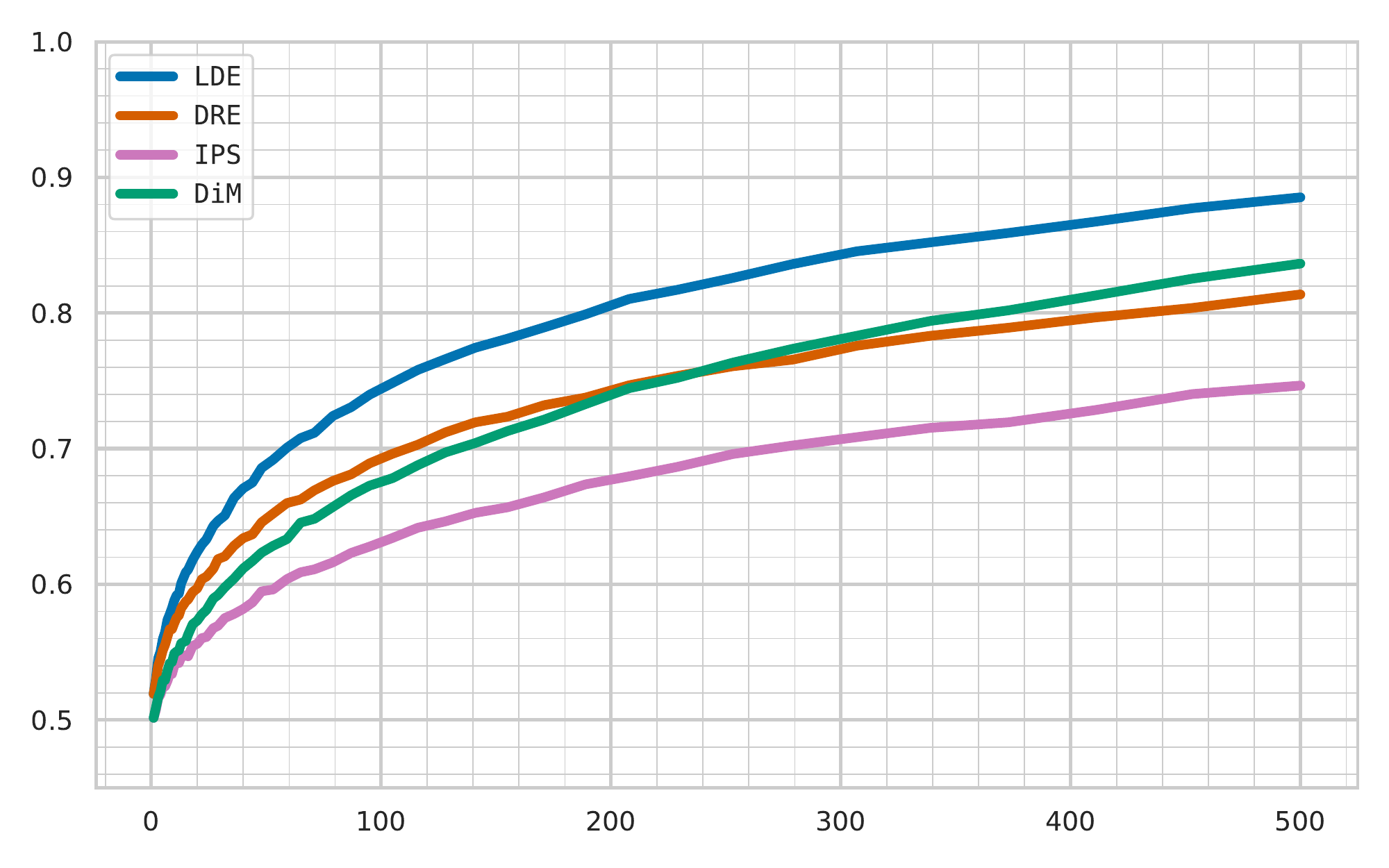}
    \caption{Reward function (left) and the percentage of correct per-state comparisons depending on the amount of historical data $m$ (right, $\alpha = .01$).}
    \label{fig:ex1.2_state}
\end{figure}

\begin{figure}[hbt!]
    \centering
    \includegraphics[width=.49\linewidth]{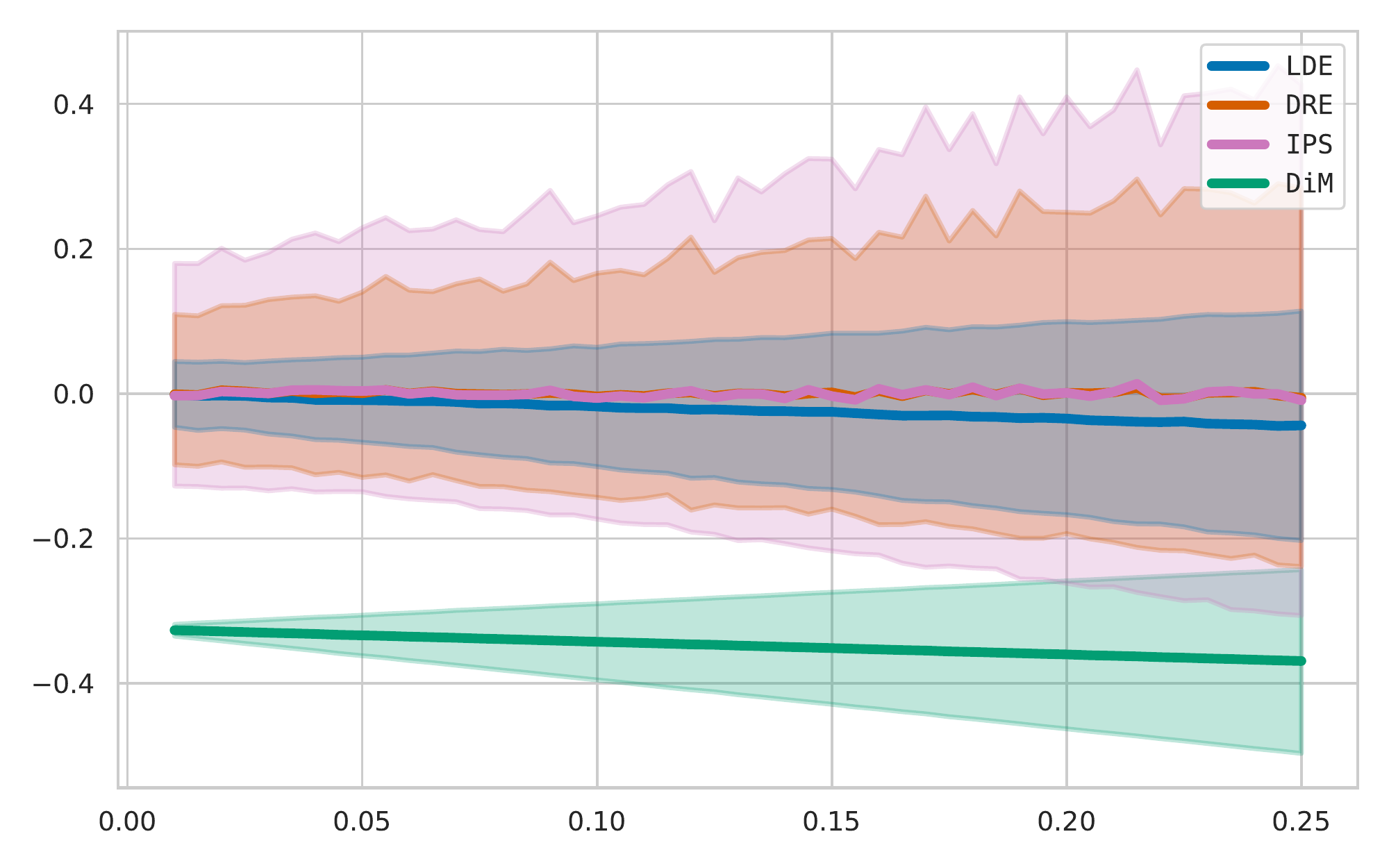}
    \includegraphics[width=.49\linewidth]{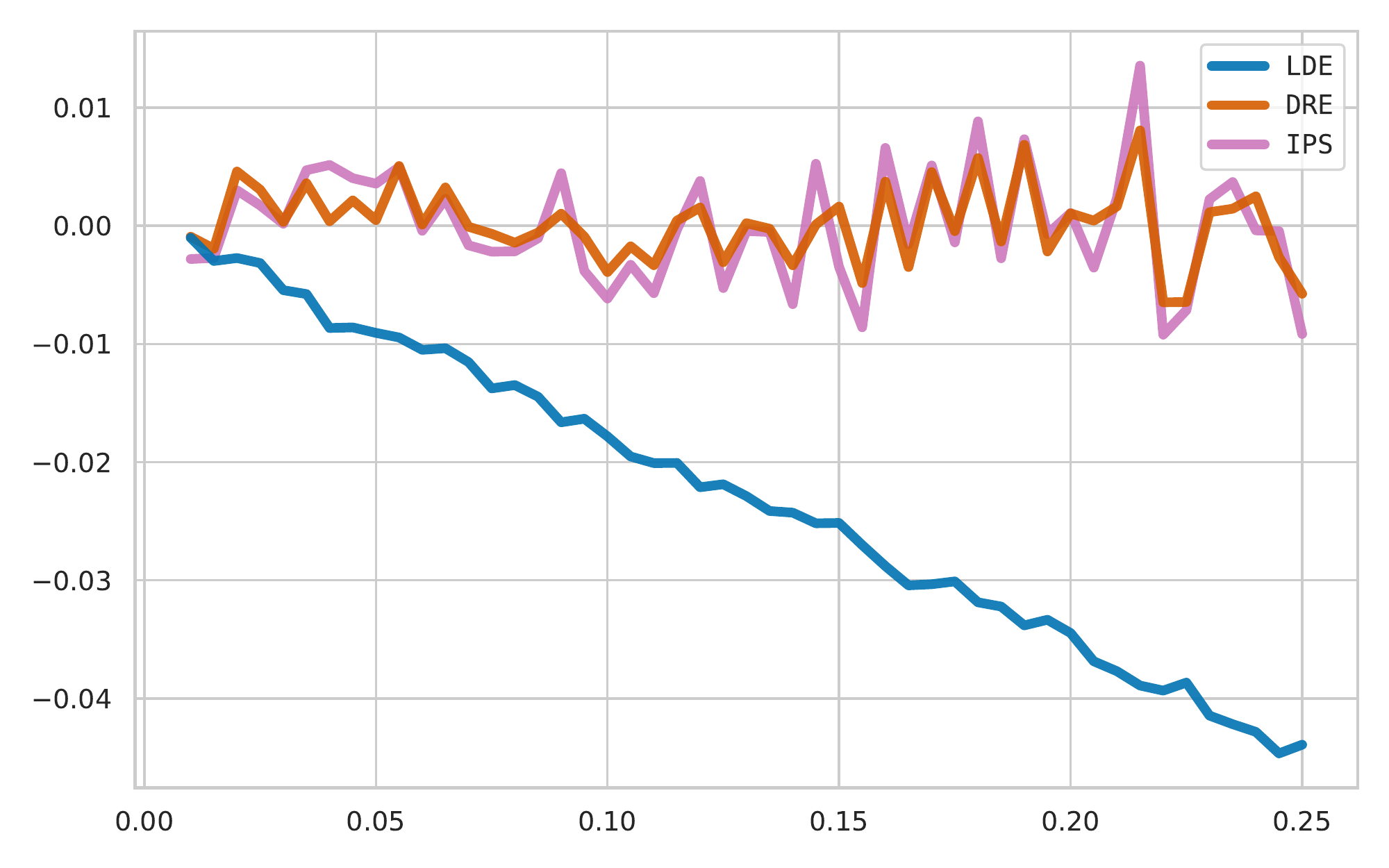}
    \\
    \includegraphics[width=.49\linewidth]{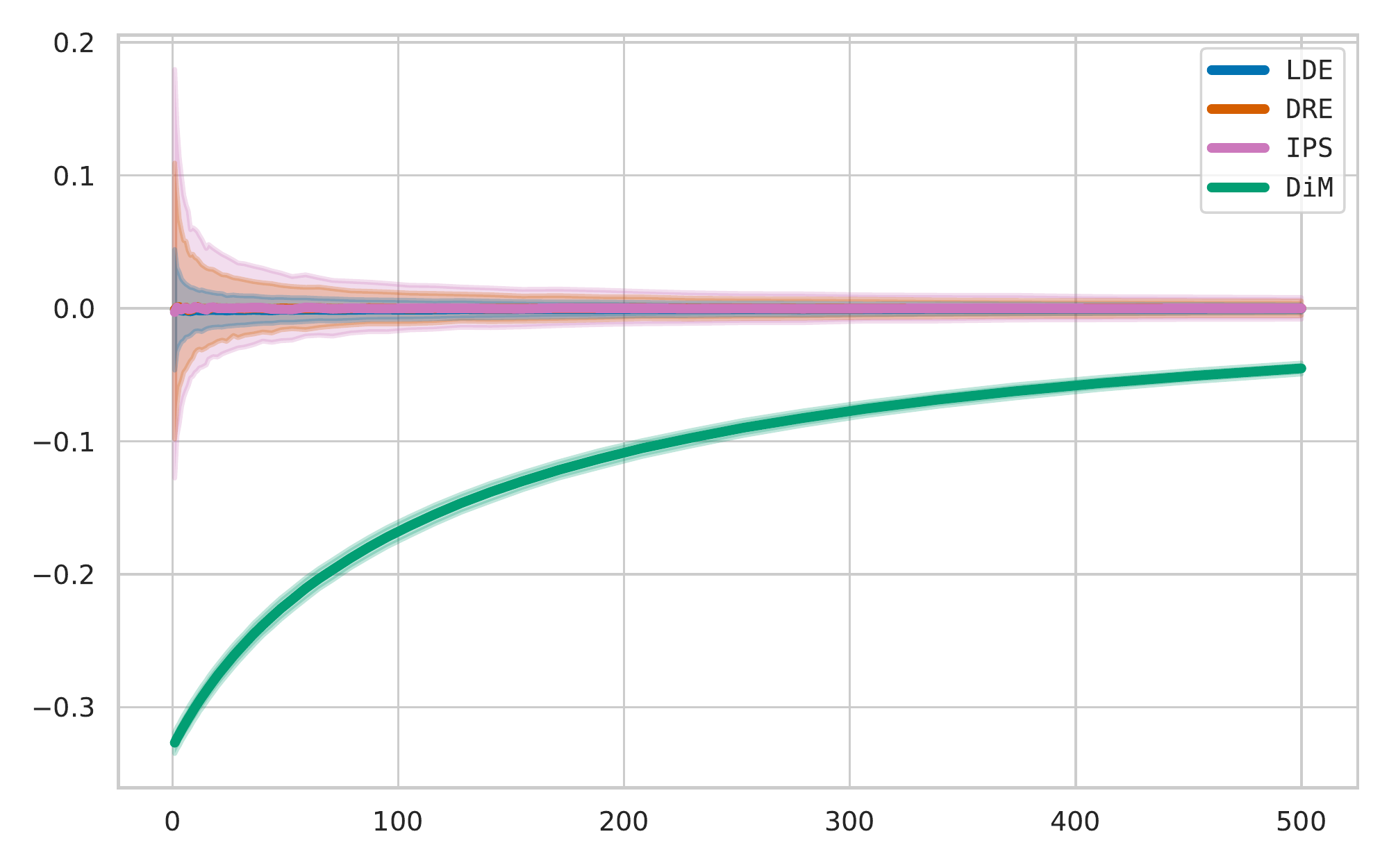}
    \includegraphics[width=.49\linewidth]{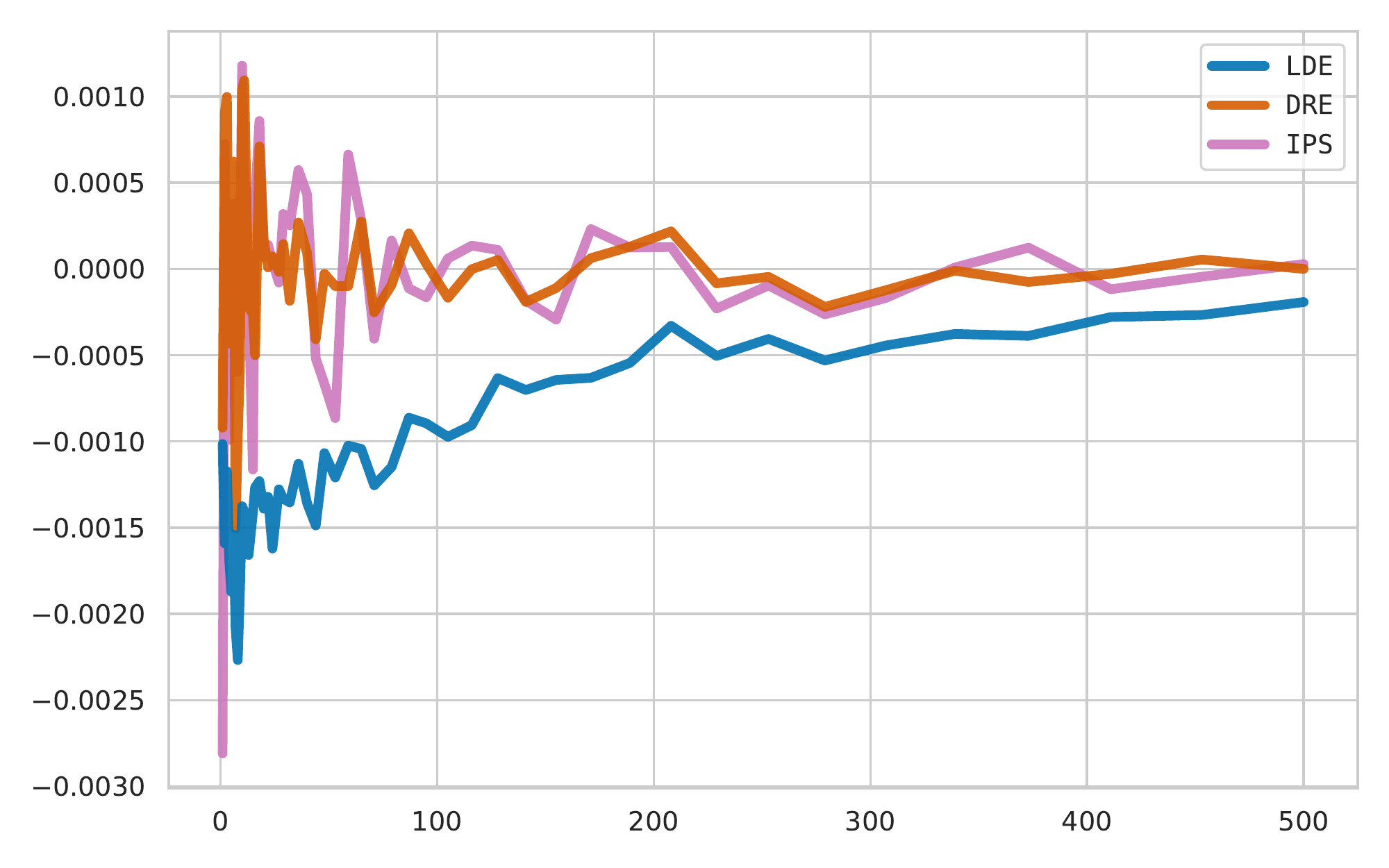}
    \caption{Policy evaluation error (normalized) depending on the value difference $\alpha$ (top, $m = 1$, see Table~\ref{tab:ex1.2_eval_a}) or the amount of historical data $m$ (bottom, $\alpha = .01$, see Table~\ref{tab:ex1.2_eval_m}).}
    \label{fig:ex1.2_eval}
\end{figure}

\begin{figure}[hbt!]
    \centering
    \includegraphics[width=.24\linewidth, trim={3em 3em 6em 3em}, clip]{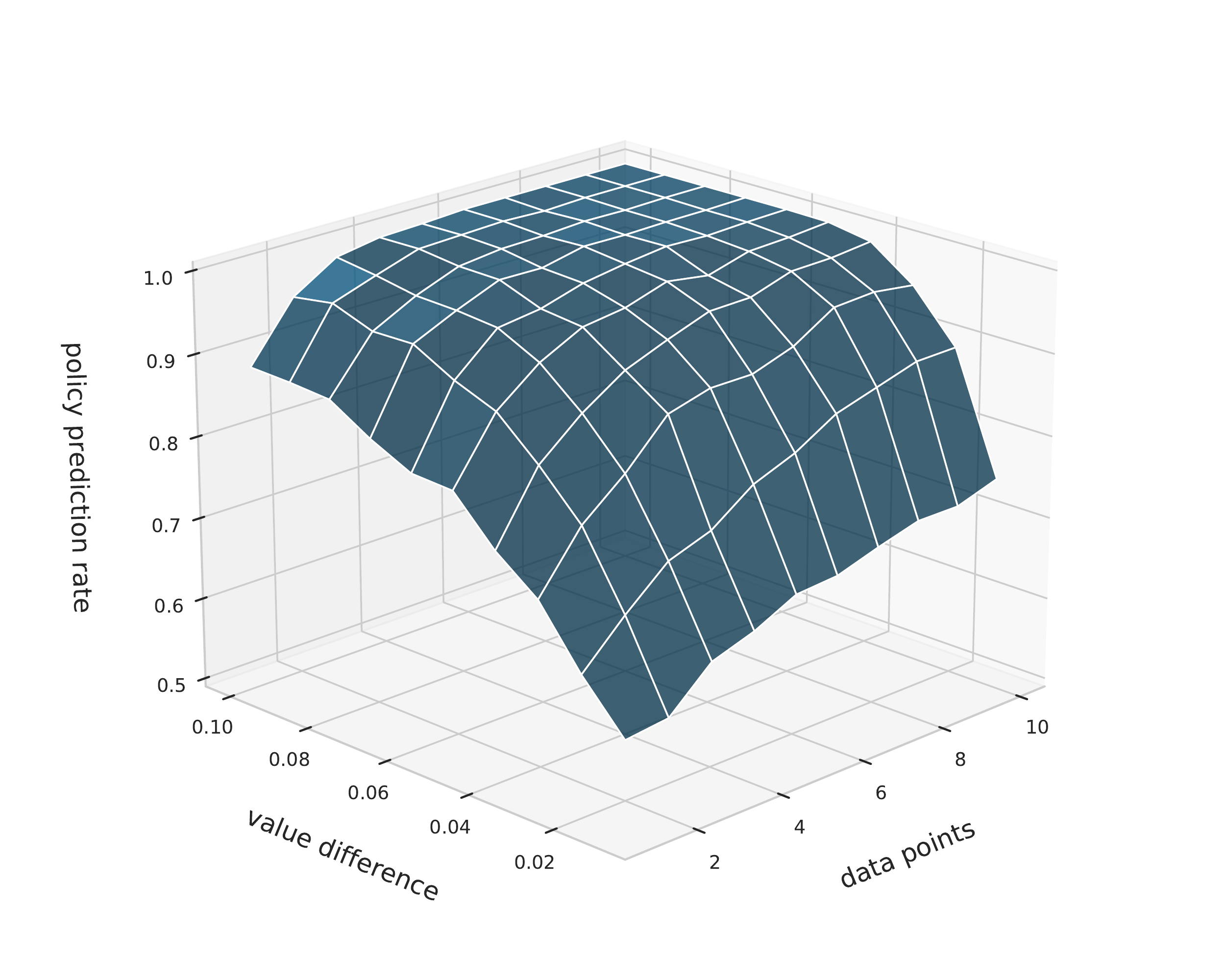}
    \includegraphics[width=.24\linewidth, trim={3em 3em 6em 3em}, clip]{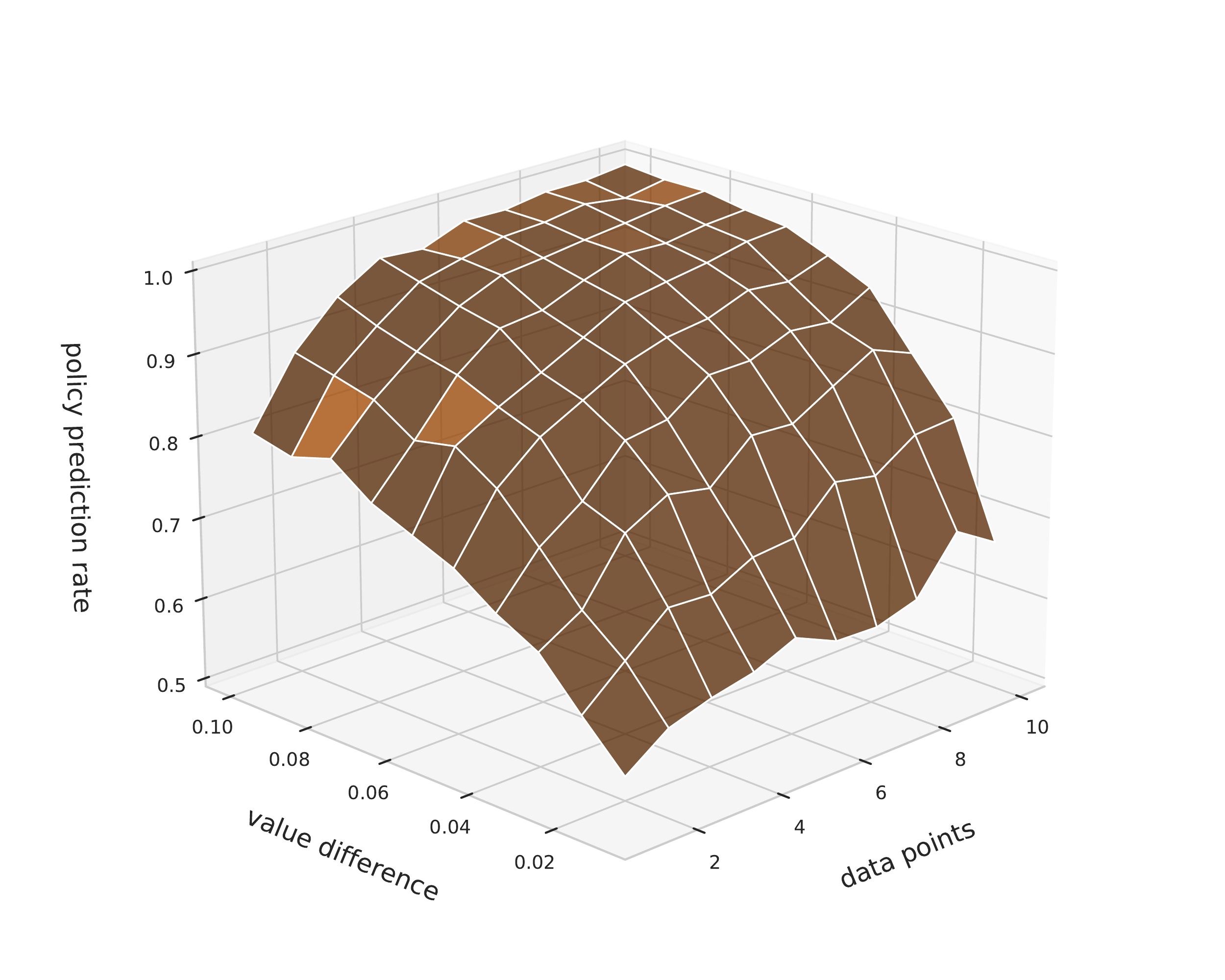}
    \includegraphics[width=.24\linewidth, trim={3em 3em 6em 3em}, clip]{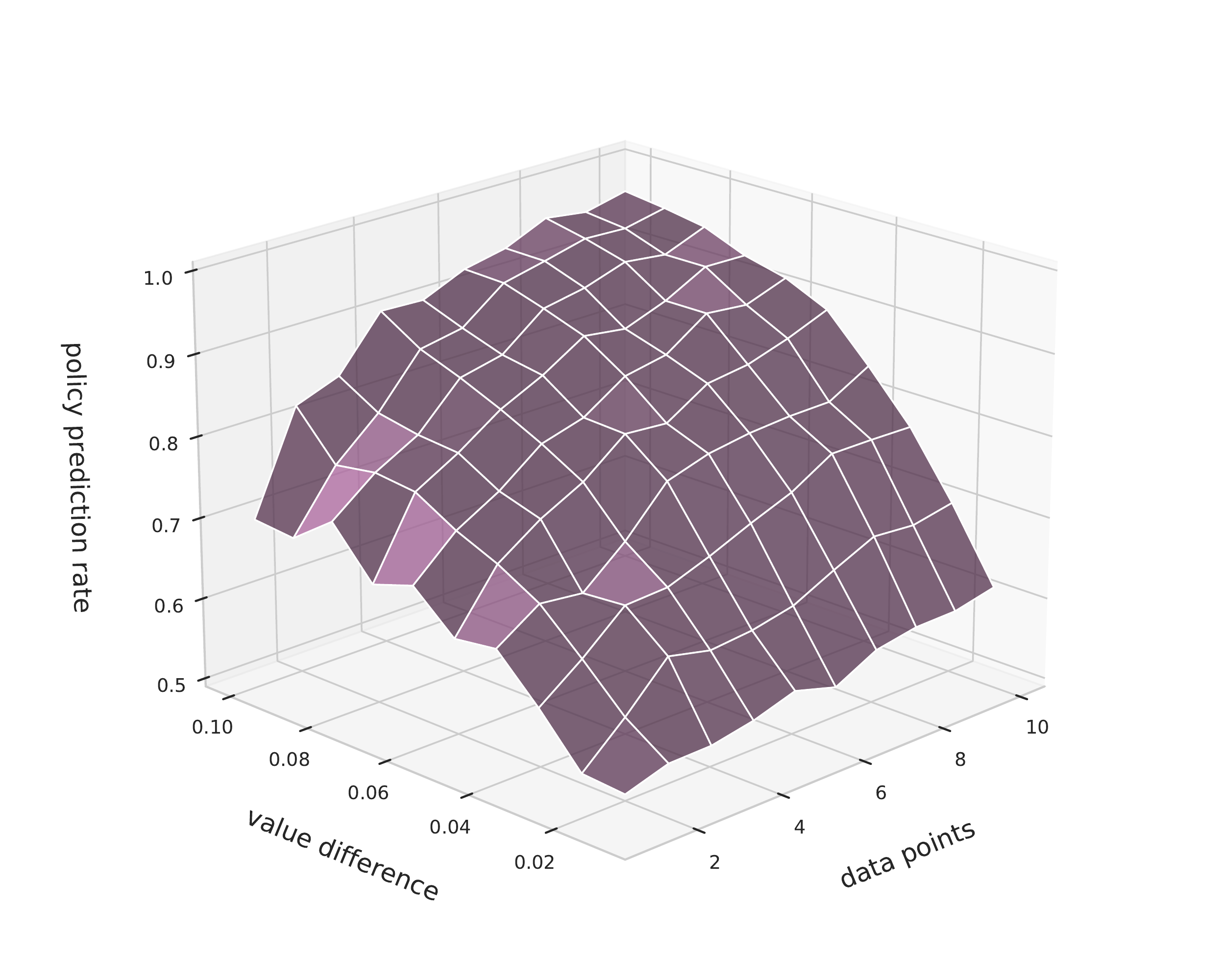}
    \includegraphics[width=.24\linewidth, trim={3em 3em 6em 3em}, clip]{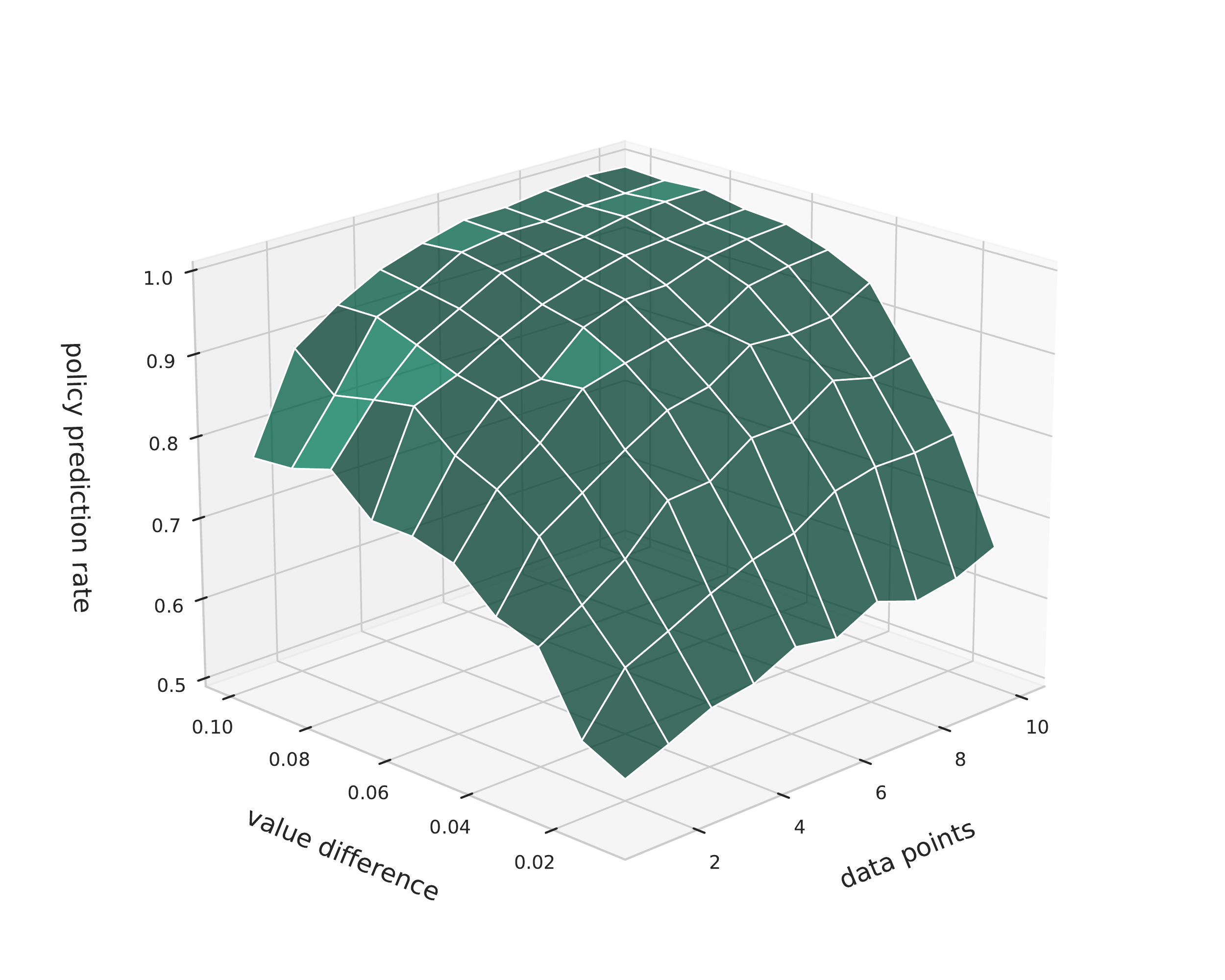}
    \\
    \includegraphics[width=.24\linewidth, trim={6em 3em 3em 3em}, clip]{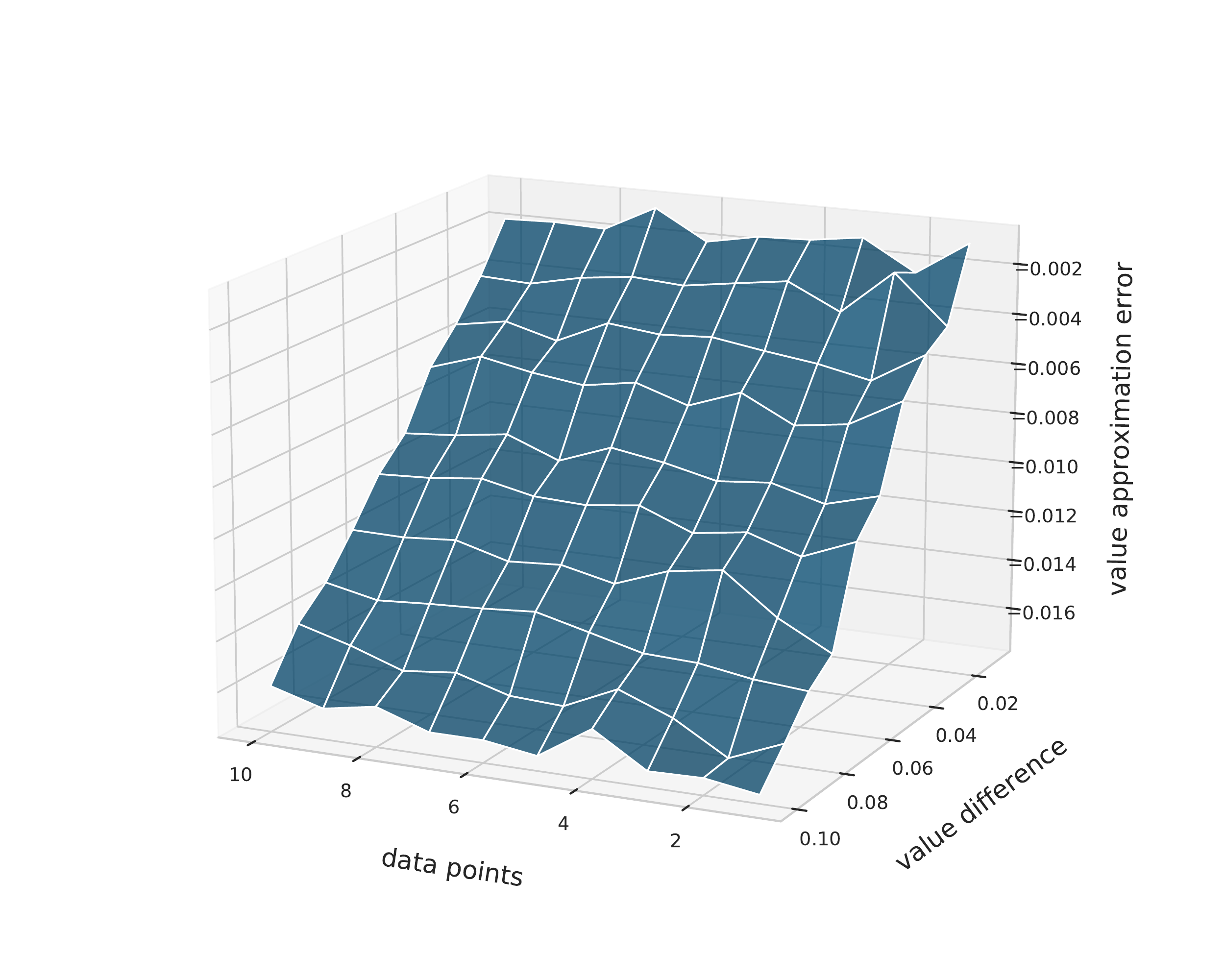}
    \includegraphics[width=.24\linewidth, trim={6em 3em 3em 3em}, clip]{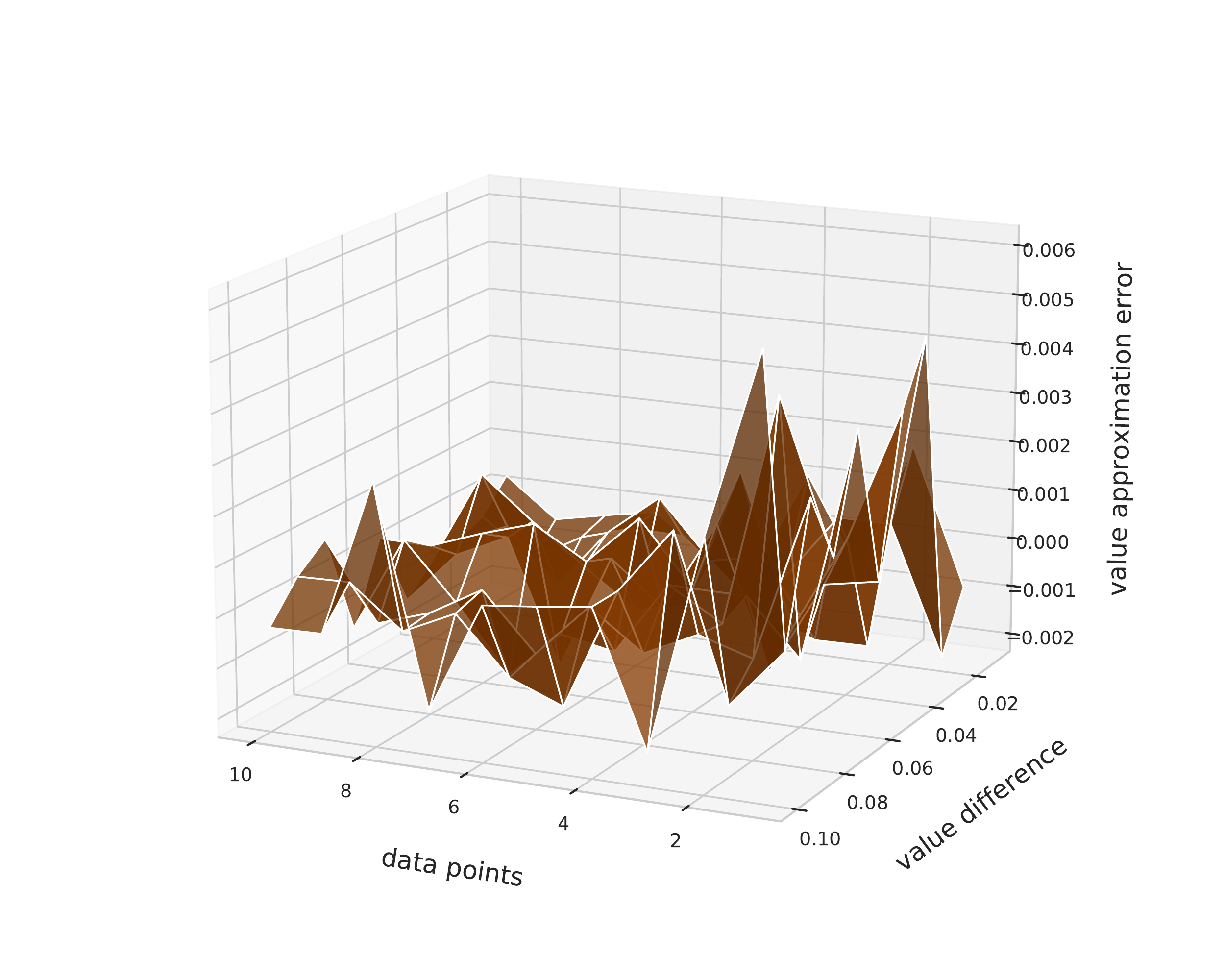}
    \includegraphics[width=.24\linewidth, trim={6em 3em 3em 3em}, clip]{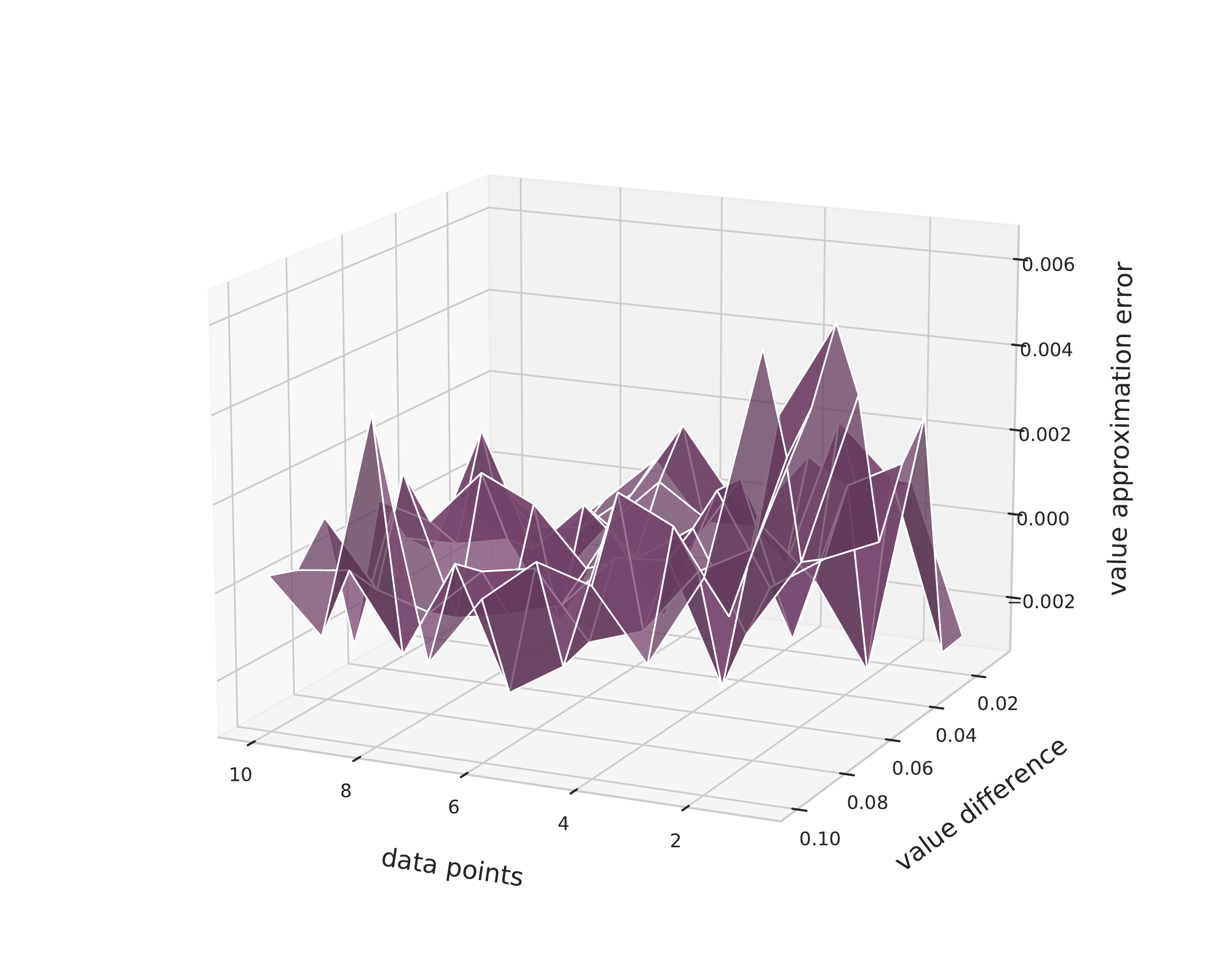}
    \includegraphics[width=.24\linewidth, trim={6em 3em 3em 3em}, clip]{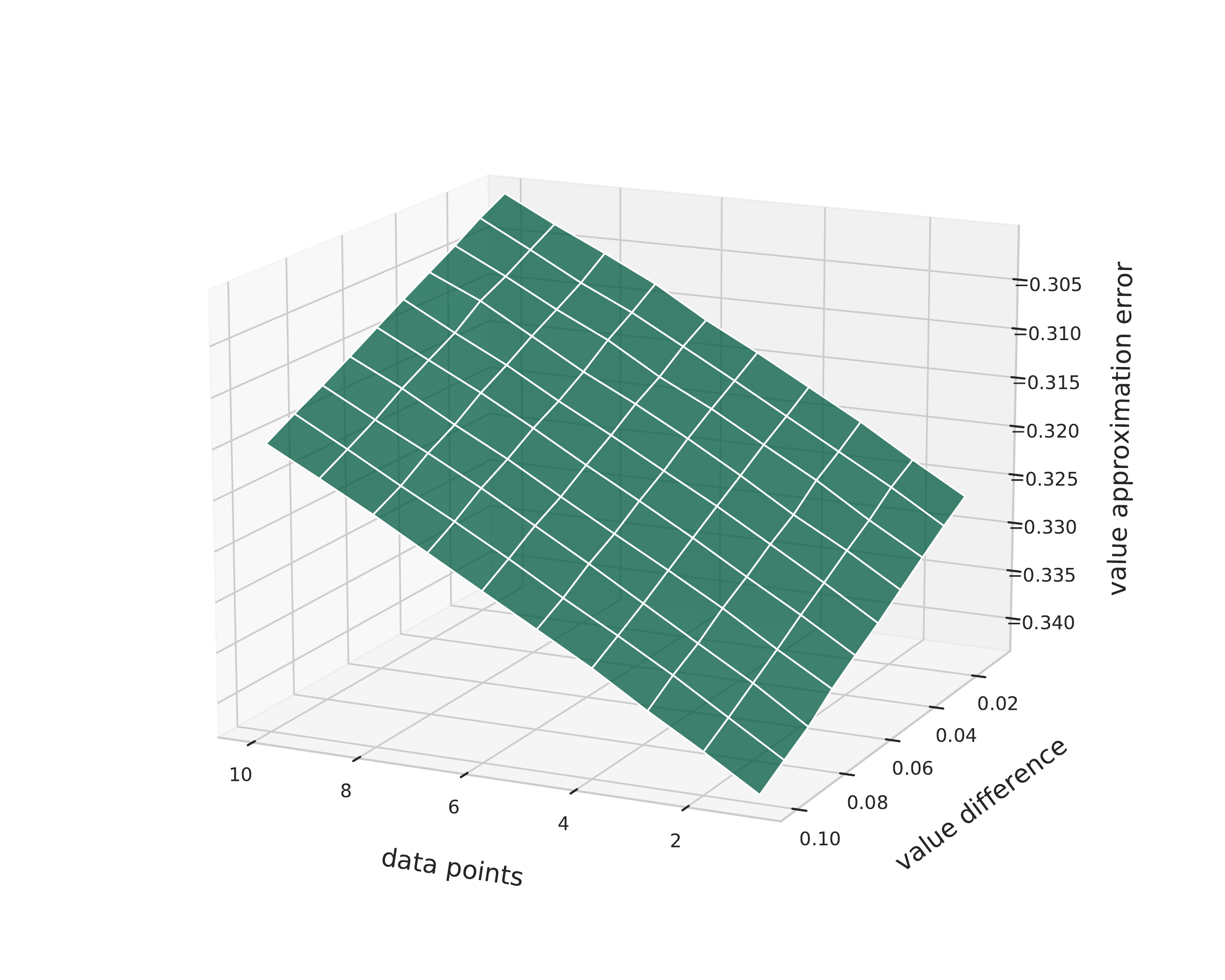}
    \caption{Policy comparison rate (top) and policy evaluation error (bottom) of the LDE, the DRE, the IPS, and the DiM respectively as a function of the value difference $\alpha$ and the number of data points $m$.}
    \label{fig:ex1.2_3d}
\end{figure}

\begin{figure}[hbt!]
    \centering
    \includegraphics[width=\linewidth, trim={4em 6em 8em 9em}, clip]{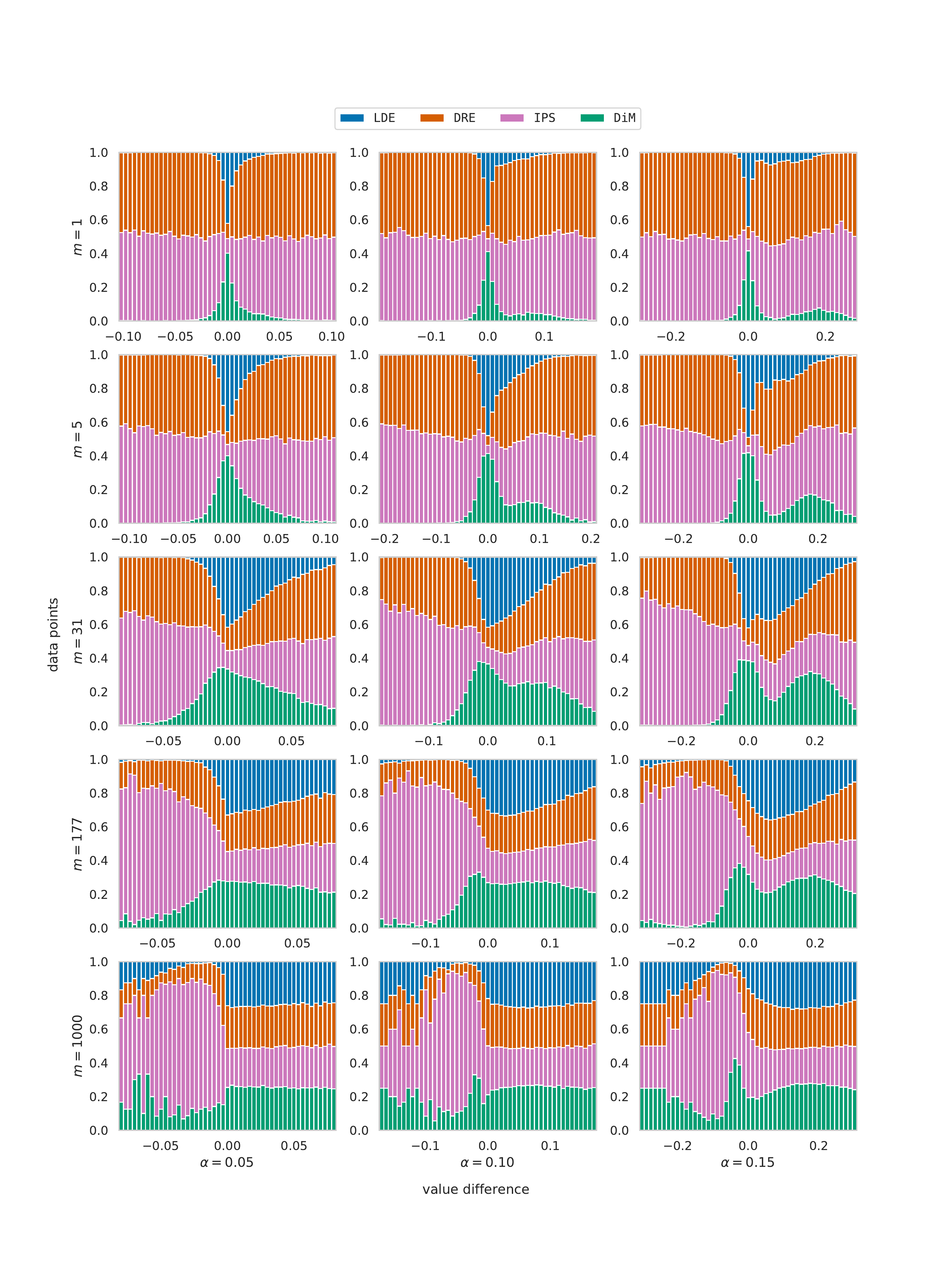}
    \caption{Normalized histograms of products of per-state differences with various values of $\alpha$ and $m$.}
    \label{fig:ex1.2_grid}
\end{figure}

This section contains supplementary figures for Example~1.2 from Section~\ref{sec:ex1}.
The reward function $r(s,a)$, given by~\eqref{eq:ex1.2_reward}, is shown in Figure~\ref{fig:ex1.2_state} (left).
The policy evaluation errors~\eqref{eq:eval_score}, depending on either the value difference $\alpha$, defined in~\eqref{eq:value_diff}, or the number of historical actions $m$, are shown in Figure~\ref{fig:ex1.2_eval}.
In Figure~\ref{fig:ex1.2_3d} we demonstrate the performance of the policy evaluation methods on the tasks of policy comparison (top) and evaluation (bottom) depending on the values of $\alpha$ and $m$.
The correlation between the differences of true values and the predicted estimates are presented in Table~\ref{tab:ex1.2_correlation}.
We verify the capacity of the LDE to perform per-state policy comparisons and show the per-state comparison rate depending on the number of data points $m$ in Figure~\ref{fig:ex1.2_state} (right) and the normalized histogram of the products of per-state differences in Figure~\ref{fig:ex1.2_grid}, as discussed in Section~\ref{sec:theory_appendix}.

We note that in this example, despite the outstanding performance on the policy comparison task (see Figure~\ref{fig:ex1.2_comp}), the LDE underperforms on the policy evaluation task (see Figure~\ref{fig:ex1.2_eval} and Tables~\ref{tab:ex1.2_eval_a} and~\ref{tab:ex1.2_eval_m}), likely due to the irregular reward distribution between the state-action pairs.

\begin{table}[hbt!]
    \centering\small
    \caption{Policy evaluation errors depending on the value difference $\alpha$, see Figure~\ref{fig:ex1.2_eval} (top).}
    \label{tab:ex1.2_eval_a}
    \begin{tabular}{ccccccccc}
        \toprule
        & \multicolumn{4}{c}{Approximation errors (average)} & \multicolumn{4}{c}{Approximation errors (std)}
        \\\cmidrule(lr){2-5}\cmidrule(lr){6-9}
        $\alpha$ & LDE & DRE & IPS & DiM & LDE & DRE & IPS & DiM
        \\\midrule
        0.05 & -9.06e-03 & \ \textbf{4.82e-04} & \ 3.57e-03 & -3.34e-01 & 3.65e-02 & 8.77e-02 & 1.28e-01 & 2.49e-02
        \\
        0.10 & -1.78e-02 & \textbf{-3.91e-03} & -6.16e-03 & -3.43e-01 & 5.59e-02 & 1.06e-01 & 1.47e-01 & 4.96e-02
        \\
        0.15 & -2.52e-02 & \ \textbf{1.63e-03} & -3.48e-03 & -3.51e-01 & 7.89e-02 & 1.40e-01 & 1.89e-01 & 7.43e-02
        \\
        0.20 & -3.45e-02 & \ \textbf{1.06e-03} & \ 1.08e-03 & -3.60e-01 & 1.02e-01 & 1.78e-01 & 2.50e-01 & 9.90e-02
        \\
        0.25 & -4.39e-02 & \textbf{-5.75e-03} & -9.15e-03 & -3.69e-01 & 1.27e-01 & 1.96e-01 & 2.60e-01 & 1.24e-01
        \\\bottomrule
    \end{tabular}
\end{table}

\begin{table}[hbt!]
    \centering\small
    \caption{Policy evaluation errors depending on the number of data points $m$, see Figure~\ref{fig:ex1.2_eval} (bottom).}
    \label{tab:ex1.2_eval_m}
    \begin{tabular}{ccccccccc}
        \toprule
        & \multicolumn{4}{c}{Approximation errors (average)} & \multicolumn{4}{c}{Approximation errors (std)}
        \\\cmidrule(lr){2-5}\cmidrule(lr){6-9}
        $m$ & LDE & DRE & IPS & DiM & LDE & DRE & IPS & DiM
        \\\midrule
        10 & -1.38e-03 & \ \textbf{1.04e-03} & \ 1.18e-03 & -3.00e-01 & 9.74e-03 & 2.37e-02 & 3.52e-02 & 5.38e-03
        \\
        27 & -1.28e-03 & \textbf{-1.72e-05} & -7.90e-05 & -2.62e-01 & 7.11e-03 & 1.34e-02 & 2.09e-02 & 5.15e-03
        \\
        71 & -1.26e-03 & \textbf{-2.52e-04} & -4.04e-04 & -1.97e-01 & 5.11e-03 & 8.59e-03 & 1.29e-02 & 4.68e-03
        \\
        189 & -5.45e-04 & \ 1.29e-04 & \ \textbf{1.24e-04} & -1.13e-01 & 3.28e-03 & 5.30e-03 & 7.82e-03 & 3.82e-03
        \\
        500 & -1.92e-04 & \ \textbf{4.63e-07} & \ 2.90e-05 & -4.52e-02 & 1.76e-03 & 3.31e-03 & 4.99e-03 & 2.51e-03
        \\\bottomrule
    \end{tabular}
\end{table}

\begin{table}[hbt!]
    \centering\small
    \caption{Correlation statistics computed on sequences~\eqref{eq:correlation} from Figure~\ref{fig:ex1.2_3d}.}
    \label{tab:ex1.2_correlation}
    \begin{tabular}{cccccccccc}
        \toprule
        & \multicolumn{4}{c}{Pearson's coefficient} & \multicolumn{4}{c}{Spearman's coefficient}
        \\\cmidrule(lr){2-5}\cmidrule(lr){6-9}
        $m$ & LDE & DRE & IPS & DiM & LDE & DRE & IPS & DiM
        \\\midrule
        1 & \textbf{0.3992} & 0.2549 & 0.1849 & 0.2920 & \textbf{0.3859} & 0.2714 & 0.1800 & 0.2695
        \\
        2 & \textbf{0.5215} & 0.3476 & 0.2614 & 0.3928 & \textbf{0.5203} & 0.3586 & 0.2577 & 0.3795
        \\
        3 & \textbf{0.6008} & 0.4029 & 0.2865 & 0.4514 & \textbf{0.6065} & 0.4121 & 0.2868 & 0.4449
        \\
        4 & \textbf{0.6683} & 0.4733 & 0.3476 & 0.5246 & \textbf{0.6771} & 0.4923 & 0.3499 & 0.5193
        \\
        5 & \textbf{0.7021} & 0.5062 & 0.3748 & 0.5546 & \textbf{0.7139} & 0.5250 & 0.3772 & 0.5521
        \\
        6 & \textbf{0.7340} & 0.5598 & 0.4269 & 0.5995 & \textbf{0.7464} & 0.5838 & 0.4336 & 0.6025
        \\
        7 & \textbf{0.7583} & 0.5838 & 0.4355 & 0.6207 & \textbf{0.7714} & 0.6061 & 0.4436 & 0.6251
        \\
        8 & \textbf{0.7791} & 0.6075 & 0.4706 & 0.6578 & \textbf{0.7917} & 0.6292 & 0.4798 & 0.6635
        \\
        9 & \textbf{0.7957} & 0.6153 & 0.4812 & 0.6750 & \textbf{0.8086} & 0.6419 & 0.4932 & 0.6807
        \\
        10 & \textbf{0.8094} & 0.6452 & 0.5056 & 0.6881 & \textbf{0.8240} & 0.6691 & 0.5188 & 0.6953
        \\\bottomrule
    \end{tabular}
\end{table}

\clearpage
\subsection{Example 2: Classification}\label{sec:ex2_appendix}
In this experiment we reframe the problem of policy comparison and evaluation of classifiers as an evaluation/comparison task in a contextual bandit environment under the limited data setting, which closely follows the setup from~\citep[Section 4.1]{dudik2014doubly}.
We consider various datasets from the UCI repository~\citep{asuncion2007uci}, deploy the policy evaluation methods on this modified task, and assess their performance.

Let $\mathcal{X}$ be the set of input data, and $\mathcal{K} := \{1, \ldots, K\}$ 
be the associated set of labels.
A classifier is a function $\pi: \mathcal{X} \rightarrow \mathcal{K}$.
In the contextual bandit formulation we view $\mathcal{X}$ as the state (context) space, $\mathcal{K}$ as the action space, and the classifier $\pi$ as a policy.
The reward space is $\mathcal{R} = \{0,1\}$ and the received reward is $1$ if the classification is correct, and $0$ otherwise.
An agent-environment interaction under a policy $\pi: \mathcal{X} \to \mathcal{K}$ is simulated by observing a state $s \in \mathcal{X}$, taking an action $a = \operatorname{argmax} \pi(s) \in \mathcal{K}$, and receiving a reward $r(s,a) = \mathbbm{1}\{a = k_s\}$, where $k_s \in \mathcal{K}$ is the true label for the element $s \in \mathcal{X}$.

By interacting with the environment in such a way, the historical data $\mathcal{D}$ is collected and the task of classifier evaluation or comparison transfers naturally to the contextual bandit setting where the policy evaluation methods can be deployed.
In our experiment we compare and evaluate $10$ target policies.
The comparison score is computed via~\eqref{eq:comp_score} and the evaluation error is the difference between the predicted and the true value, i.e. $V_{method}(\pi) - V(\pi)$.
For both metrics we report the average and the standard deviation over 1,000 tests.
Concretely, for a given setting a single test consists of the following steps:
\begin{enumerate}
    \item Randomly split the input data $\mathcal{X}$ into equally-sized disjoint subsets $\mathcal{X}_{tr}$ and $\mathcal{X}_{ts}$ to serve as the training and test sets respectively (the test and training sets are guaranteed to have at least one instance of each class unless there is only one instance of that class in the entire dataset);
    \item Train the target policies $\{\pi_1, \ldots, \pi_{10}\}$ on the training set $\mathcal{X}_{tr}$ using the Direct Loss Minimization~\citep{mcallester2010direct} over 1,000 epochs with $\varepsilon = 0.1$;
    \item Sample a behavioral policy $\nu$ from the normal distribution and then normalize with the softmax function;
    \item Collect the historical data $\mathcal{D}$ by deploying the behavioral policy $\nu$ on the training set $\mathcal{X}_{ts}$ so that each test example is observed only once;
    \item Employ the policy evaluation methods to evaluate and compare the target policies $\{\pi_1, \dots, \pi_{10}\}$ on the set of historical data $\mathcal{D}$.
\end{enumerate}

For each dataset from Table~\ref{tab:ex2_datasets} we repeat tests 1,000 times in the following way: steps 1--2 are performed 10 times, and for each split of the input data $\mathcal{X}$ and a set of target policies $\{\pi_1, \ldots, \pi_{10}\}$, the steps 3--4 are executed 100 times, i.e. each collection of trained policies is compared and evaluated 100 times under different instances of historical data $\mathcal{D}$.

\begin{figure}[hbt!]
    \centering
    \includegraphics[width=\linewidth]{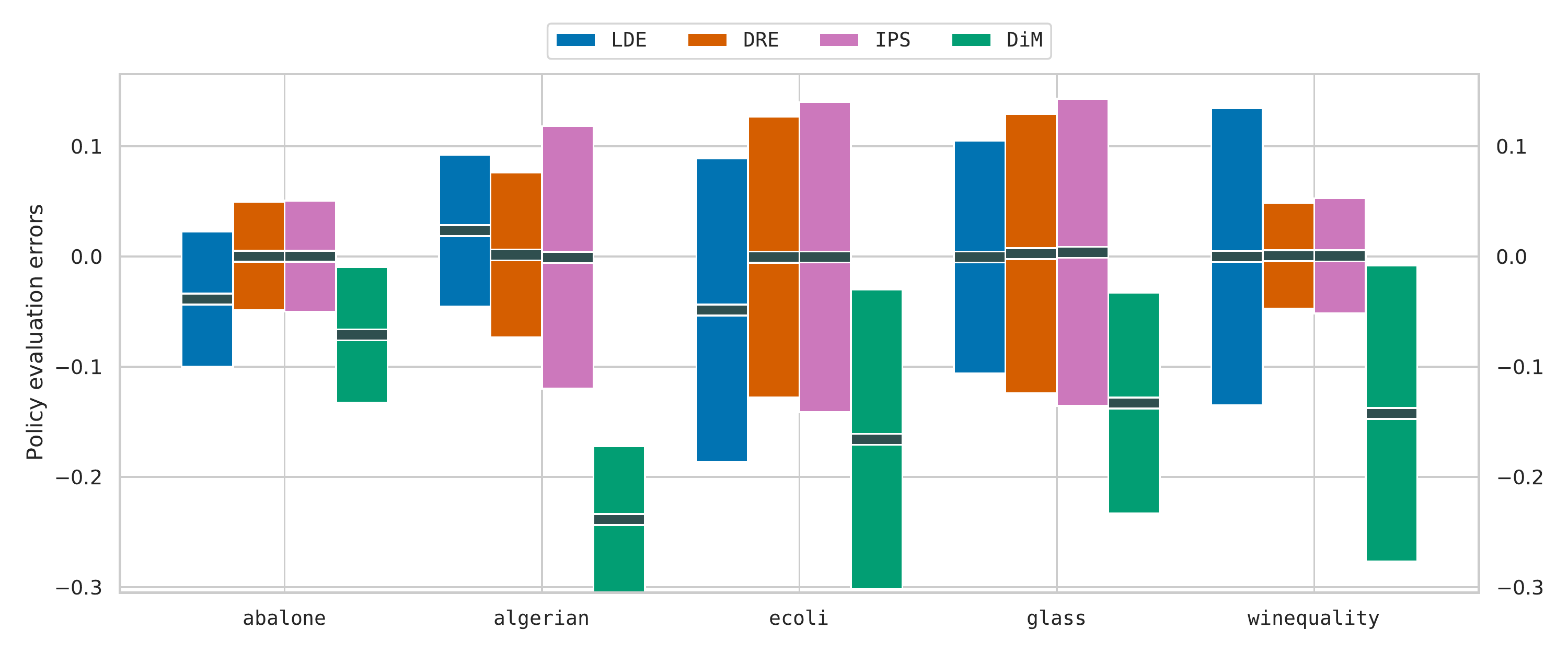}
    \caption{Policy evaluation errors on various datasets, see Table~\ref{tab:ex2_eval}.}
    \label{fig:ex2_eval}
\end{figure}

The statistical distribution of the policy comparison scores is presented in Figure~\ref{fig:ex2_comp} and Table~\ref{tab:ex2_comp}.
The evaluation results in the form of the value approximation errors are shown in Figure~\ref{fig:ex2_eval} and Table~\ref{tab:ex2_eval}.
As one would expect, even though the LDE performs well on the policy comparison task, the value estimate predictions are not precise due to the complexity of this scenario.

Our experiment is closely related to the one considered in~\citep{dudik2014doubly}; however, there are three mild differences.
First, we consider the policy comparison task in addition to the policy evaluation task, and thus we train 10 target policies for each test, as opposed to the single target policy in the original experiment.
Second, for the purpose of nontrivial comparison, we require the target policies to be distinct, and thus each target policy is trained on a subset of the training data $\mathcal{X}_{tr}$, consisting of $80\%$ randomly sampled training examples.
Third, we sample the behavioral policy $\nu$ from the normal distribution, rather than utilizing a uniformly random policy, since otherwise the policy comparison results coincide for the LDE and the DRE, and the IPS and the DiM.
Since we consider a different behavior policy, the policy evaluation errors reported in the current paper differ from the ones in the original experiment.

\begin{table}[hbt!]
    \centering\small
    \caption{Datasets considered in Example~\ref{sec:ex2}.}
    \label{tab:ex2_datasets}
    \begin{tabular}{lccccc}
        \toprule
        & \texttt{abalone} & \texttt{algerian} & \texttt{ecoli} & \texttt{glass} & \texttt{winequality}
        \\\midrule
        Size & 4177 & 244 & 336 & 214 & 1599
        \\
        Input features & 8 & 11 & 8 & 10 & 11
        \\
        Classes & 29 & 2 & 8 & 7 & 10
        \\\bottomrule
    \end{tabular}
\end{table}

\begin{table}[hbt!]
    \centering\small
    \caption{Policy comparison scores on classification datasets, computed via~\eqref{eq:comp_score}, see Figure~\ref{fig:ex2_comp}.}
    \label{tab:ex2_comp}
    \begin{tabular}{lcccc}
        \toprule
        & \multicolumn{4}{c}{Policy comparison score: average (std)}
        \\\cmidrule(lr){2-5}
        Dataset & LDE & DRE & IPS & DiM
        \\\midrule
        \texttt{abalone} & \textbf{0.8753} (0.057) & 0.8402 (0.071) & 0.8402 (0.072) & 0.8693 (0.059)
        \\
        \texttt{algerian} & \textbf{0.9976} (0.009) & 0.9240 (0.079) & 0.8447 (0.112) & 0.8964 (0.087)
        \\
        \texttt{ecoli} & \textbf{0.8859} (0.073) & 0.8272 (0.097) & 0.8405 (0.094) & 0.8828 (0.074)
        \\
        \texttt{glass} & \textbf{0.8355} (0.080) & 0.7738 (0.107) & 0.7725 (0.107) & 0.8279 (0.079)
        \\
        \texttt{winequality} & 0.9328 (0.044) & 0.9009 (0.061) & 0.9258 (0.048) & \textbf{0.9359} (0.043)
        \\\bottomrule
    \end{tabular}
\end{table}

\begin{table}[hbt!]
    \centering\small
    \caption{Policy evaluation errors on classification datasets, see Figure~\eqref{fig:ex2_eval}.}
    \label{tab:ex2_eval}
    \begin{tabular}{lcccc}
        \toprule
        & \multicolumn{4}{c}{Policy evaluation error: average (std)}
        \\\cmidrule(lr){2-5}
        Dataset & LDE & DRE & IPS & DiM
        \\\midrule
        \texttt{abalone} & -3.85e-02 (0.061) & \ 3.87e-04 (0.049) & \ \textbf{3.37e-04} (0.050) & -7.10e-02 (0.061)
        \\
        \texttt{algerian} & \ 2.36e-02 (0.069) & \ 1.43e-03 (0.075) & \textbf{-7.45e-04} (0.119) & -2.39e-01 (0.066)
        \\
        \texttt{ecoli} & -4.85e-02 (0.138) & -5.28e-04 (0.127) & \textbf{-4.85e-04} (0.141) & -1.66e-01 (0.136)
        \\
        \texttt{glass} & \textbf{-4.96e-04} (0.106) & \ 2.70e-03 (0.127) & \ 3.89e-03 (0.139) & -1.33e-01 (0.100)
        \\
        \texttt{winequality} & \textbf{-5.24e-06} (0.135) & \ 7.90e-04 (0.048) & \ 6.57e-04 (0.052) & -1.42e-01 (0.134)
        \\\bottomrule
    \end{tabular}
\end{table}

\clearpage
\subsection{Example 3: RL Environments}\label{sec:ex3_appendix}
In this example we reframe the problem of policy comparison in some commonly used reinforcement learning environments as a policy comparison task in a contextual bandit setting.
Specifically, we consider the RL environments implemented in the \texttt{PyBullet}~(\url{https://pybullet.org/}) engine.
For each environment we train the target policies using the \texttt{Stable-Baselines3}~\citep{stable-baselines3} repository and compare the obtained policies by deploying the policy evaluation methods under the limited historical data setting.

In this setting an RL environment is the tuple $(\mathcal{S}_{rl}, \mathcal{A}_{rl}, \mathcal{R}_{rl})$, and for an agent $\pi$ an interaction with the environment is an iterative process consisting of a various number of timesteps.
On each timestep $t \ge 0$ the agent observes a state $s_t^{rl} \in \mathcal{S}_{rl}$, takes an action $a_t^{rl} \in \mathcal{A}_{rl}$, and receives a reward $r_t^{rl} \in \mathcal{R}_{rl}$.
This process is continued until termination, either by the environment or by the user.
The full process from beginning to termination forms an episode $\{(s_t^{rl}, a_t^{rl}, r_t^{rl})\}_{t=0}^T$ with $0 \le T \le \infty$.

To reframe this setting as a contextual bandit, we limit the episode length to 100 timesteps and view the whole episode as a single agent-environment interaction.
Namely, given a recorded episode $\{(s_t^{rl}, a_t^{rl}, r_t^{rl})\}_{t=0}^{99}$, we view it as an interaction with the bandit environment as follows:
\begin{itemize}
    \item The observed state is $s := s_0^{rl} \in \mathcal{S}_{rl}$;
    \item The taken action is $a := \{a_0^{rl}, \ldots, a_{99}^{rl}\} \in \mathcal{A}_{rl}^{100}$;
    \item The received reward is $r(s,a) := \sum_{t=0}^{99} r_t^{rl}$.
\end{itemize}
In order to establish the probability of the policy $\pi$ taking an action $a$ from a state $s$, i.e. following the trajectory $\{a_0^{rl}, \ldots, a_{99}^{rl}\}$ from the state $s_0^{rl}$, we employ the following \textit{probability proxy}:
\begin{equation}\label{eq:ex3_prob_proxy}
    \pi(a|s) = \prod_{t=0}^{99} \exp \left( 1 - \left( 1 - \Bigg(\frac{\pi(s_t^{rl}) - a_t^{rl}}{a_{high} - a_{low}}\Bigg)^2 \right)^{-1} \right)
\end{equation}
where $a_{high}$ and $a_{low}$ are the bounds of the action space $\mathcal{A}$ for the given environment.
In cases when the action space is multi-dimensional, the probability proxy is computed coordinate-wise and then averaged.
We also note that if an episode terminated before timestep 100, the action sequence, the cumulative reward, and the probability proxy are adjusted accordingly.

As a result, any set of logged interactions with the reinforcement learning environment can be treated as the historical data in a bandit setting and the policy comparison can be performed.
In this experiment we consider 5 target policies that are trained using different optimization algorithms:
\begin{itemize}
    \item $\pi_{a2c}$ is trained with Advantage Actor Critic (A2C)~\citep{mnih2016asynchronous};
    \item $\pi_{ddpg}$ is trained with Deep Deterministic Policy Gradient (DDPG)~\citep{lillicrap2015continuous};
    \item $\pi_{ppo}$ is trained with Proximal Policy Optimization (PPO)~\citep{schulman2017proximal};
    \item $\pi_{sac}$ is trained with Soft Actor Critic (SAC)~\citep{haarnoja2018soft};
    \item $\pi_{td3}$ is trained with Twin Delayed DDPG (TD3)~\citep{fujimoto2018addressing}.
\end{itemize}

Thus in our setting a policy comparison test in a given environment consists of the following steps:
\begin{enumerate}
    \item The target policies $\{\pi_{a2s}, \pi_{ddpg}, \pi_{ppo}, \pi_{sac}, \pi_{td3}\}$ are trained for 1,000 epoches with the default values of hyperparameters, as implemented in \texttt{Stable-Baselines3};
    \item Each target policy is deployed once and the recorded interactions of up to 100 timesteps form the historical dataset $\mathcal{D}$;
    \item The policy evaluation methods (the LDE, the DRE, the IPS, and the DiM) are deployed to compare the target policies $\{\pi_{a2s}, \pi_{ddpg}, \pi_{ppo}, \pi_{sac}, \pi_{td3}\}$ under the historical data $\mathcal{D}$ and the probability proxy~\eqref{eq:ex3_prob_proxy}.
\end{enumerate}

\begin{figure}[hbt!]
    \centering
    \includegraphics[width=.49\linewidth]{./images/scores_InvertedPendulumBulletEnv-v0.pdf}
    \includegraphics[width=.49\linewidth]{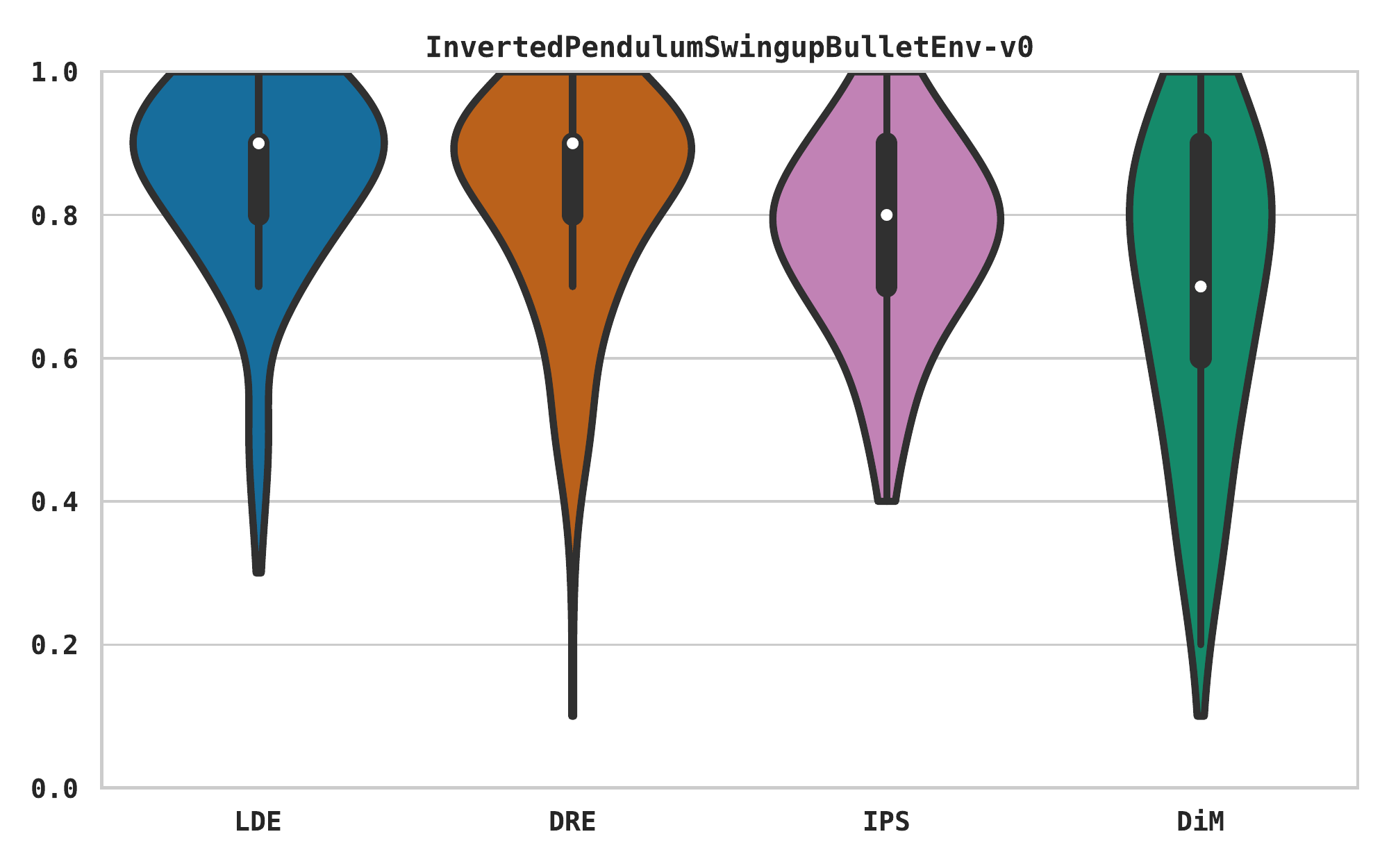}
    \\
    \includegraphics[width=.49\linewidth]{./images/scores_ReacherBulletEnv-v0.pdf}
    \includegraphics[width=.49\linewidth]{./images/scores_Walker2DBulletEnv-v0.pdf}
    \\
    \includegraphics[width=.49\linewidth]{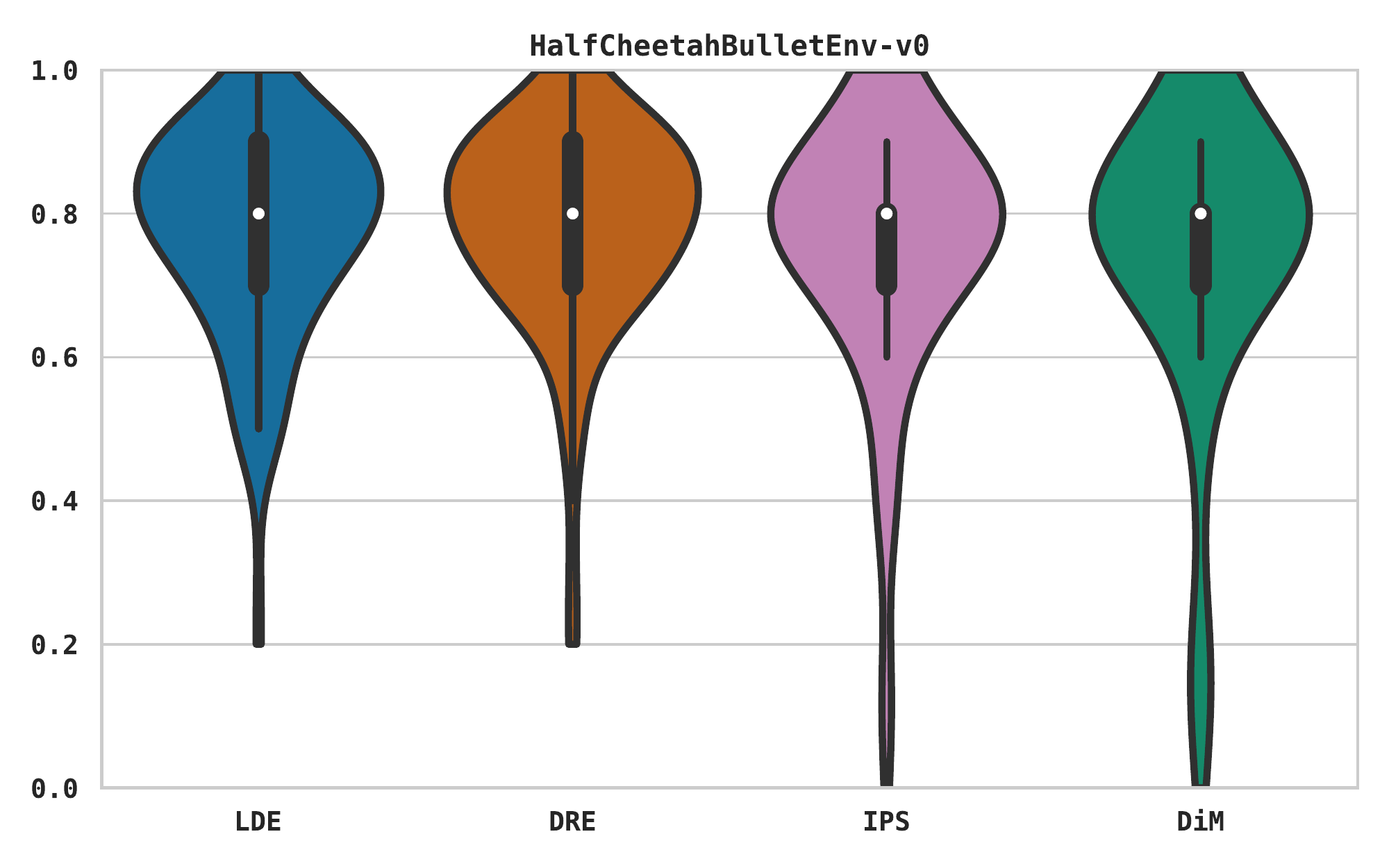}
    \includegraphics[width=.49\linewidth]{./images/scores_AntBulletEnv-v0.pdf}
    \\
    \includegraphics[width=.49\linewidth]{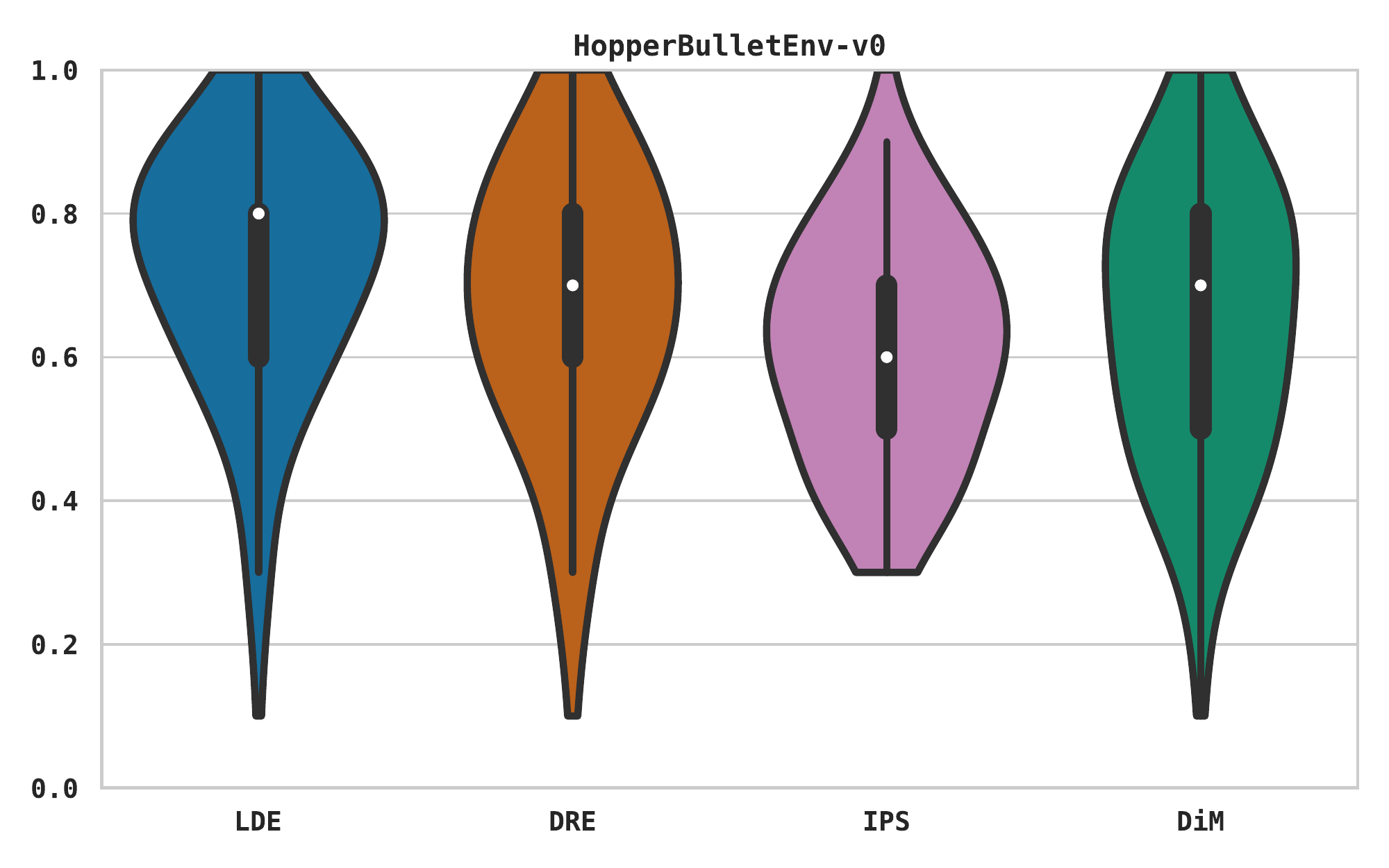}
    \includegraphics[width=.49\linewidth]{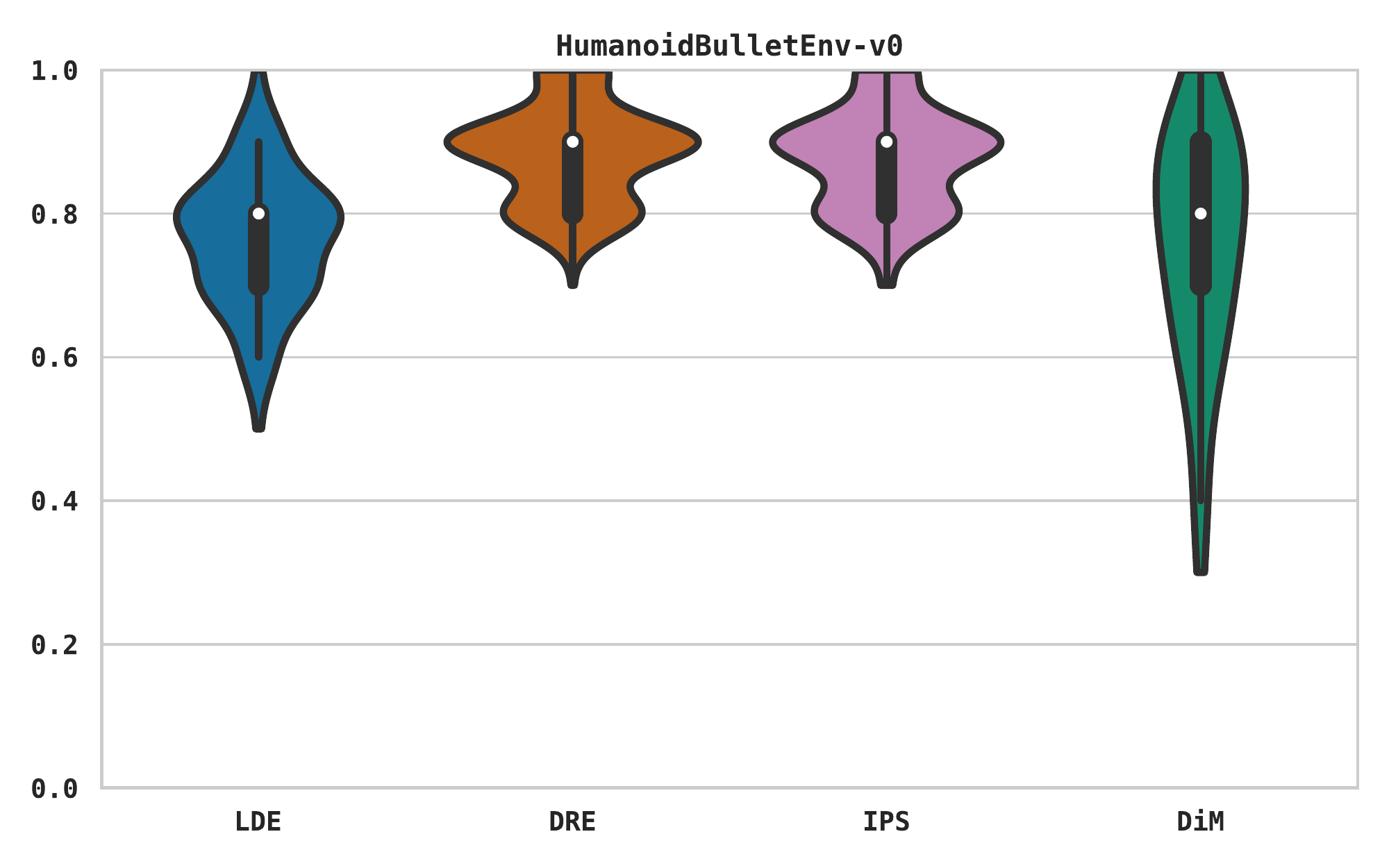}
    \caption{Policy comparison on PyBullet RL environments, see Table~\ref{tab:ex3_appendix}.}
    \label{fig:ex3_comp_appendix}
\end{figure}

For each RL environment we perform 225 tests as follows: step 1 is repeated 15 times and for each set of target policies $\{\pi_{a2s}, \pi_{ddpg}, \pi_{ppo}, \pi_{sac}, \pi_{td3}\}$ step 2 is executed 15 times, i.e. the historical data $\mathcal{D}$ is sampled and the policy comparison is performed.
The resulting policy comparison scores on various RL environments are presented in Table~\ref{tab:ex3_appendix} and in Figure~\ref{fig:ex3_comp_appendix}.

\begin{table}[hbt!]
    \centering\small
    \caption{Policy comparison scores on PyBullet RL environments, computed via~\eqref{eq:comp_score}.}
    \label{tab:ex3_appendix}
    \begin{tabular}{lcccc}
        \toprule
        & \multicolumn{4}{c}{Policy comparison score: average (std)}
        \\\cmidrule(lr){2-5}
        Environment & LDE & DRE & IPS & DiM
        \\\midrule
        \texttt{InvertedPendulum-v0} & \textbf{0.8249} (0.109) & 0.7831 (0.146) & 0.7929 (0.124) & 0.3551 (0.258)
        \\
        \texttt{InvertedPendulumSwingup-v0} & \textbf{0.8533} (0.136) & 0.8222 (0.153) & 0.7693 (0.133) & 0.6987 (0.207)
        \\
        \texttt{Reacher-v0} & \textbf{0.7076} (0.218) & 0.6818 (0.196) & 0.6693 (0.205) & 0.6582 (0.224)
        \\
        \texttt{Walker2D-v0} & \textbf{0.6969} (0.166) & 0.6289 (0.168) & 0.6160 (0.132) & 0.5458 (0.176)
        \\
        \texttt{HalfCheetah-v0} & 0.7756 (0.139) & \textbf{0.7831} (0.137) & 0.7556 (0.174) & 0.7418 (0.203)
        \\
        \texttt{Ant-v0} & \textbf{0.8760} (0.110) & 0.8680 (0.121) & 0.7498 (0.120) & 0.4400 (0.219)
        \\
        \texttt{Hopper-v0} & \textbf{0.7276} (0.171) & 0.6729 (0.188) & 0.6098 (0.154) & 0.6516 (0.191)
        \\
        \texttt{Humanoid-v0} & 0.7636 (0.090) & \textbf{0.8844} (0.067) & 0.8764 (0.070) & 0.7542 (0.160)
        \\\bottomrule
    \end{tabular}
\end{table}

\end{document}